\documentclass{article}

\usepackage[preprint]{neurips_2026}

\usepackage[utf8]{inputenc}
\usepackage[T1]{fontenc}
\PassOptionsToPackage{hyphens}{url}
\usepackage{hyperref}
\usepackage{url}
\usepackage{booktabs}
\usepackage{longtable}
\usepackage{multirow}
\usepackage{siunitx}
\usepackage{amsfonts}
\usepackage{amsmath, amssymb}
\usepackage{nicefrac}
\usepackage{microtype}
\usepackage{xcolor}
\usepackage{graphicx}
\graphicspath{{figures/}}

\title{Probing Routing-Conditional Calibration in
       Attention-Residual Transformers}

\author{%
  \textbf{Wenhao Liang}\,$^{1}$\quad
  \textbf{Lin Yue}\,$^{1}$\quad
  \textbf{Wei Emma Zhang}\,$^{2}$\quad
  \textbf{Miao Xu}\,$^{3}$\\
  \textbf{Mingyu Guo}\,$^{1}$\quad
  \textbf{Olaf Maennel}\,$^{1}$\quad
  \textbf{Weitong Chen}\,$^{2}$\\[0.4em]
  $^{1}$Adelaide University\\
  $^{2}$Australian Institute for Machine Learning (AIML), Adelaide University\\
  $^{3}$The University of Queensland
}

\begin{document}
\maketitle

\begin{abstract}
Post-hoc calibration is usually evaluated as a function of logits
or softmax confidence alone, even as routing-augmented architectures
(MoE, adaptive computation, dynamic-token transformers) increasingly
accompany predictions with sample-specific internal routing
traces and are paired with claims that those traces carry
calibration-relevant uncertainty. We ask a basic question: do these
traces provide stable routing-specific evidence for post-hoc
calibration beyond confidence? We study this question in Attention-Residual
transformers~\citep{team2026attention} through a matched-confidence
diagnostic suite that stratifies examples by routing-derived
state, compares subgroup gaps against within-bin
routing-permutation nulls, and evaluates matched post-hoc probes
differing only in their auxiliary feature. Across our completed
AR runs, scalar routing summaries do not provide stable
evidence of routing-conditional miscalibration: weighted gaps
remain small or seed-sensitive, and only $1$ of $30$ within-bin
permutation tests rejects the conditional-null at
$\alpha\!=\!0.05$ (only on one seed; not stable across seeds in
that cell).
\textbf{AR-CondCal}, a deliberately minimal $2$-D
Nadaraya--Watson probe on confidence and routing-depth variance,
lies within the seed-variance band of matched confidence-only and
predictive-entropy controls and does not reliably improve
worst-routing-tertile ECE; bandwidth-sensitivity checks over
Scott multiples, CV-NLL, and a global-ECE oracle do not change
this. A full-vector MLP over $(c, H_1, \ldots, H_L)$ can appear
to improve over a linear confidence baseline, but the apparent
gain disappears once a confidence-only MLP at the same capacity
is included as a control, and shuffled routing profiles achieve
comparable performance. In the evaluated AR setting, apparent routing-aware calibration
gains should not be read as internal-state calibration until
matched-confidence, bandwidth-sensitivity, capacity-matched, and
permutation controls rule out common confounds.
\end{abstract}

\section{Introduction}
\label{sec:intro}

Deep classifiers often attain high Top-1 accuracy while producing
miscalibrated softmax probabilities~\citep{guo2017calibration,minderer2021revisiting}. Post-hoc calibration is the standard
remedy: fit a small transform of the logits on a held-out set,
leaving the backbone untouched. Standard post-hoc
methods---including temperature-, vector-, histogram-, isotonic-,
and kernel/binning-based calibrators (full set in
Tab.~\ref{tab:calibration_block_ar})---share one structural
property: each is a function of the classifier's softmax (or
logits) alone, with no access to any sample-level internal signal
that the classifier itself computes. Whether calibration error
can be captured by the softmax output alone is the implicit
modelling assumption on which these methods stand.

Dynamic routing and adaptive-computation architectures, from
sparsely gated MoE layers and learned layer-skipping networks to
dynamic-token vision transformers, increasingly expose
sample-specific internal traces alongside the final softmax output,
making those traces a natural auxiliary signal to audit for
post-hoc calibration~\citep{shazeer2017outrageously,fedus2022switch,wang2018skipnet,rao2021dynamicvit}.
We instantiate the question on Attention-Residual
(AR)~\citep{team2026attention} transformers, where each sample
passes through an internal routing trajectory that is not part of
the model's softmax output (mechanism in
\S\ref{sec:method}). A natural extension is then to summarise
this trajectory by an entropy-like scalar feature, analogous to
predictive entropy or top-class confidence. For the AR routing
statistics tested here, this hypothesis fails under
matched-confidence testing: scalar routing summaries do not
yield stable evidence of conditional miscalibration.

On Swin-Tiny~\citep{liu2021swin} $+$ Block-AR
(CIFAR-10~\citep{krizhevsky2009learning}, seed~$0$), the
scalar-routing gap is visually plausible and yields the only
nominal rejection in the $30$-run sweep; however, the rejection
is not stable across the other seeds in the same cell
(\S\ref{sec:rq1}, Tab.~\ref{tab:phenomenon}). \textbf{AR-CondCal}, a controlled matched-kernel probe on
$(c, r_{\mathrm{std}})$ (Fig.~\ref{fig:overview}), tracks the
same picture: it does not reliably improve worst-tertile ECE;
Prop.~2 (App.~\ref{app:theory}) gives the corresponding
confidence-only subgroup blind-spot motivation, and
App.~\ref{app:probe-sanity} shows that full-profile MLP uplift
is not routing-specific under capacity-matched controls.

\textbf{Contributions.}
\textbf{(1)~Matched-confidence falsification protocol} for
routing-conditional calibration claims, combining tertile
stratification, weighted-integrated matched-confidence gaps, and
a within-bin routing-permutation test. The protocol is
architecture-agnostic in design.
\textbf{(2)~Scalar-routing negative diagnostic result.}
Across $30$ completed AR runs, evaluated scalar routing summaries
do not yield stable evidence: only $1/30$ within-bin permutation
tests rejects the conditional-null at $\alpha\!=\!0.05$ (smallest
$p\!=\!0.042$, on one Swin~$+$~Block-AR seed; the other two seeds
in that cell do not reject). The matched-kernel probe
\textbf{AR-CondCal} is similarly within the seed-variance band of
confidence-only and predictive-entropy controls, with no
evaluated bandwidth choice producing a reliable worst-tertile
improvement (App.~\ref{app:bw-sensitivity}).
\textbf{(3)~Capacity-controlled false-positive audit.} A
full-profile MLP appears to improve over a linear
confidence-only baseline, but the gain disappears under a
capacity-matched confidence-only MLP and under shuffled-routing
controls (App.~\ref{app:probe-sanity}), closing a route along
which routing-aware calibration claims could be falsely supported.

\section{Diagnostic protocol}
\label{sec:method}

\begin{figure}[!ht]
  \centering
  \includegraphics[width=0.95\linewidth]{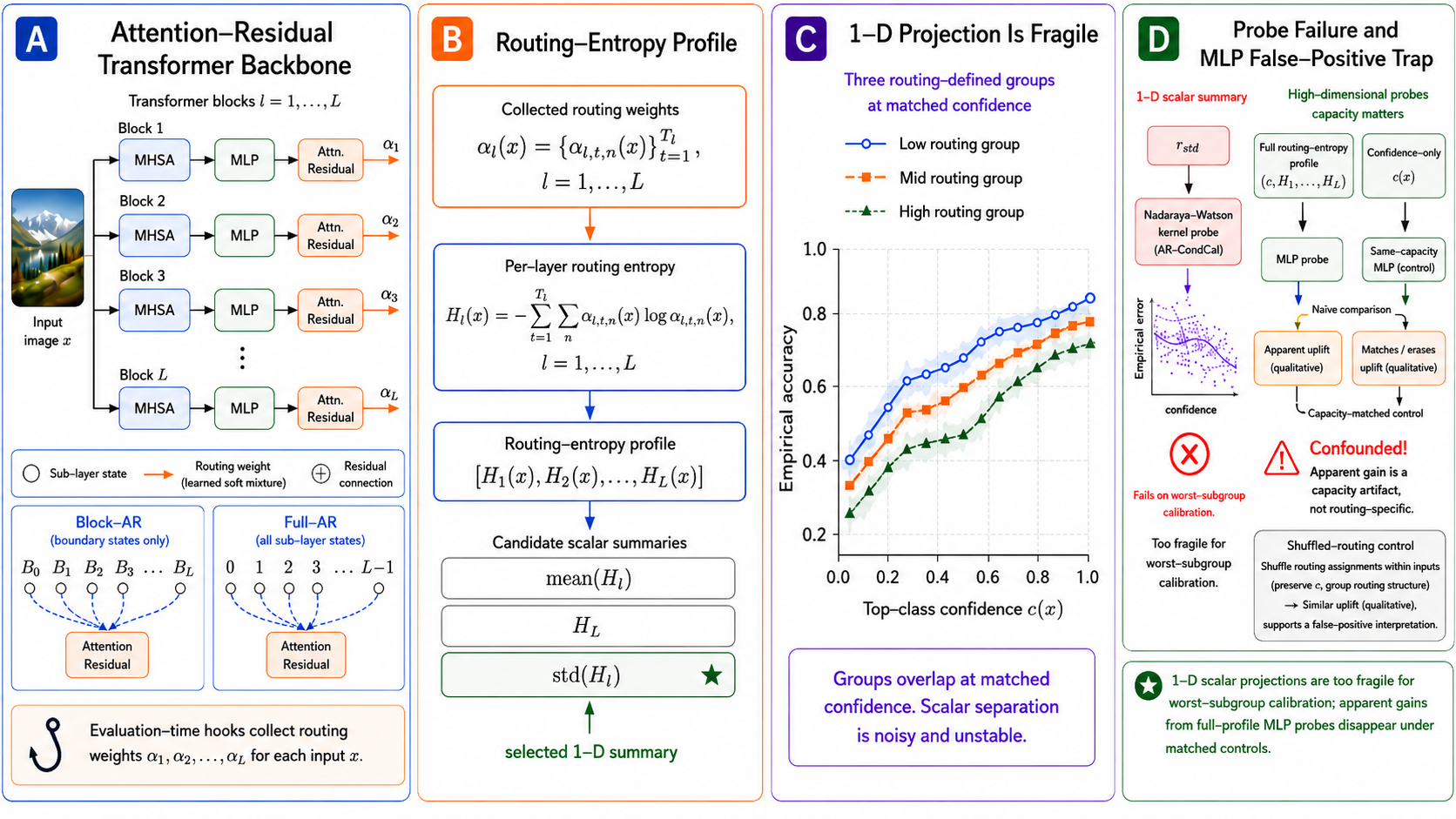}
  \caption{Overview of the routing-conditional calibration
    analysis.
    \emph{(A)} Attention-Residual backbones expose per-sample
    routing weights $\alpha_l(x)$ collected via evaluation-time
    hooks.
    \emph{(B)} Routing-entropy profile: weights define a
    per-layer profile $[H_1(x), \dots, H_L(x)]$, from which we
    select a $1$-D scalar summary (e.g., $r_{\mathrm{std}}(x) =
    \mathrm{std}_l(H_l(x))$).
    \emph{(C)} $1$-D projection is fragile: routing-defined
    groups overlap at matched confidence, and scalar separation
    is noisy and seed-unstable.
    \emph{(D)} Probe failure and MLP false-positive trap:
    AR-CondCal uses the $1$-D summary in a $2$-D kernel
    calibrator but does not reliably improve worst-tertile ECE;
    an MLP on the full $L$-D profile shows an apparent uplift
    that vanishes once capacity-matched confidence-only and
    shuffled-routing controls are added
    (App.~\ref{app:probe-sanity}). Schematic on
    Swin-Tiny~$+$~Block-AR (CIFAR-10, seed-$0$); concrete numbers
    in Tab.~\ref{tab:calibration_block_ar} and
    \S\ref{sec:experiments}.}
  \label{fig:overview}
\end{figure}

Figure~\ref{fig:overview} gives an overview of the three pieces this
section formalises: the routing weights $\alpha_l$ produced by
Attention-Residual sub-layers, the scalar routing-derived features we
extract from them, and the 2-D post-hoc calibrator that consumes
confidence together with one of those features.

\subsection{Preliminaries}
\label{sec:method-arcondcal}
For a $K$-way classifier with logits $z \in \mathbb{R}^K$,
$p = \mathrm{softmax}(z) \in \Delta^{K-1}$, and $c(x) = \max_k p_k$,
output-only post-hoc calibrators fit a map
$\hat\eta : \mathbb{R}^K \to \Delta^{K-1}$ or
$\hat\eta : \Delta^{K-1} \to \Delta^{K-1}$; routing-aware probes
additionally condition on an auxiliary internal feature such as
$r_{\mathrm{std}}(x)$. All accuracy values are Top-$1$ (Acc@1).

Throughout, we report global equal-width ECE (15
bins)~\citep{guo2017calibration,naeini2015obtaining}, equal-mass adaptive ECE
(AdaECE)~\citep{nixon2019measuring}, NLL, Brier score, and (the
core diagnostic metric central to this work) \emph{worst-tertile ECE}: for a fixed per-sample routing feature
$\rho(x)$, split the evaluation set into approximately equal-count
tertiles defined by the empirical $33$rd / $67$th percentiles of
$\rho$ on that same set (ties at the cut-points are assigned by
strict inequality), compute equal-width ECE within each tertile,
and report the worst of the three values (illustrated in
Fig.~\ref{fig:wt-ece-illustration}, App.~\ref{app:wt-ece-illustration}).
In this general definition, $\rho$ denotes the chosen
routing feature under evaluation. Sec.~\ref{sec:rq1} instantiates
$\rho$ as the \emph{aggregate} routing
entropy $r_{\mathrm{agg}}(x) = \tfrac{1}{L}\sum_l H_l(x)$ for the
diagnostic gap analysis, and Tab.~\ref{tab:calibration_block_ar}
instantiates it as the \emph{depth-variance} feature
$r_{\mathrm{std}}(x) = \mathrm{std}_l\, H_l(x)$, matching the feature
that AR-CondCal itself consumes so that worst-tertile ECE directly
tests whether the calibrator fixes the subgroup the method targets.

Formally, let $\mathcal{D}_{\mathrm{eval}} = \{(x_i, y_i)\}_{i=1}^{n}$
be the evaluation set and let $\rho_i = \rho(x_i)$ be the per-sample
routing feature of interest; let $q_{1/3}$ and $q_{2/3}$ be the
empirical $33$rd and $67$th percentiles of $\{\rho_i\}_{i=1}^{n}$ on
$\mathcal{D}_{\mathrm{eval}}$, and define the three tertile index sets
\begin{equation}
\mathcal{S}_{\text{low}}  = \{i : \rho_i \le q_{1/3}\},\quad
\mathcal{S}_{\text{mid}}  = \{i : q_{1/3} < \rho_i \le q_{2/3}\},\quad
\mathcal{S}_{\text{high}} = \{i : \rho_i > q_{2/3}\}.
\end{equation}
Writing $\mathrm{ECE}_{15}(\mathcal{S})$ for the $15$-bin equal-width
ECE~\citep{guo2017calibration,naeini2015obtaining} computed on the sub-sample
indexed by $\mathcal{S}$ (the same definition used for the global ECE
column of Tab.~\ref{tab:calibration_block_ar}), worst-tertile ECE is
\begin{equation}
\mathrm{ECE}^{\mathrm{wt}}(\mathcal{D}_{\mathrm{eval}})
\;=\; \max_{t \in \{\text{low},\,\text{mid},\,\text{high}\}}
\mathrm{ECE}_{15}(\mathcal{S}_{t}).
\end{equation}
In Tab.~\ref{tab:calibration_block_ar} we take
$\mathcal{D}_{\mathrm{eval}}$ to be the $5000$-sample held-out test
half (\S\ref{sec:experiments}); the tertile cut-points $q_{1/3},
q_{2/3}$ are therefore computed on that same $5000$-sample half, not
on the full test set used for the diagnostic gap analysis of
Sec.~\ref{sec:rq1}.

This diagnostic metric is conceptually motivated by the subgroup /
conditional-calibration literature
(multicalibration~\citep{hebert2018multicalibration}, verified
uncertainty calibration~\citep{kumar2019verified}), but the exact
tertile instantiation above is ours and is not a benchmark imported
from a specific prior paper. The metric is a stress test for a
\emph{hypothesised} routing-conditioned failure mode: if a calibrator's
average ECE improvement is driven by easy-routing samples while a
routing-defined subgroup remains miscalibrated, worst-tertile ECE will
catch it; if no such subgroup miscalibration is present, worst-tertile ECE
will simply track global ECE on each tertile.
SmoothECE~\citep{blasiok2023smooth} is reported in the appendix as a
binning-robustness cross-check.

\subsection{Residual routing weights}
The AR sub-layer at depth $l$ attends over $T_l$ prior states
$V_l \in \mathbb{R}^{T_l \times B \times N \times D}$. With
RMS-normalised keys $K_l = \mathrm{RMSN}(V_l)$ and a pseudo-query
$w_l \in \mathbb{R}^D$, the routing weights are
\begin{equation}
\alpha_{l, t, n}(x) = \frac{\exp(w_l^\top K_{l, t, n})}
                           {\sum_{t'} \exp(w_l^\top K_{l, t', n})},
\qquad
\alpha_l \in \mathbb{R}^{T_l \times B \times N}.
\end{equation}
Block-AR caps $T_l$ at the number of block-boundary states at depth
$l$; Full-AR takes $T_l = l$, except that the $l\!=\!1$ sub-layer
($T_1\!=\!1$, degenerate) is uniform-by-construction and is excluded
from the routing-entropy profile. The batch dimension $B$ in this
shape is implementation-only; all subsequent statistics are
computed per input $x$ after averaging over the token index $n$.
We extract $\alpha_l$ at evaluation time by registering forward
hooks on each AR sub-layer; no re-training is required, and the
backbone weights are never touched.

\subsection{From routing weights to a per-sample feature}
The natural per-layer summary is the normalised Shannon entropy
\begin{equation}
H_l(x) = \frac{1}{N \log T_l} \sum_{n=1}^{N}
  \Bigl(-\sum_{t=1}^{T_l} \alpha_{l, t, n}(x) \log \alpha_{l, t, n}(x)\Bigr)
  \in [0, 1],
\end{equation}
where $H_l = 0$ when routing is fully concentrated on a single prior
state and $H_l = 1$ when it is uniform. After the $T_l\!=\!1$
exclusion above, every retained sub-layer satisfies $T_l \ge 2$ and
the $\log T_l$ normaliser is well-defined. Given $L$ AR sub-layers, each sample
yields a vector $(H_1(x), \ldots, H_L(x))$, and the question
is how to summarise it into a scalar that helps calibration. We test
seven candidates (Tab.~\ref{tab:feature_ablation}): aggregate
entropy $r_{\mathrm{agg}}(x) = \tfrac{1}{L}\sum_l H_l(x)$, last-layer
entropy $H_L$, the \emph{depth-variance} $r_{\mathrm{std}}(x) =
\mathrm{std}_l(H_l(x))$, label-free min--max rescaled to $[0,1]$,
the concentration $1 - r_{\mathrm{agg}}(x)$, entropy-weighted
confidence, the non-routing control predictive entropy, and a $1$-D
confidence-only control. As established by our $30$-run AR
sweep (\S\ref{sec:rq2}, Tab.~\ref{tab:ar-sweep}), the
feature-ablation comparison (Tab.~\ref{tab:feature_ablation}),
and the capacity-controlled audit (\S\ref{sec:probe},
App.~\ref{app:probe-sanity}), compressing the per-layer profile
into any single scalar yields a fragile $1$-D projection:
across all seven candidates the point-estimate ECE range is
narrow (Tab.~\ref{tab:feature_ablation}) and indistinguishable
from confidence-only under the protocol's CIs. We carry forward two
named features as canonical $1$-D projections used elsewhere in
the paper: $r_{\mathrm{agg}}$ for the diagnostic gap analysis in
\S\ref{sec:rq1}, and $r_{\mathrm{std}}$ as the second calibration
feature used by AR-CondCal to test whether a single-feature
kernel smoother recovers any signal beyond confidence;
\S\ref{sec:probe} reports the answer under capacity-matched
controls.




\subsection{AR-CondCal}
\label{sec:ar-condcal}

\textbf{Rather than proposing a state-of-the-art calibrator, we
design AR-CondCal as a deliberately minimal diagnostic tool} to
test whether one routing-derived scalar is low-dimensional and
locally smooth enough for a Nadaraya--Watson estimator to exploit.
It is included in the comparison as a controlled $2$-D matched-kernel
probe; it is not tuned to maximise any global metric, and its role is
to expose the limitations of scalar routing summaries on this substrate
(\S\ref{sec:rq3}).

Let the calibration set be
\(\mathcal{D}_{\mathrm{cal}} = \{(c_i, r_i, t_i)\}_{i=1}^{n}\), where
\(c_i = c(x_i) = \max_k p_k(x_i)\) is the top-class confidence,
\(r_i = r_{\mathrm{std}}(x_i) = \operatorname{std}_{l} H_l(x_i)\)
is the depth-variance routing feature, and \(t_i\) is the correctness
indicator:
\begin{equation}
t_i = \mathbf{1}\{y_i = \hat{y}_i\},
\qquad
\hat{y}_i = \operatorname*{arg\,max}_{k} p_k(x_i).
\end{equation}
AR-CondCal targets the conditional top-$1$ correctness probability
\(g(c, r) = \Pr(t = 1 \mid c, r)\), estimated by a $2$-D
Nadaraya--Watson regression~\citep{nadaraya1964estimating,watson1964smooth}
with Gaussian kernel \(K_h\):
\begin{equation}
\hat{g}(c, r)
=
\frac{
\sum_{i=1}^{n} K_h\bigl(c - c_i,\, r - r_i\bigr)\, t_i
}{
\sum_{i=1}^{n} K_h\bigl(c - c_i,\, r - r_i\bigr)
}.
\end{equation}
We use per-dimension Scott's-rule bandwidths~\citep{scott2015multivariate}
\(h_j = \sigma_j n^{-1/(m+4)}\), where \(m=2\) is the feature dimension,
\(j \in \{c,r\}\), and \(\sigma_j\) is computed on
\(\mathcal{D}_{\mathrm{cal}}\).

\textbf{Inference.}
We project the kernel-predicted target into the range achievable
by per-sample temperature scaling,
$\tilde c(x) = \Pi_{[1/K + \epsilon,\, 1 - \epsilon]}\bigl(\hat g(c(x), r_{\mathrm{std}}(x))\bigr)$,
since $\mathrm{softmax}(z/\tau)_{\hat y} \in (1/K, 1)$ for any
$\tau > 0$; we then solve
$\mathrm{softmax}(z(x)/\tau(x))_{\hat y(x)} = \tilde c(x)$ by
bisection in $\tau$. Across the evaluated cells, the pooled
clipping rates are $0.02\%$ at the lower bound and $0.06\%$ at
the upper bound; the worst per-cell rate is $0.10\%$ at either
bound, and metrics are unchanged at four-decimal precision
(App.~\ref{app:impl}). Top-$1$ accuracy is preserved.
AR-CondCal has no parametric training step, no backbone
modification, and no tuning beyond the standard bandwidth rule (for the minimal parameter overhead of the underlying
AR backbone itself, see App.~\ref{app:overhead}).

\subsection{Scope of the AR-CondCal construction}
AR-CondCal is a minimal probe: the standard $2$-D-kernel
calibrator family, with one routing-derived feature
$r_{\mathrm{std}}$ in place of the second auxiliary. Within the
matched-kernel calibrator family with a single
architecture-internal feature, AR-CondCal is competitive on global ECE
(Tab.~\ref{tab:calibration_block_ar}). The probe is not a safety,
robustness, or OOD-detection method. The feature choice
is empirical and substrate-dependent
(Tab.~\ref{tab:feature_ablation}). Appendix~\ref{app:theory} gives three supporting propositions
and a Nadaraya--Watson consistency fact; none is a finite-sample
ECE-dominance claim.

\section{Experiments}
\label{sec:experiments}

The main illustrative calibration benchmark in
Tab.~\ref{tab:calibration_block_ar} and
Fig.~\ref{fig:bench-main} uses Swin-Tiny Block-AR on
CIFAR-10. The diagnostic sweeps in \S\ref{sec:rq2} and
\S\ref{sec:probe} aggregate over the completed
protocol-matched AR cells described in
App.~\ref{app:replication-plan}.\footnote{We deploy the
protocol on Swin / DeiT~\citep{touvron2021training} /
ViT~\citep{dosovitskiy2020image} runs on CIFAR-10 and CIFAR-100;
single-seed Tiny-ImageNet ViT is descriptive only. The
empirical conclusion is bounded to the evaluated AR setting
and the evaluated probe family.} \textbf{The main Swin-Tiny /
DeiT calibration benchmarks are aggregated over three
independent training seeds} (Tab.~\ref{tab:calibration_block_ar},
Fig.~\ref{fig:bench-main}, Tab.~\ref{tab:ar-sweep}).
Tab.~\ref{tab:ar-sweep} also includes clearly marked single-seed
ViT-B/$16$ pilot rows as descriptive scope checks.
Protocol-matched replication arms for additional datasets are
summarised in Appendix~\ref{app:replication-plan}. We evaluate on
two deliberately distinct populations, documented here once and
referenced by name throughout.

\noindent\textbf{Evaluation populations.}
The \emph{diagnostic gap analysis} (Tab.~\ref{tab:phenomenon},
Fig.~\ref{fig:priority2}, and Fig.~\ref{fig:priority2-full} in
Appendix~\ref{app:full-ar-pathology}) is computed on the full
$10{,}000$-sample test set. It does not fit any calibrator, so no
held-out split is needed; we prefer the larger sample because the
per-bin max-gap statistic is unstable at $n=5000$ (small routing-tail
bins produce artificial spikes). The \emph{calibration benchmark}
(Tab.~\ref{tab:calibration_block_ar}) uses a $50/50$ split of the
test set with seed~$42$: the first 5000 samples are the calibration
set on which every post-hoc calibrator is fit, and the second 5000
samples are the held-out test half on which all
Tab.~\ref{tab:calibration_block_ar} metrics are computed. Bootstrap
CIs come from 500 resamples of this 5000-sample test half. These are
two different populations by design; the main-body text notes the
choice wherever it matters. The choice of an equal $50/50$ split
rather than a smaller calibration fraction is defended quantitatively
in Appendix~\ref{app:calsize} (Fig.~\ref{fig:calsize}): at
$n_{\mathrm{cal}} = 5000$ every method in
Tab.~\ref{tab:calibration_block_ar} is at or near its sample-size
plateau, so the method-ranking we report is not an artifact of an
aggressive calibration-split choice. Method rankings are
descriptive only (no pairwise multiplicity correction); our
conclusion rests on matched-control and permutation patterns.

\noindent\textbf{Metrics.}
Unless stated otherwise, models are trained from scratch for 300
epochs on Swin-Tiny, DeiT-Small, and ViT-B/$16$, with three
independent seeds ($s_0,s_1,s_2$) for the multi-seed Swin-Tiny and
DeiT-Small cells; the ViT-B/$16$ cells are descriptive single-seed
runs. The main comparison
(Tab.~\ref{tab:calibration_block_ar}) reports equal-width ECE
($15$ bins; binning-based), AdaECE ($15$ equal-count bins;
binning-robust), NLL, Brier, and our local worst-group metric
(\emph{worst-tertile ECE}, defined in
Sec.~\ref{sec:method-arcondcal}). AR-CondCal preserves the argmax;
not all multiclass baselines do, so we interpret
Tab.~\ref{tab:calibration_block_ar} via calibration metrics and
matched-control comparisons, not accuracy ranking. ECE, AdaECE,
and worst-tertile
ECE are reported as mean $\pm 1$ std across the three training
seeds. NLL and Brier are reported as seed means in the main
table, with paired $\Delta$NLL$/\Delta$Brier relative to the
uncalibrated model reported in App.~\ref{app:paired-delta} to
isolate calibrator-induced effects from cross-seed
raw-model-quality variance. Within each seed, all calibrators use the same seed-$42$
$50/50$ cal/test split defined above. We do not rely on any single ECE-style metric: MCE, Classwise ECE, and
SmoothECE~\citep{blasiok2023smooth} (a kernel-smoothed
calibration error robust to binning choices) are in the
appendix (Tab.~\ref{tab:cwece-supp}); we use them to
cross-check rather than re-rank the main table.

\textbf{Baselines.}
Our external-literature baselines are Temperature Scaling (TS), Vector
Scaling (VS), and Classwise TS~\citep{guo2017calibration}; Ensemble
TS~\citep{zhang2020mix}; Parametric TS~\citep{tomani2022parameterized}; Histogram
Binning~\citep{zadrozny2001obtaining}; BBQ~\citep{naeini2015obtaining}; Isotonic
Regression~\citep{zadrozny2002transforming}; and SB-ECE
TS~\citep{karandikar2021soft}. For sanity-check comparisons, we
also include two internal comparators: LC (a post-hoc
logit-normalisation baseline inspired by, but distinct from, the
training-time LogitNorm loss~\citep{wei2022mitigating}) and RCMMC
(routing-conditioned monotone margin-and-entropy binning). Neither
is externally published and neither is part of this paper's
contribution; both remain in the comparison as internal sanity
checks, with full definitions in
Appendix~\ref{app:internal-baselines}. We additionally report a
\emph{matched-kernel control suite} (\textbf{NW family}, last three
rows of Tab.~\ref{tab:calibration_block_ar}): an identical $2$-D
Nadaraya--Watson estimator with the same bandwidth applied to
$(c)$ alone, $(c,\mathrm{PredEntropy})$, and
$(c,r_{\mathrm{std}})\!=\!\textbf{AR-CondCal}$. The three rows
share kernel family and bandwidth and differ only in the second
feature.

\subsection{Diagnostic protocol and the fragility of the 1-D
matched-confidence gap}
\label{sec:rq1}

\begin{figure}[t]
  \centering
  \includegraphics[width=\linewidth]{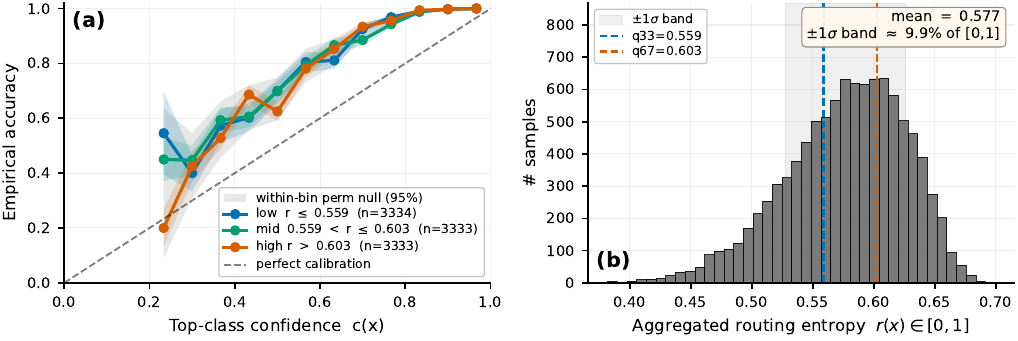}
  \caption{\textbf{Matched-confidence gaps on Swin-Tiny + Block-AR
    seed-$0$} (CIFAR-10, ep $=$ 299, full test set).
    \emph{(a)} Top-1 accuracy vs top-class confidence by
    tertile of $r_{\mathrm{agg}}(x)$; gray band $=$ within-bin
    permutation null at $95\%$. \emph{(b)} $r_{\mathrm{agg}}(x)$
    distribution. Quantitative gap statistics in
    Tab.~\ref{tab:phenomenon}; this seed produces the only
    seed-level nominal rejection ($p\!=\!0.042$) in the $30$-run
    sweep (\S\ref{sec:rq2}, Tab.~\ref{tab:ar-sweep}), and the
    rejection is not stable across the other seeds in the same
    cell.}
  \label{fig:priority2}
\end{figure}

Stratifying the full $10{,}000$-sample test set by tertile of
the aggregate routing entropy
$r_{\mathrm{agg}}(x) = \tfrac{1}{L}\sum_l H_l(x)$ produces three
populations of roughly equal size. Within each tertile we bin by
top-class confidence in $15$ equal-width bins and plot empirical
accuracy (Fig.~\ref{fig:priority2}). Across the confidence bins
shared by both tertiles, we report the max absolute low-vs-high
tertile gap, a weighted-integrated mean of that gap, a
$5000$-resample bootstrap CI on each, and a within-confidence-bin
permutation $p$-value against the conditional-null (robustness
checks in App.~\ref{app:gap-robustness}).

Under our rigorous evaluation protocol on
Swin-Tiny~\citep{liu2021swin} CIFAR-10 seed-$0$
(Tab.~\ref{tab:phenomenon}\footnote{Per-seed raw Acc@1 for these
checkpoints is reported in
Tab.~\ref{tab:perseed-swin-c10-diagnostic}.},
Fig.~\ref{fig:priority2}), Block-AR
yields a max-gap of $0.346$ with within-bin permutation
$p\!=\!0.042$, the \emph{only} seed-level nominal rejection in
the $30$-run AR sweep (\S\ref{sec:rq2}, Tab.~\ref{tab:ar-sweep}).
Full-AR yields max-gap $0.052$ and $p\!=\!0.888$, well inside the
within-bin permutation null. Under a global null, $30$ tests at
$\alpha\!=\!0.05$ yield $\sim\!1.5$ expected nominal rejections,
and $p\!=\!0.042$ does not survive a Bonferroni correction
($\alpha/30\!\approx\!0.0017$). The $1/30$ rejection therefore
reads as a seed-level \emph{instability example} of a noisy $1$-D
projection, not a stable substrate-level failure mode: \textbf{subgroups defined
by a single routing-derived scalar are not consistently separable
from the within-bin permutation null in this cell.}

\subsection{The fragility of 1-D routing projections}
\label{sec:rq2}

While Tab.~\ref{tab:phenomenon} presents the detailed autopsy
of the seed that yields the only nominal rejection, the same
fragility is
consistently observed across our completed protocol-matched
substrate sweep. The same protocol applied across
$30$ runs (Swin-Tiny, DeiT-Small, ViT-B/$16$; Block-AR and
Full-AR; multiple seeds on CIFAR-10/CIFAR-100, plus
single-seed Tiny-ImageNet ViT descriptive rows) supports the
same picture. Per-run max-gap spans $0.049$ to $0.500$,
weighted-integrated gap spans $0.010$ to $0.035$, and within-bin
permutation $p$-values span $0.042$ to $0.949$; $1$ of $30$ runs
rejects the conditional-null at $\alpha\!=\!0.05$ (smallest
$p\!=\!0.042$, on Swin-Tiny + Block-AR ($b\!=\!2$) seed-$0$).
The point-estimate Block versus Full ordering is not preserved
on either backbone (DeiT seed-$0$: Block $0.095$, Full $0.124$),
and the max-gap is strongly seed-sensitive (Swin-Tiny Block-AR
at $b\!=\!2$: max-gap $0.346$ at seed-$0$ vs $0.100$ at
seed-$2$).

\textbf{The $1$-D projection of routing uncertainty is unstable
across seeds and is not a reliable target for
single-routing-feature non-parametric post-hoc calibration under
this protocol.} The max-over-bins gap compounds
two projections (first to a tertile membership indicator on a
single routing scalar, then to the extreme of $15$ confidence
bins), and the resulting summary inherits the noise of both.
The single-cell point estimate cannot be read as evidence of a
hypothesised routing-conditioned failure mode; the same picture holds across
the full $30$-run AR sweep
(Tab.~\ref{tab:ar-sweep}, App.~\ref{app:replication-plan}).
The remaining question is whether the calibration-relevant
routing content exists in the underlying data, or only in the
noise floor of this projection. \S\ref{sec:rq3} and
\S\ref{sec:probe} answer it.

\begin{figure}[t]
  \centering
  \includegraphics[width=\linewidth]{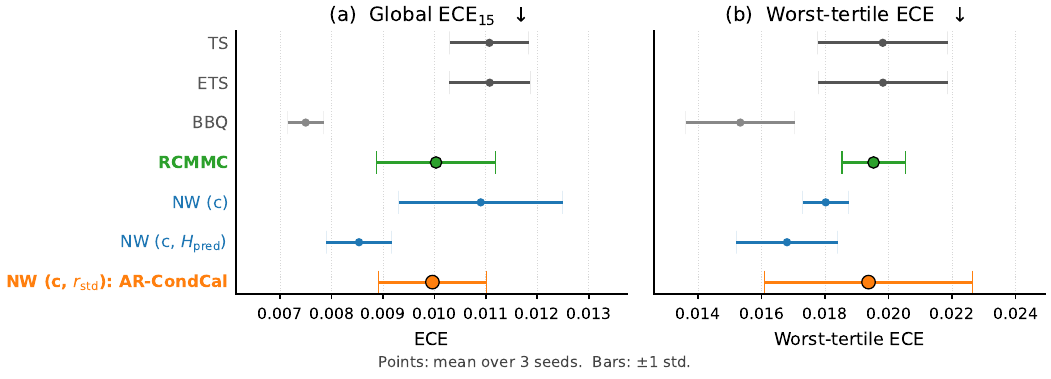}
  \caption{Global ECE$_{15}$ (left) and worst-tertile ECE (right)
    on the CIFAR-10 diagnostic benchmark (Block-AR, Swin-Tiny,
    ep $=$ 299). Points denote the mean over three training seeds
    ($s_0,s_1,s_2$); horizontal bars denote $\pm 1$ standard
    deviation. The last three rows are the matched-kernel
    \textbf{NW family}: an identical $2$-D Nadaraya--Watson
    estimator with the same bandwidth applied to $(c)$ alone, to
    $(c, H_{\mathrm{pred}})$, and to $(c, r_{\mathrm{std}})\!=\!$
    \textbf{AR-CondCal}. \emph{The routing feature does not
    reliably separate AR-CondCal from the matched non-routing
    controls on either metric:} the three NW-family rows fall
    within each other's seed-variance bands on global ECE, and
    on worst-tertile ECE the matched-kernel rows
    (Conf-only, Conf$+$PredEntropy, AR-CondCal) all sit within a
    narrow $\sim\!0.003$ std band of each other. Full method list (all $14$ calibrators)
    in App.~\ref{app:bench-main-full},
    Fig.~\ref{fig:bench-main-full}.}
  \label{fig:bench-main}
\end{figure}

\begin{table}[t]
\centering
\setlength{\tabcolsep}{3pt}
\footnotesize
\caption{\textbf{Diagnostic protocol on Swin-Tiny CIFAR-10 seed-$0$.}
Samples are stratified by tertile of aggregate routing entropy
$r_{\mathrm{agg}}$ and compared at matched top-class confidence.
Max gap is the largest absolute low-vs-high tertile accuracy gap
over confidence bins shared by both tertiles among $15$
equal-width bins. Wt.\ gap is the bin-support-weighted mean gap,
with a percentile bootstrap $95\%$ CI ($B\!=\!5000$). Bins reports
shared/total confidence bins; Min/$q_{25}$/Med reports
per-shared-bin support
$\min(n_{\mathrm{low}}, n_{\mathrm{high}})$, ruling out empty-tail
artifacts. Null $q_{.975}$ is the $97.5$th percentile of the
within-bin routing-permutation null distribution for the max-gap
statistic ($P\!=\!5000$). On Block-AR the observed Max gap of
$0.346$ produces a within-bin permutation $p\!=\!0.042$, the
\emph{only} seed-level nominal rejection in the $30$-run AR sweep
(\S\ref{sec:rq2}, Tab.~\ref{tab:ar-sweep}). On Full-AR the Max gap
is well within the null and the permutation test does not reject. Percentile-bootstrap CIs for
absolute-gap statistics need not contain the original point
estimate; see App.~\ref{app:gap-robustness}.
Statistics illustrated in Fig.~\ref{fig:tab1-illustration}.}
\label{tab:phenomenon}
\begin{tabular}{lccccccc}
\toprule
Variant & Max gap & Wt.\ gap & $95\%$ CI (Wt.) & Bins & Min/$q_{25}$/Med & Null $q_{.975}$ & Perm.\ $p$ \\
\midrule
Block-AR & $0.346$ & $0.016$ & $[0.012,\,0.030]$ & $12/15$ & $11/86/143$ & $0.379$ & $\mathbf{0.042}$ \\
Full-AR  & $0.052$ & $0.016$ & $[0.018,\,0.043]$ & $13/15$ & $41/222/280$ & $0.204$ & $0.888$ \\
\bottomrule
\end{tabular}
\end{table}

\subsection{A matched-kernel probe exposes the limits of scalar
routing summaries}
\label{sec:rq3}

If the $1$-D routing-tertile gap is dominated by noise
(\S\ref{sec:rq2}), what does a calibrator built on top of it
do? AR-CondCal is the minimal probe of that question. It is a
$2$-D Nadaraya--Watson estimator on $(c, r_{\mathrm{std}})$ that
shares a kernel family with the $1$-D Conf-only and $2$-D
Conf$+$predictive-entropy controls and swaps in one
routing-derived scalar. The qualitative pattern below
replicates across an $8$-cell $\times$ $3$-seed cross-substrate
sweep (Figs.~\ref{fig:cross-cell-main}--\ref{fig:cross-cell-supp},
App.~\ref{app:cross-cell-sweep}); we present the Swin-Tiny
CIFAR-10 Block-AR cell here for continuity.

To isolate what the routing feature buys at fixed estimator
family, we fit three calibrators with shared kernel and varying
second feature. On global ECE (\%, $3$-seed sweep,
Fig.~\ref{fig:bench-main} / Tab.~\ref{tab:calibration_block_ar}),
AR-CondCal ($1.00\!\pm\!0.10$) is indistinguishable from its
matched controls Conf-only ($1.09\!\pm\!0.16$) and
Conf$+$PredEntropy ($0.85\!\pm\!0.06$): the routing feature does
not separate within the NW family. The non-matched columns of
Tab.~\ref{tab:calibration_block_ar} (TS/ETS $1.11\!\pm\!0.08$,
RCMMC $1.00\!\pm\!0.12$, BBQ $0.75\!\pm\!0.04$) span a similar
range; AR-CondCal is not a SOTA claim.

\textbf{Worst-tertile ECE is not reliably improved by the
routing feature.} The mean worst-tertile ECE of AR-CondCal (\%)
($1.94\!\pm\!0.33$) is comparable to Conf-only
($1.80\!\pm\!0.07$) and Conf$+$PredEntropy
($1.68\!\pm\!0.16$, right panel of
Fig.~\ref{fig:bench-main}); the routing feature does not move
worst-tertile ECE beyond the seed-variance band of the matched
non-routing controls. Adding a valid routing-derived feature to a
$2$-D kernel estimator does not yield a stable, reliable
resolution for the most miscalibrated subgroup; kernel and
binning calibrators have worst-tertile seed-std an order of
magnitude wider than TS/ETS. Across the cross-substrate sweep no advantage separates from
seed variance (Fig.~\ref{fig:cross-cell-main}). Read alongside
\S\ref{sec:rq2}, this is consistent with the confidence-only
blind-spot bound in Proposition~2 and the bandwidth-shrinkage
calculation in Proposition~4 (App.~\ref{app:theory}). A
bandwidth-sensitivity sweep over Scott multiples, CV-NLL, and a
global-ECE oracle (App.~\ref{app:bw-sensitivity}; CIFAR-100
temporal-ensembling cross-check in
App.~\ref{app:temporal-cross-dataset}) preserves the same
picture: global ECE moves modestly, worst-tertile ECE does not.
Reliability diagrams for $4$ representative AR cells are in
App.~\ref{app:reliability}; per-cell full-method calibration
tables for the other six $3$-seed AR cells are in
App.~\ref{app:per-cell-cal}.

\begin{table}[t]
\centering
\setlength{\tabcolsep}{2pt}
\small
\caption{\textbf{Post-hoc calibration on Block-AR (Swin-Tiny, CIFAR-10, ep $=$ 299; $5000$-sample test half, seed~$42$ split, $3$ training seeds).} ECE, AdaECE, worst-tert.\ ECE: mean\,$\pm$\,$1$ std over seeds. NLL, Brier: seed means (pooled cross-seed std is dominated by raw-model quality variance; paired $\Delta$NLL/$\Delta$Brier vs the uncalibrated model in Tab.~\ref{tab:paired-delta}, App.~\ref{app:paired-delta}). AR-CondCal is a routing-feature kernel probe targeting top-$1$ correctness, not a SOTA calibrator; NLL / Brier are reported as guardrails, not optimisation targets. \textbf{Bold} = best mean; \underline{underlined} = $2$nd-best (raw row excluded); ranking markers are for orientation, not statistically resolved. Worst-tert.\ ECE splits the test half into tertiles of $r_{\mathrm{std}}$. $\Delta$Acc@$1$ is relative to the uncalibrated row. MCE / Classwise ECE / SmoothECE in Tab.~\ref{tab:cwece-supp}.}
\label{tab:calibration_block_ar}
\begin{tabular}{p{2.6cm}cccccc}
\toprule
Method & ECE $\downarrow$ & AdaECE $\downarrow$ & NLL $\downarrow$ & Brier $\downarrow$ & Worst-tert.\ ECE $\downarrow$ & $\Delta$Acc@$1$ (\%) \\
\midrule
No calibration & $0.1514\,{\scriptstyle\pm 0.0028}$ & $0.1514\,{\scriptstyle\pm 0.0028}$ & $0.427$ & $0.175$ & $0.1581\,{\scriptstyle\pm 0.0049}$ & $0.00\,{\scriptstyle\pm 0.00}$ \\
Temp. Scaling (TS) & $0.0111\,{\scriptstyle\pm 0.0008}$ & $0.0123\,{\scriptstyle\pm 0.0004}$ & $0.295$ & $0.146$ & $0.0198\,{\scriptstyle\pm 0.0020}$ & $0.00\,{\scriptstyle\pm 0.00}$ \\
Ensemble TS & $0.0111\,{\scriptstyle\pm 0.0008}$ & $0.0123\,{\scriptstyle\pm 0.0004}$ & $0.295$ & $0.146$ & $0.0198\,{\scriptstyle\pm 0.0020}$ & $0.00\,{\scriptstyle\pm 0.00}$ \\
Vector Scaling & $0.0094\,{\scriptstyle\pm 0.0022}$ & $0.0106\,{\scriptstyle\pm 0.0025}$ & $\mathbf{0.288}$ & $\mathbf{0.142}$ & $0.0169\,{\scriptstyle\pm 0.0022}$ & $+0.24\,{\scriptstyle\pm 0.16}$ \\
Classwise TS & $0.0098\,{\scriptstyle\pm 0.0010}$ & $0.0115\,{\scriptstyle\pm 0.0004}$ & $0.294$ & $0.145$ & $0.0182\,{\scriptstyle\pm 0.0009}$ & $0.00\,{\scriptstyle\pm 0.00}$ \\
Parametric TS & $0.0113\,{\scriptstyle\pm 0.0010}$ & $\underline{0.0094\,{\scriptstyle\pm 0.0005}}$ & $0.298$ & $0.146$ & $0.0179\,{\scriptstyle\pm 0.0007}$ & $0.00\,{\scriptstyle\pm 0.00}$ \\
Histogram Binning & $0.0100\,{\scriptstyle\pm 0.0016}$ & $0.0099\,{\scriptstyle\pm 0.0026}$ & $0.323$ & $0.150$ & $0.0206\,{\scriptstyle\pm 0.0045}$ & $-0.01\,{\scriptstyle\pm 0.01}$ \\
BBQ & $\mathbf{0.0075\,{\scriptstyle\pm 0.0004}}$ & $0.0127\,{\scriptstyle\pm 0.0036}$ & $0.324$ & $0.150$ & $\mathbf{0.0153\,{\scriptstyle\pm 0.0017}}$ & $0.00\,{\scriptstyle\pm 0.00}$ \\
Isotonic Regression & $0.0108\,{\scriptstyle\pm 0.0020}$ & $0.0111\,{\scriptstyle\pm 0.0017}$ & $0.323$ & $0.150$ & $0.0195\,{\scriptstyle\pm 0.0025}$ & $-0.01\,{\scriptstyle\pm 0.01}$ \\
SB-ECE TS & $0.0105\,{\scriptstyle\pm 0.0010}$ & $0.0109\,{\scriptstyle\pm 0.0008}$ & $0.295$ & $0.145$ & $0.0184\,{\scriptstyle\pm 0.0041}$ & $0.00\,{\scriptstyle\pm 0.00}$ \\
\midrule
RCMMC & $0.0100\,{\scriptstyle\pm 0.0012}$ & $0.0101\,{\scriptstyle\pm 0.0012}$ & $\underline{0.293}$ & $\underline{0.145}$ & $0.0195\,{\scriptstyle\pm 0.0010}$ & $0.00\,{\scriptstyle\pm 0.00}$ \\
Conf-only & $0.0109\,{\scriptstyle\pm 0.0016}$ & $0.0099\,{\scriptstyle\pm 0.0018}$ & $0.299$ & $0.146$ & $0.0180\,{\scriptstyle\pm 0.0007}$ & $0.00\,{\scriptstyle\pm 0.00}$ \\
Conf + PredEntropy & $\underline{0.0085\,{\scriptstyle\pm 0.0006}}$ & $\mathbf{0.0090\,{\scriptstyle\pm 0.0025}}$ & $0.299$ & $0.146$ & $\underline{0.0168\,{\scriptstyle\pm 0.0016}}$ & $0.00\,{\scriptstyle\pm 0.00}$ \\
\textbf{AR-CondCal} & $0.0100\,{\scriptstyle\pm 0.0010}$ & $0.0100\,{\scriptstyle\pm 0.0017}$ & $0.306$ & $0.148$ & $0.0194\,{\scriptstyle\pm 0.0033}$ & $0.00\,{\scriptstyle\pm 0.00}$ \\
\bottomrule
\end{tabular}
\end{table}

\subsection{Capacity-controlled audit of the full-profile probe}
\label{sec:probe}

To audit whether the candidate routing signal is destroyed by
dimensionality reduction or smoothing, we run a supervised probe
with no $1$-D projection: a parametric two-layer MLP on
$(c, H_1, \ldots, H_L)$ predicting
$|\mathrm{conf}(x){-}\mathrm{correct}(x)|$, against linear and
capacity-matched non-linear controls.

\textbf{Naive comparison suggests uplift.} Against a linear
confidence-only baseline, the full-vector MLP shows a positive
held-out $R^2$ uplift on the three Block-AR cells in
Tab.~\ref{tab:routing-probe} (mean $+0.074\!\pm\!0.038$),
which in isolation could suggest signal in the full profile.

\textbf{The naive comparison is insufficient.} It confounds
routing information with non-linear model capacity. We therefore
add a \emph{capacity-matched confidence-only MLP} (same
architecture, $c$ as sole input), and a \emph{shuffled-routing
full-vector MLP} (training-fold $H$ permuted across samples).
Across $18$ seed-level cells in $10$ substrate $\times$ variant
groups (App.~\ref{app:probe-sanity} for breakdown), the conf-only
MLP beats the full-vector MLP in pooled held-out
$R^2$ ($0.602\!\pm\!0.094$ vs $0.559\!\pm\!0.140$) and on $9/10$
per-substrate aggregates (full $-$ conf-MLP range $-0.21$ to
$-0.004$; c10 Swin Block-AR slightly positive at
$+0.022\!\pm\!0.009$, within seed std); shuffling $H$ within the
training fold leaves the full-vector $R^2$ essentially
unchanged ($\Delta\!=\!-0.002$). Full per-model pooled values
are tabulated in App.~\ref{app:probe-sanity}.

\textbf{Conclusion.}
The apparent uplift over the linear confidence baseline reflects
non-linear modelling of confidence and increased input dimensionality,
not routing-specific information. We do not claim routing is
universally irrelevant; under the probes evaluated here, no stable
routing-specific signal survives the controls. Fig.~\ref{fig:mechanism}b
shows this visually: $r_{\mathrm{std}}$ distributions for correct and
incorrect predictions overlap on Block-AR ($\rho\!\approx\!-0.01$, n.s.).

\begin{figure}[t]
  \centering
  \includegraphics[width=\linewidth]{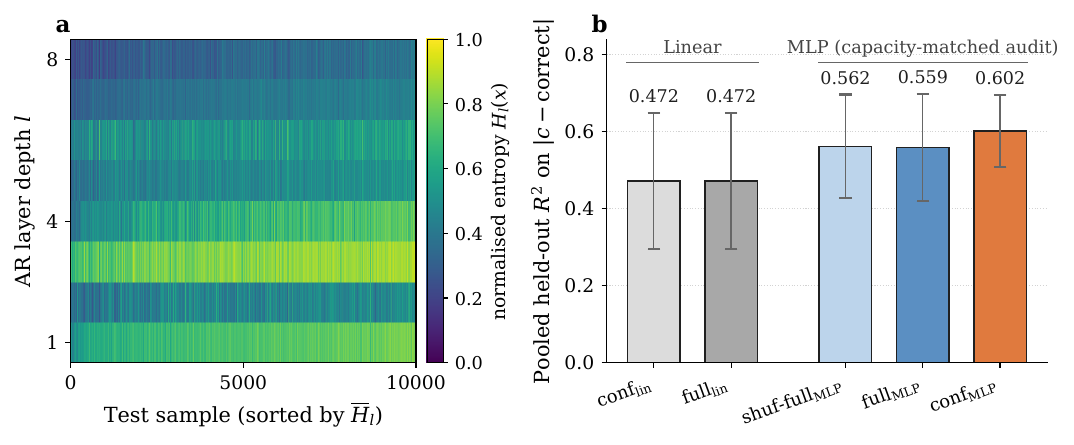}
  \caption{\textbf{Capacity-controlled audit of the full-profile
    probe.}
    \textbf{(a)}~Routing-entropy heatmap (Swin-Tiny $+$ Block-AR,
    CIFAR-10, seed-$1$): deep AR layers collapse to
    near-deterministic routing, early layers retain dispersion.
    \textbf{(b)}~Pooled held-out $R^2$ on
    $|\mathrm{conf}(x){-}\mathrm{correct}(x)|$ across $18$
    seed-level AR cells for five regressors over
    $(c, H_1,\ldots,H_L)$. The full-profile MLP does not isolate
    routing-specific signal once capacity-matched and
    shuffled-routing controls are included
    (App.~\ref{app:probe-sanity}).}
  \label{fig:mechanistic-autopsy}
\end{figure}

\section{Related work}
\label{sec:related}

\textbf{Post-hoc calibration and internal-state uncertainty.}
Standard post-hoc methods (cited in \S\ref{sec:intro} and
\S\ref{sec:experiments}; full set in
Tab.~\ref{tab:calibration_block_ar}) share one structural property:
each is a function of the classifier's softmax or logits alone, with
no access to any per-sample internal computation. A separate line
of work uses internal representations rather than only the
softmax: feature-distance, Bayesian last-layer, and ensemble
approaches~\citep{ovadia2019can}. None specifically tests whether
a \emph{dynamic-routing trajectory} carries calibration-relevant
information beyond the softmax: the question we ask here.

\textbf{Multi-metric and subgroup calibration.}
\citet{nixon2019measuring} and \citet{roelofs2022mitigating}
recommend AdaECE / classwise ECE / MCE / CIs alongside ECE, and
\citet{blasiok2023smooth} add SmoothECE; we report all of these.
Multicalibration~\citep{hebert2018multicalibration} and
verified uncertainty
calibration~\citep{kumar2019verified} require calibration within
identifiable subgroups. Our worst-tertile ECE is not a replacement for those benchmarks; it
is an architecture-induced subgroup stress test, with subgroups
defined by routing-derived state. Calibration-aware
training~\citep{mukhoti2020calibrating,thulasidasan2019mixup} is
orthogonal; this paper is strictly post-hoc.

\textbf{Vision-transformer and AR-architecture calibration.}
\citet{minderer2021revisiting} report modern transformers are
typically better calibrated than ResNets yet still benefit from
post-hoc methods; \citet{hendrycks2019benchmarking} and
\citet{ovadia2019can} measure calibration under shift. Attention
Residuals~\citep{team2026attention} replace the fixed residual skip
with a learned depth-wise soft mixture (block variant capped at
block boundaries); we use the authors' implementation verbatim and
propose no architectural change. We frame this paper as a
diagnostic negative result on a specific assumption: that scalar
routing summaries support routing-aware calibration. Our focus is
the \emph{conditional} structure of in-distribution miscalibration,
not corruption behaviour.

\section{Discussion and limitations}
\label{sec:discussion}

\looseness=-1
The three protocols (\S\S\ref{sec:rq2}--\ref{sec:probe})
converge: no scalar projection is seed-stable, AR-CondCal does not
reliably improve worst-tertile ECE, and full-vector MLP probes
fail capacity-matched and shuffled-routing controls. Conclusions are bounded to AR routing and the evaluated probes;
no safety, OOD, robustness, or universal-absence claim is made. The protocol is
architecture-agnostic by design; cross-architecture validation
and richer (higher-dimensional or non-scalar) descriptors of
internal routing dynamics are open work.
\label{sec:conclusion}



\newpage
\begingroup\small
\bibliographystyle{plainnat}
\bibliography{refs}
\endgroup

\newpage
\appendix

\section{Worst-tertile ECE: bin-level construction}
\label{app:wt-ece-illustration}

\begin{figure}[h]
  \centering
  \includegraphics[width=\linewidth,trim={0 140 0 100},clip]{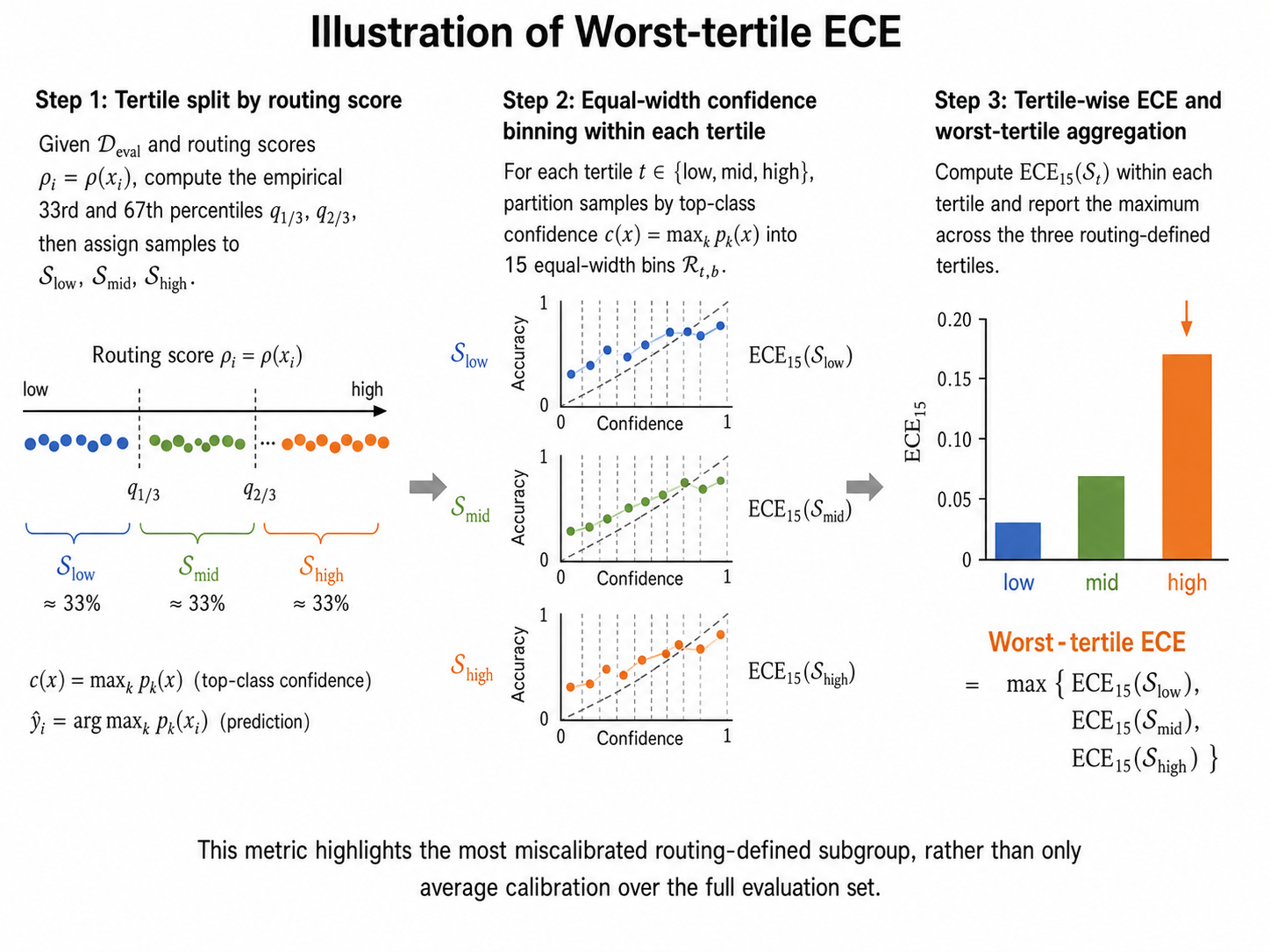}
  \caption{Visual companion to Steps~1--3, in the same notation:
    split $\mathcal{D}_{\mathrm{eval}}$ by routing score $\rho_i$
    into $\mathcal{S}_{\text{low}}, \mathcal{S}_{\text{mid}},
    \mathcal{S}_{\text{high}}$; equal-width confidence binning
    within each tertile yields the bin sets $\mathcal{R}_{t,b}$;
    worst-tertile ECE is the maximum of
    $\mathrm{ECE}_{15}(\mathcal{S}_{\text{low}})$,
    $\mathrm{ECE}_{15}(\mathcal{S}_{\text{mid}})$,
    $\mathrm{ECE}_{15}(\mathcal{S}_{\text{high}})$. This metric
    highlights the most miscalibrated routing-defined subgroup,
    rather than only the average calibration over the full
    evaluation set.}
  \label{fig:wt-ece-illustration}
\end{figure}

Worst-tertile ECE is the diagnostic metric introduced in
\S\ref{sec:method-arcondcal} and used as the routing-conditional
headline score throughout the paper. The body definition (Eqs.~(1)
and~(2) in \S\ref{sec:method-arcondcal}) reports the final
aggregation; this appendix unpacks the within-tertile
$\mathrm{ECE}_{15}(\mathcal{S}_{t})$ operator into its bin-level
components, in the same notation as the body.
Fig.~\ref{fig:wt-ece-illustration} accompanies the three steps below
as a visual companion.

\textbf{Step 1: tertile split by routing score.}
Given the evaluation set
$\mathcal{D}_{\mathrm{eval}} = \{(x_i, y_i)\}_{i=1}^{n}$ and a
per-sample routing feature $\rho_i = \rho(x_i)$, compute the
empirical $33$rd and $67$th percentiles $q_{1/3}, q_{2/3}$ of
$\{\rho_i\}_{i=1}^{n}$ and assign each sample to one of the index
sets $\mathcal{S}_{\text{low}}, \mathcal{S}_{\text{mid}},
\mathcal{S}_{\text{high}}$ as in Eq.~(1) of
\S\ref{sec:method-arcondcal}. By construction each tertile contains
approximately $n/3$ samples (ties at the cut-points are assigned by
strict inequality, matching the body).

\textbf{Step 2: equal-width confidence binning within each tertile.}
For tertile $t \in \{\text{low}, \text{mid}, \text{high}\}$ and
$b \in \{1, \dots, 15\}$, define the $b$-th equal-width confidence
bin restricted to that tertile,
\begin{equation}
\mathcal{R}_{t,b} \;=\; \bigl\{\, i \in \mathcal{S}_{t} \,:\,
   c(x_i) \in \bigl[\tfrac{b-1}{15},\; \tfrac{b}{15}\bigr) \,\bigr\},
\end{equation}
where $c(x) = \max_k p_k(x)$ is the top-class confidence used
throughout. The within-bin empirical accuracy and mean confidence are
\begin{equation}
\mathrm{acc}(\mathcal{R}_{t,b})
   \;=\; \frac{1}{|\mathcal{R}_{t,b}|}
        \sum_{i \in \mathcal{R}_{t,b}} \mathbf{1}[\hat y_i = y_i],
\qquad
\mathrm{conf}(\mathcal{R}_{t,b})
   \;=\; \frac{1}{|\mathcal{R}_{t,b}|}
        \sum_{i \in \mathcal{R}_{t,b}} c(x_i),
\end{equation}
with $\hat y_i = \arg\max_k p_k(x_i)$.

\textbf{Step 3: tertile-wise ECE and worst-tertile aggregation.}
The within-tertile equal-width $\mathrm{ECE}_{15}$ operator that
appears in the body definition expands as
\begin{equation}
\mathrm{ECE}_{15}(\mathcal{S}_{t})
   \;=\; \sum_{b=1}^{15}
        \frac{|\mathcal{R}_{t,b}|}{|\mathcal{S}_{t}|}\;
        \bigl|\,\mathrm{acc}(\mathcal{R}_{t,b})
              - \mathrm{conf}(\mathcal{R}_{t,b})\,\bigr|,
\end{equation}
i.e.\ the standard $15$-bin equal-width
ECE~\citep{guo2017calibration,naeini2015obtaining} computed on the
sub-sample indexed by $\mathcal{S}_{t}$. Worst-tertile ECE is then
the maximum of these three quantities (Eq.~(2) of
\S\ref{sec:method-arcondcal}); it reports the calibration error of
the most miscalibrated routing-defined subgroup rather than the
global average over $\mathcal{D}_{\mathrm{eval}}$, and is therefore
the natural stress test for the routing-conditional miscalibration
hypothesis investigated in the body.

\section{Reliability diagrams on Block Attention Residual}
\label{app:reliability}

\begin{figure}[h]
  \centering
  \includegraphics[width=\linewidth]{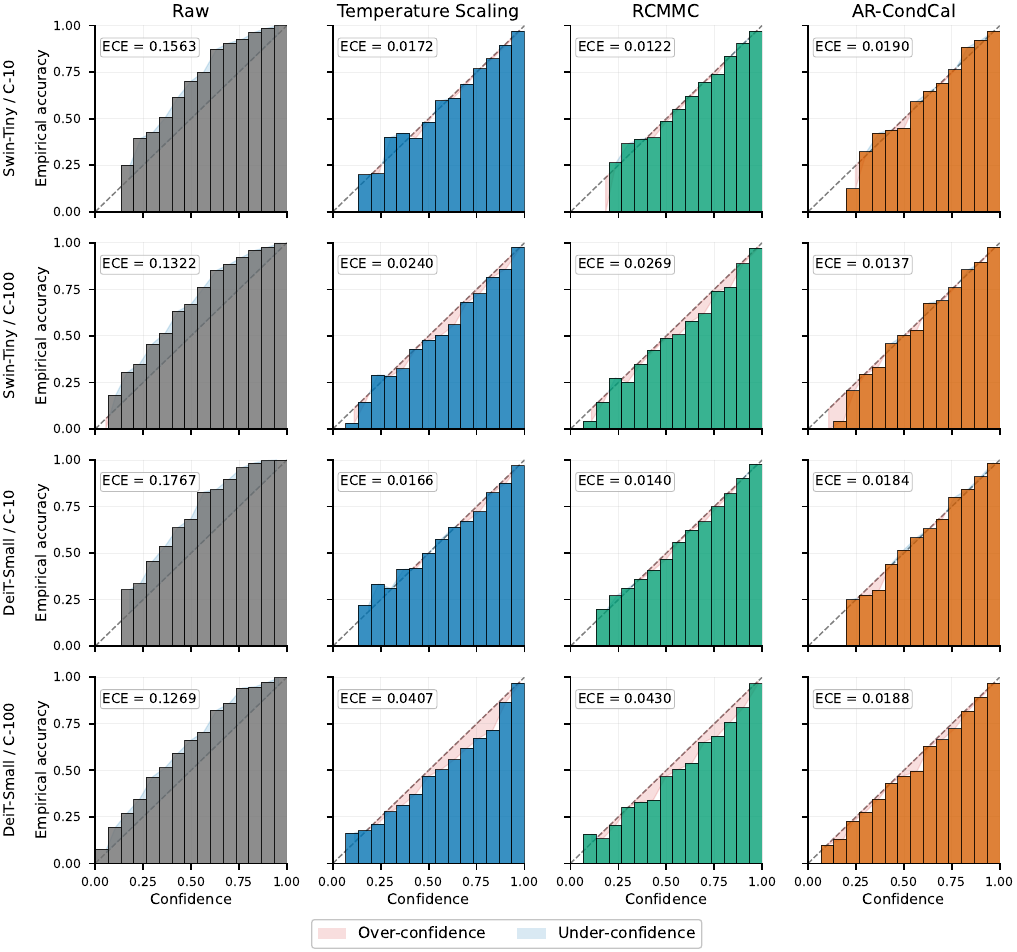}
  \caption{Reliability diagrams on Block-AR for four substrates (rows:
    Swin-Tiny / DeiT-Small $\times$ CIFAR-10 / CIFAR-100) under four
    calibrators (columns). Bars show empirical accuracy in each of $15$
    equal-width confidence bins; dashed line is perfect calibration; red
    shading marks over-confidence (predicted $>$ accuracy) and blue
    marks under-confidence, both relative to the bin's mean confidence.
    Per-panel global ECE is annotated in the upper-left corner. All
    cells use the seed-$0$ ep\,$=$\,$299$ checkpoint with the 50/50
    cal/test split.}
  \label{fig:reliability-block}
\end{figure}

Figure~\ref{fig:reliability-block} shows reliability diagrams ($15$
equal-width confidence bins, seed $0$, ep\,$=$\,$299$, 50/50 split) for
the four Block-AR cells (rows) and four calibration settings (columns):
uncalibrated, Temperature Scaling, RCMMC, and AR-CondCal. Each panel
is annotated with its own global ECE; per-bar height is empirical
accuracy at that confidence bin. The qualitative picture is consistent
across substrates: all three calibrators substantially flatten the raw
over-confidence gap, AR-CondCal sits within the TS/RCMMC band on global
ECE on every cell, and bar-level residuals in the high-confidence tail
persist for \emph{every} method, matching the worst-tertile-stress-test pattern
that the main body identifies as the open subgroup-calibration problem.

\section{Calibration-set-size sensitivity}
\label{app:calsize}

\begin{figure}[h]
  \centering
  \includegraphics[width=0.75\linewidth]{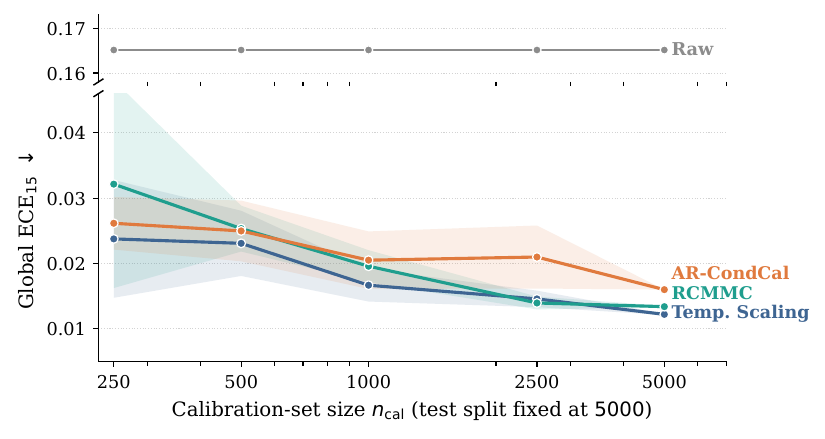}
  \caption{Calibration-set-size sensitivity (Block-AR, Swin-Tiny,
    CIFAR-10). Mean $\pm$ 1-std across 5 random calibration subsamples
    per size; test split fixed at 5000 samples. TS is the most
    sample-efficient; AR-CondCal degrades more gracefully than RCMMC at
    small $n_\text{cal}$, but neither 2-D calibrator catches TS at any
    tested size.}
  \label{fig:calsize}
\end{figure}

We re-fit each calibrator on sub-samples of the 5000-sample calibration
split, holding the 5000-sample test split fixed. Each size is repeated
5 times with different random calibration subsets. Figure
\ref{fig:calsize} reports mean global ECE with $\pm$1-std shaded bands.

Two effects are relevant to the paper's conservative framing.
\textbf{(a)} Temperature Scaling is the most sample-efficient method
on this substrate: its ECE rises only from 0.0122 at $n_\text{cal}=5000$ to 0.0238 at
$n_\text{cal}=250$. This is consistent with the intuition that a
single scalar calibrator parameter needs very little calibration
data.
\textbf{(b)} RCMMC is the \emph{least} sample-robust of the 2-D
calibrators; its ECE degrades by a factor of $\approx 2.4$ over the same
range ($0.0134 \to 0.0322$). AR-CondCal degrades more gracefully
(factor $\approx 1.6$, $0.0160 \to 0.0262$). At
$n_\text{cal} \le 1000$, AR-CondCal and RCMMC are statistically
indistinguishable within the $\pm$1-std band. This yields a
conservative empirical finding: relative to RCMMC, AR-CondCal is
less sample-sensitive at small calibration sizes; this comparison
does not imply a reliable routing-feature gain over the matched
confidence-only controls.

\section{MCE, Classwise ECE, and SmoothECE (complement to Table~\ref{tab:calibration_block_ar})}
\label{app:cwece}

Table~\ref{tab:cwece-supp} reports the three diagnostic metrics we
chose not to place in the main comparison: MCE (worst-bin
$|\mathrm{acc}-\mathrm{conf}|$ over $15$ equal-width bins), classwise
ECE~\citep{kull2019beyond} (mean of per-class ECE over the $10$
CIFAR-10 classes), and SmoothECE~\citep{blasiok2023smooth}. We include SmoothECE in particular
as a binning-robust cross-check: the equal-width-ECE column of the
main table and the SmoothECE column here rank the calibrators
consistently (up to the Vector-Scaling row, which is the only method
that substantially improves per-class calibration beyond what its
global ECE suggests), so no single binning choice is carrying the
ordering in Tab.~\ref{tab:calibration_block_ar}. None of these
appendix metrics changes the paper's main-body interpretation:
AR-CondCal continues not to dominate TS / ETS / RCMMC globally, nor
does it uniformly improve the worst-tertile metric.

%
\begin{table}[!ht]
\centering
\caption{Appendix diagnostic metrics on Block-AR, complement to Tab.~\ref{tab:calibration_block_ar}. MCE is the worst-bin $|\mathrm{acc}-\mathrm{conf}|$ among $15$ equal-width confidence bins (bins with $<5$ samples ignored); \emph{MCE is included only as a stress-test metric --- the worst-bin statistic is highly sensitive to tail-bin support and its cross-seed std (e.g., $\pm 0.37$ for AR-CondCal) reflects this instability rather than a per-method calibration property}. Classwise ECE is the mean of per-class ECE over the $10$ CIFAR-10 classes. SmoothECE is the kernel-smoothed calibration error of Bla\c{s}iok \& Nakkiran (2024), included as a robustness check on the equal-width-ECE ordering. Bold = best mean; \underline{underlined} = $2$nd-best mean (raw row excluded from ranking). No single metric should be read as the whole calibration picture; we report this triad so readers can cross-check the main table against a binning-robust alternative.}
\label{tab:cwece-supp}
\setlength{\tabcolsep}{4pt}
\begin{tabular}{lccc}
\toprule
Method & MCE $\downarrow$ & Classwise ECE $\downarrow$ & SmoothECE $\downarrow$ \\
\midrule
No calibration & $0.2237\,{\scriptstyle\pm 0.0465}$ & $0.0371\,{\scriptstyle\pm 0.0052}$ & $0.1650\,{\scriptstyle\pm 0.0196}$ \\
Temp. Scaling (TS) & $0.1898\,{\scriptstyle\pm 0.0078}$ & $0.0132\,{\scriptstyle\pm 0.0045}$ & $0.0156\,{\scriptstyle\pm 0.0006}$ \\
Ensemble TS & $0.1898\,{\scriptstyle\pm 0.0079}$ & $0.0132\,{\scriptstyle\pm 0.0045}$ & $0.0156\,{\scriptstyle\pm 0.0006}$ \\
Vector Scaling & $0.0744\,{\scriptstyle\pm 0.0251}$ & $\mathbf{0.0086\,{\scriptstyle\pm 0.0023}}$ & $0.0189\,{\scriptstyle\pm 0.0034}$ \\
Classwise TS & $0.0701\,{\scriptstyle\pm 0.0792}$ & $\underline{0.0122\,{\scriptstyle\pm 0.0041}}$ & $\underline{0.0155\,{\scriptstyle\pm 0.0012}}$ \\
Parametric TS & $0.0880\,{\scriptstyle\pm 0.1423}$ & $0.0126\,{\scriptstyle\pm 0.0043}$ & $\mathbf{0.0152\,{\scriptstyle\pm 0.0019}}$ \\
Histogram Binning & $0.0931\,{\scriptstyle\pm 0.0930}$ & $0.0180\,{\scriptstyle\pm 0.0068}$ & $0.0225\,{\scriptstyle\pm 0.0083}$ \\
BBQ & $0.0657\,{\scriptstyle\pm 0.0219}$ & $0.0179\,{\scriptstyle\pm 0.0068}$ & $0.0199\,{\scriptstyle\pm 0.0081}$ \\
Isotonic Regression & $0.1080\,{\scriptstyle\pm 0.2306}$ & $0.0176\,{\scriptstyle\pm 0.0064}$ & $0.0214\,{\scriptstyle\pm 0.0055}$ \\
SB-ECE TS & $0.0787\,{\scriptstyle\pm 0.0891}$ & $0.0135\,{\scriptstyle\pm 0.0046}$ & $0.0180\,{\scriptstyle\pm 0.0022}$ \\
RCMMC & $0.0715\,{\scriptstyle\pm 0.0737}$ & $0.0126\,{\scriptstyle\pm 0.0043}$ & $0.0173\,{\scriptstyle\pm 0.0031}$ \\
\midrule
NW (Conf only) & $0.0820\,{\scriptstyle\pm 0.1079}$ & $0.0134\,{\scriptstyle\pm 0.0046}$ & $0.0195\,{\scriptstyle\pm 0.0050}$ \\
NW (Conf $+$ Pred. Ent.) & $\underline{0.0524\,{\scriptstyle\pm 0.1548}}$ & $0.0133\,{\scriptstyle\pm 0.0044}$ & $0.0204\,{\scriptstyle\pm 0.0056}$ \\
\textbf{NW (Conf $+$ Routing): AR-CondCal} & $\mathbf{0.0412\,{\scriptstyle\pm 0.3702}}$ & $0.0134\,{\scriptstyle\pm 0.0045}$ & $0.0175\,{\scriptstyle\pm 0.0040}$ \\
\bottomrule
\end{tabular}
\end{table}

\section{Paired NLL and Brier deltas vs uncalibrated model}
\label{app:paired-delta}

Tab.~\ref{tab:calibration_block_ar} reports NLL and Brier as
seed means rather than mean\,$\pm$\,std because the pooled
cross-seed standard deviations of these proper scores are
dominated by raw-model accuracy/NLL variance rather than by
calibrator-induced effects (raw $\sigma_{\mathrm{NLL}} \!\approx\!
0.33$ for the $3$ training seeds of the c10-swin Block-AR cell).
A paired view (subtracting the per-seed raw NLL/Brier from the
calibrated NLL/Brier within the same seed and only then summarising
across seeds) removes that nuisance variance and isolates the
calibrator-induced effect.
Tab.~\ref{tab:paired-delta} reports the paired
$\Delta\mathrm{NLL}_{m,s} = \mathrm{NLL}_{m,s} -
\mathrm{NLL}_{\mathrm{raw},s}$ and $\Delta\mathrm{Brier}_{m,s}$,
aggregated as mean\,$\pm$\,$1$ std over the $3$ seeds. The
$\Delta$-stds collapse to the $0.001$--$0.015$ band, two orders of
magnitude smaller than the raw-NLL pooled std, and the
calibrator-induced effects become directly comparable across
methods. Vector Scaling delivers the strongest paired improvement
on both proper scores, followed by RCMMC and Classwise TS;
AR-CondCal achieves a $\Delta\mathrm{NLL} = -0.106$ (i.e., a
$0.011$ smaller NLL improvement than VS) with one of the smallest
$\Delta$-stds in the table ($\pm 0.0018$), consistent with its
stable paired proper-score effect on this metric.

\begin{table}[!ht]
\centering
\setlength{\tabcolsep}{6pt}
\small
\caption{\textbf{Paired proper-score changes relative to the
uncalibrated model} (Block-AR, Swin-Tiny, CIFAR-10, ep $=$ $299$;
complement to Tab.~\ref{tab:calibration_block_ar}). For each
training seed $s$, we compute $\Delta\mathrm{NLL}_{m,s} =
\mathrm{NLL}_{m,s} - \mathrm{NLL}_{\mathrm{raw},s}$ and analogously
for Brier, then report mean\,$\pm$\,$1$ std over seeds. This paired
view removes the dominant cross-seed raw-model variance and
isolates the calibrator-induced effect (raw $\sigma_{\mathrm{NLL}}
\!\approx\! 0.33$ collapses to $\Delta$-stds in the $0.001$--$0.015$
band). \textbf{Bold} = most-negative (best) mean; \underline{underlined}
= $2$nd-best.}
\label{tab:paired-delta}
\begin{tabular}{lcc}
\toprule
Method & $\Delta$NLL vs raw $\downarrow$ & $\Delta$Brier vs raw $\downarrow$ \\
\midrule
Temp. Scaling (TS) & $-0.1154\,{\scriptstyle\pm 0.0042}$ & $-0.0262\,{\scriptstyle\pm 0.0082}$ \\
Ensemble TS & $-0.1153\,{\scriptstyle\pm 0.0042}$ & $-0.0262\,{\scriptstyle\pm 0.0082}$ \\
Vector Scaling & $\mathbf{-0.1228\,{\scriptstyle\pm 0.0130}}$ & $\mathbf{-0.0284\,{\scriptstyle\pm 0.0104}}$ \\
Classwise TS & $-0.1162\,{\scriptstyle\pm 0.0061}$ & $-0.0266\,{\scriptstyle\pm 0.0090}$ \\
Parametric TS & $-0.1087\,{\scriptstyle\pm 0.0101}$ & $-0.0253\,{\scriptstyle\pm 0.0087}$ \\
Histogram Binning & $-0.0862\,{\scriptstyle\pm 0.0096}$ & $-0.0222\,{\scriptstyle\pm 0.0055}$ \\
BBQ & $-0.0835\,{\scriptstyle\pm 0.0048}$ & $-0.0217\,{\scriptstyle\pm 0.0065}$ \\
Isotonic Regression & $-0.0823\,{\scriptstyle\pm 0.0154}$ & $-0.0222\,{\scriptstyle\pm 0.0058}$ \\
SB-ECE TS & $-0.1133\,{\scriptstyle\pm 0.0041}$ & $-0.0263\,{\scriptstyle\pm 0.0079}$ \\
RCMMC & $\underline{-0.1166\,{\scriptstyle\pm 0.0061}}$ & $\underline{-0.0267\,{\scriptstyle\pm 0.0089}}$ \\
Conf-only & $-0.1107\,{\scriptstyle\pm 0.0028}$ & $-0.0255\,{\scriptstyle\pm 0.0076}$ \\
Conf $+$ PredEntropy & $-0.1107\,{\scriptstyle\pm 0.0023}$ & $-0.0258\,{\scriptstyle\pm 0.0072}$ \\
\textbf{AR-CondCal} & $-0.1055\,{\scriptstyle\pm 0.0018}$ & $-0.0247\,{\scriptstyle\pm 0.0079}$ \\
\bottomrule
\end{tabular}
\end{table}

\section{Cross-substrate calibration sweep}
\label{app:cross-cell-sweep}

This appendix replicates the structure of
Tab.~\ref{tab:calibration_block_ar} across the seven additional
$3$-seed AR cells we evaluated: Block-AR and Full-AR variants of
Swin-Tiny and DeiT-Small on CIFAR-10 and CIFAR-100 (plus the
c10-swin Block-AR anchor for orientation against the main table).
For compactness we report only the four calibrators most relevant
to the paper's central claim, namely TS as the seed-stable parametric anchor in Tab.~\ref{tab:calibration_block_ar} and AR-CondCal on the worst-tertile ECE side.

The pattern of \S\ref{sec:rq3} replicates substrate-by-substrate:
AR-CondCal sits within the cross-method seed-variance band on
global ECE / AdaECE on every cell, and on worst-tertile ECE stays
within roughly $1$ cross-method std-sum of the leader pack but
never separates from the strongest baseline by more than its own
seed-std. The qualitative \emph{ranking-stability} claim that method ordering once seeds are pooled is robust to the per-seed std band holds across all displayed cells.

\begin{figure}[!ht]
  \centering
  \includegraphics[width=\linewidth]{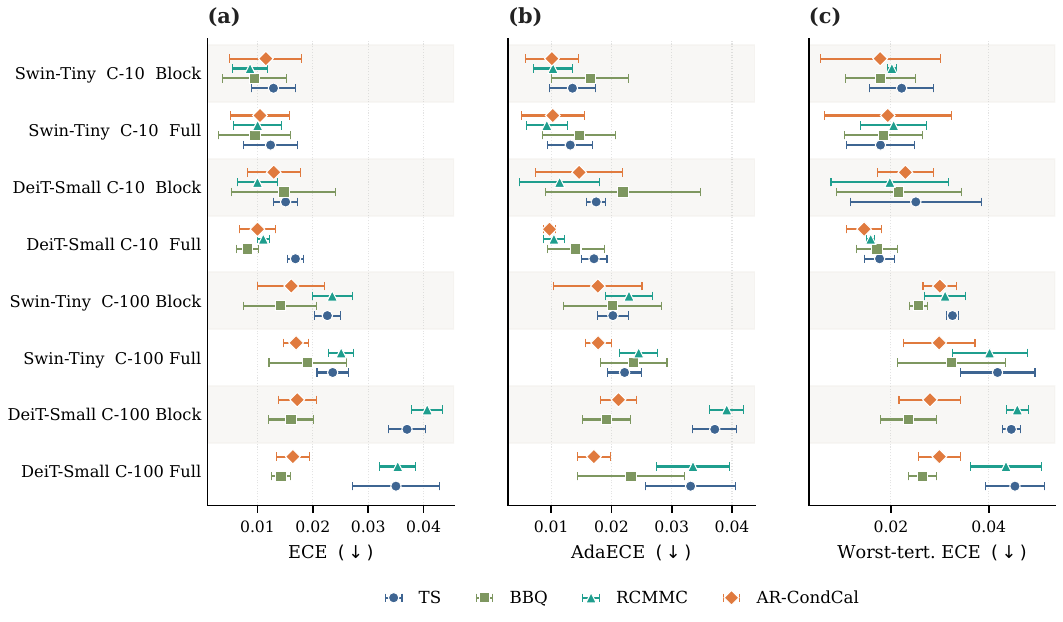}
  \caption{\textbf{Cross-cell calibration sweep --- main metrics.}
  Mean over $3$ training seeds (markers) with horizontal $\pm 1$ std
  bars, on a $5000$-sample test half (cal/test split seed $42$). Eight
  $($backbone, dataset, variant$)$ AR cells (rows) compared on three
  headline calibration metrics (panels). Four calibrators per cell:
  TS (\(\bullet\)), BBQ (\(\blacksquare\)), RCMMC (\(\blacktriangle\)),
  and \textbf{AR-CondCal (\(\blacklozenge\), highlighted)}. Among the
  displayed rows, AR-CondCal often falls near the leader pack on
  worst-tertile ECE but no advantage separates from seed variance,
  the \emph{ranking-stability} claim of \S\ref{sec:rq3}.}
  \label{fig:cross-cell-main}
\end{figure}

\begin{figure}[!ht]
  \centering
  \includegraphics[width=\linewidth]{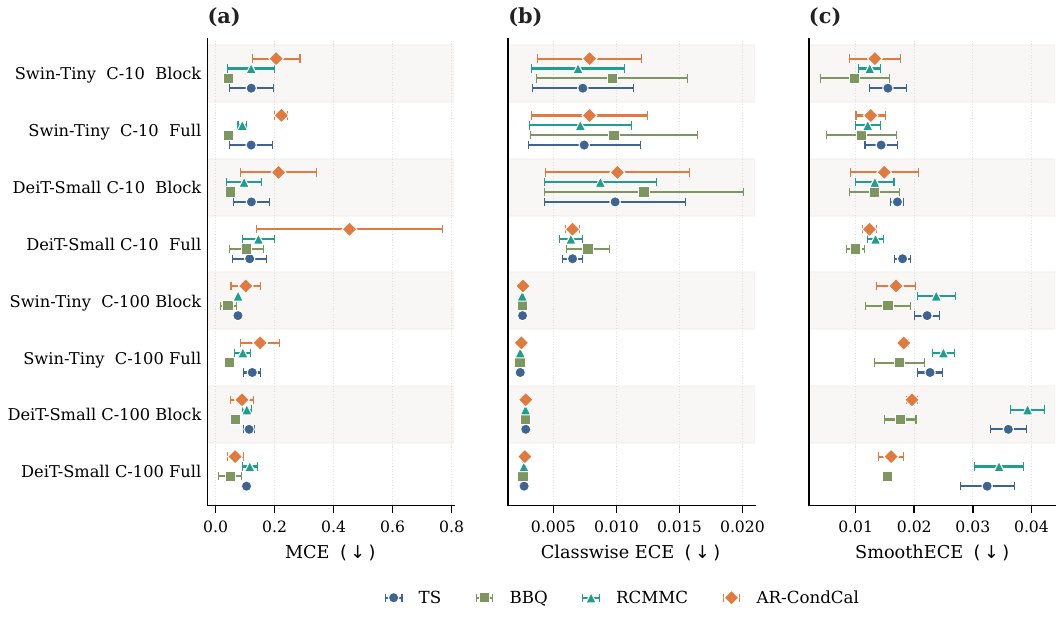}
  \caption{\textbf{Cross-cell calibration sweep --- supplementary
  metrics (complement to Fig.~\ref{fig:cross-cell-main}).} Same eight
  cells, four calibrators, and protocol as
  Fig.~\ref{fig:cross-cell-main}; panels report MCE (worst-bin
  $|\mathrm{acc}-\mathrm{conf}|$ among $15$ equal-width confidence
  bins), Classwise ECE (mean of per-class ECE), and the
  binning-robust SmoothECE~\citep{blasiok2023smooth}.}
  \label{fig:cross-cell-supp}
\end{figure}

\paragraph{Supplementary metrics across the same cells.}
The same conclusion is reinforced by Fig.~\ref{fig:cross-cell-supp}:
on the binning-robust SmoothECE panel, BBQ wins on every displayed
cell and AR-CondCal is consistently the runner-up; on the MCE
panel, BBQ again dominates and the AR-CondCal MCE advantage
observed on the single c10-swin Block-AR cell of
Tab.~\ref{tab:cwece-supp} does not generalise.

\section{Per-cell full-method calibration tables}
\label{app:per-cell-cal}

For completeness, this appendix reports the full $14$-method
calibration comparison from Tab.~\ref{tab:calibration_block_ar}
(ECE, AdaECE, NLL, Brier, worst-tertile ECE; $3$-seed mean $\pm$
$1$ std) replicated on the seven additional 3-seed AR cells of our
sweep: c10-Swin Full-AR, c10-DeiT Block / Full-AR, c100-Swin Block /
Full-AR, and c100-DeiT Block / Full-AR. These per-cell tables are
the source of the consolidated cross-cell figures
(Fig.~\ref{fig:cross-cell-main}, Fig.~\ref{fig:cross-cell-supp}).

{\setlength{\tabcolsep}{3pt}\small
\setlength{\LTcapwidth}{\textwidth}
\begin{longtable}{p{2.8cm}ccccc}
\caption{Per-cell full-method calibration sweep (seven $3$-seed AR cells;
$5000$-sample test half, seed $42$ split; $3$-seed mean\,$\pm$\,$1$ std).
Same $14$-method protocol and ordering as Tab.~\ref{tab:calibration_block_ar};
\textbf{Bold} = best mean within a cell$\cdot$column block,
\underline{underline} = $2$nd-best (raw row excluded from ranking).
Cell sub-headers separate the seven post-Tab.~\ref{tab:calibration_block_ar} cells.}
\label{tab:cal-per-cell-combined} \\
\toprule
Method & ECE $\downarrow$ & AdaECE $\downarrow$ & NLL $\downarrow$ & Brier $\downarrow$ & Worst-tert.\ ECE $\downarrow$ \\
\midrule
\endfirsthead
\multicolumn{6}{l}{\textit{Tab.~\ref{tab:cal-per-cell-combined} continued from previous page.}} \\
\toprule
Method & ECE $\downarrow$ & AdaECE $\downarrow$ & NLL $\downarrow$ & Brier $\downarrow$ & Worst-tert.\ ECE $\downarrow$ \\
\midrule
\endhead
\midrule
\multicolumn{6}{r}{\textit{continued on next page}} \\
\endfoot
\bottomrule
\endlastfoot

\multicolumn{6}{l}{\textbf{Cell 1.} Full-AR --- Swin-Tiny / CIFAR-10}\\
\midrule
No calibration & $0.1423\,{\scriptstyle\pm 0.0138}$ & $0.1422\,{\scriptstyle\pm 0.0138}$ & $0.515$ & $0.219$ & $0.1536\,{\scriptstyle\pm 0.0093}$ \\
Temp. Scaling (TS) & $0.0124\,{\scriptstyle\pm 0.0049}$ & $0.0132\,{\scriptstyle\pm 0.0037}$ & $0.397$ & $0.192$ & $0.0179\,{\scriptstyle\pm 0.0070}$ \\
Ensemble TS & $0.0124\,{\scriptstyle\pm 0.0049}$ & $0.0132\,{\scriptstyle\pm 0.0037}$ & $0.397$ & $0.192$ & $\underline{0.0179\,{\scriptstyle\pm 0.0070}}$ \\
Vector Scaling & $0.0109\,{\scriptstyle\pm 0.0040}$ & $0.0119\,{\scriptstyle\pm 0.0059}$ & $\mathbf{0.391}$ & $\mathbf{0.190}$ & $0.0186\,{\scriptstyle\pm 0.0061}$ \\
Classwise TS & $0.0109\,{\scriptstyle\pm 0.0037}$ & $\underline{0.0099\,{\scriptstyle\pm 0.0028}}$ & $0.397$ & $\underline{0.192}$ & $0.0191\,{\scriptstyle\pm 0.0060}$ \\
Parametric TS & $0.0126\,{\scriptstyle\pm 0.0065}$ & $0.0116\,{\scriptstyle\pm 0.0083}$ & $0.400$ & $0.193$ & $\mathbf{0.0171\,{\scriptstyle\pm 0.0051}}$ \\
Histogram Binning & $0.0122\,{\scriptstyle\pm 0.0052}$ & $0.0144\,{\scriptstyle\pm 0.0065}$ & $0.428$ & $0.197$ & $0.0206\,{\scriptstyle\pm 0.0066}$ \\
BBQ & $\mathbf{0.0095\,{\scriptstyle\pm 0.0065}}$ & $0.0147\,{\scriptstyle\pm 0.0061}$ & $0.431$ & $0.197$ & $0.0185\,{\scriptstyle\pm 0.0079}$ \\
Isotonic Regression & $0.0137\,{\scriptstyle\pm 0.0070}$ & $0.0112\,{\scriptstyle\pm 0.0064}$ & $0.437$ & $0.197$ & $0.0209\,{\scriptstyle\pm 0.0069}$ \\
SB-ECE TS & $0.0135\,{\scriptstyle\pm 0.0021}$ & $0.0137\,{\scriptstyle\pm 0.0056}$ & $0.399$ & $0.192$ & $0.0198\,{\scriptstyle\pm 0.0068}$ \\
RCMMC & $\underline{0.0100\,{\scriptstyle\pm 0.0044}}$ & $\mathbf{0.0093\,{\scriptstyle\pm 0.0034}}$ & $\underline{0.397}$ & $0.192$ & $0.0205\,{\scriptstyle\pm 0.0068}$ \\
Conf-only & $0.0117\,{\scriptstyle\pm 0.0047}$ & $0.0114\,{\scriptstyle\pm 0.0048}$ & $0.402$ & $0.193$ & $0.0201\,{\scriptstyle\pm 0.0075}$ \\
Conf + PredEntropy & $0.0115\,{\scriptstyle\pm 0.0079}$ & $0.0105\,{\scriptstyle\pm 0.0040}$ & $0.406$ & $0.194$ & $0.0207\,{\scriptstyle\pm 0.0054}$ \\
\textbf{AR-CondCal} & $0.0105\,{\scriptstyle\pm 0.0053}$ & $0.0103\,{\scriptstyle\pm 0.0052}$ & $0.407$ & $0.194$ & $0.0193\,{\scriptstyle\pm 0.0130}$ \\
\midrule
\multicolumn{6}{l}{\textbf{Cell 2.} Block-AR --- DeiT-Small / CIFAR-10}\\
\midrule
No calibration & $0.1729\,{\scriptstyle\pm 0.0033}$ & $0.1728\,{\scriptstyle\pm 0.0034}$ & $0.640$ & $0.269$ & $0.1783\,{\scriptstyle\pm 0.0025}$ \\
Temp. Scaling (TS) & $0.0151\,{\scriptstyle\pm 0.0021}$ & $0.0175\,{\scriptstyle\pm 0.0015}$ & $0.496$ & $0.233$ & $0.0251\,{\scriptstyle\pm 0.0134}$ \\
Ensemble TS & $0.0151\,{\scriptstyle\pm 0.0021}$ & $0.0175\,{\scriptstyle\pm 0.0015}$ & $0.496$ & $0.233$ & $0.0251\,{\scriptstyle\pm 0.0134}$ \\
Vector Scaling & $0.0145\,{\scriptstyle\pm 0.0037}$ & $0.0151\,{\scriptstyle\pm 0.0026}$ & $\mathbf{0.483}$ & $\mathbf{0.229}$ & $\mathbf{0.0176\,{\scriptstyle\pm 0.0048}}$ \\
Classwise TS & $\underline{0.0123\,{\scriptstyle\pm 0.0018}}$ & $0.0151\,{\scriptstyle\pm 0.0015}$ & $0.494$ & $0.232$ & $\underline{0.0177\,{\scriptstyle\pm 0.0075}}$ \\
Parametric TS & $0.0130\,{\scriptstyle\pm 0.0051}$ & $\underline{0.0125\,{\scriptstyle\pm 0.0083}}$ & $0.498$ & $0.234$ & $0.0240\,{\scriptstyle\pm 0.0110}$ \\
Histogram Binning & $0.0138\,{\scriptstyle\pm 0.0068}$ & $0.0143\,{\scriptstyle\pm 0.0028}$ & $0.531$ & $0.238$ & $0.0210\,{\scriptstyle\pm 0.0080}$ \\
BBQ & $0.0148\,{\scriptstyle\pm 0.0094}$ & $0.0219\,{\scriptstyle\pm 0.0129}$ & $0.539$ & $0.239$ & $0.0216\,{\scriptstyle\pm 0.0128}$ \\
Isotonic Regression & $0.0152\,{\scriptstyle\pm 0.0067}$ & $0.0158\,{\scriptstyle\pm 0.0060}$ & $0.538$ & $0.238$ & $0.0206\,{\scriptstyle\pm 0.0064}$ \\
SB-ECE TS & $0.0173\,{\scriptstyle\pm 0.0012}$ & $0.0171\,{\scriptstyle\pm 0.0002}$ & $0.499$ & $0.233$ & $0.0265\,{\scriptstyle\pm 0.0109}$ \\
RCMMC & $\mathbf{0.0100\,{\scriptstyle\pm 0.0036}}$ & $\mathbf{0.0114\,{\scriptstyle\pm 0.0066}}$ & $\underline{0.493}$ & $\underline{0.231}$ & $0.0198\,{\scriptstyle\pm 0.0120}$ \\
Conf-only & $0.0147\,{\scriptstyle\pm 0.0105}$ & $0.0150\,{\scriptstyle\pm 0.0093}$ & $0.504$ & $0.234$ & $0.0224\,{\scriptstyle\pm 0.0126}$ \\
Conf + PredEntropy & $0.0152\,{\scriptstyle\pm 0.0083}$ & $0.0143\,{\scriptstyle\pm 0.0084}$ & $0.505$ & $0.234$ & $0.0211\,{\scriptstyle\pm 0.0097}$ \\
\textbf{AR-CondCal} & $0.0129\,{\scriptstyle\pm 0.0048}$ & $0.0146\,{\scriptstyle\pm 0.0072}$ & $0.513$ & $0.235$ & $0.0229\,{\scriptstyle\pm 0.0057}$ \\
\midrule
\multicolumn{6}{l}{\textbf{Cell 3.} Full-AR --- DeiT-Small / CIFAR-10}\\
\midrule
No calibration & $0.1716\,{\scriptstyle\pm 0.0015}$ & $0.1715\,{\scriptstyle\pm 0.0016}$ & $0.435$ & $0.167$ & $0.1782\,{\scriptstyle\pm 0.0059}$ \\
Temp. Scaling (TS) & $0.0169\,{\scriptstyle\pm 0.0014}$ & $0.0171\,{\scriptstyle\pm 0.0022}$ & $0.283$ & $0.133$ & $0.0177\,{\scriptstyle\pm 0.0031}$ \\
Ensemble TS & $0.0168\,{\scriptstyle\pm 0.0014}$ & $0.0171\,{\scriptstyle\pm 0.0022}$ & $0.283$ & $0.133$ & $0.0177\,{\scriptstyle\pm 0.0031}$ \\
Vector Scaling & $0.0162\,{\scriptstyle\pm 0.0042}$ & $0.0165\,{\scriptstyle\pm 0.0013}$ & $\mathbf{0.280}$ & $\mathbf{0.133}$ & $0.0195\,{\scriptstyle\pm 0.0033}$ \\
Classwise TS & $0.0162\,{\scriptstyle\pm 0.0009}$ & $0.0155\,{\scriptstyle\pm 0.0004}$ & $0.283$ & $0.133$ & $0.0173\,{\scriptstyle\pm 0.0021}$ \\
Parametric TS & $\mathbf{0.0078\,{\scriptstyle\pm 0.0011}}$ & $\underline{0.0092\,{\scriptstyle\pm 0.0021}}$ & $0.282$ & $0.133$ & $\mathbf{0.0140\,{\scriptstyle\pm 0.0014}}$ \\
Histogram Binning & $0.0126\,{\scriptstyle\pm 0.0042}$ & $0.0124\,{\scriptstyle\pm 0.0021}$ & $0.306$ & $0.136$ & $0.0173\,{\scriptstyle\pm 0.0028}$ \\
BBQ & $\underline{0.0082\,{\scriptstyle\pm 0.0019}}$ & $0.0141\,{\scriptstyle\pm 0.0047}$ & $0.313$ & $0.137$ & $0.0171\,{\scriptstyle\pm 0.0042}$ \\
Isotonic Regression & $0.0102\,{\scriptstyle\pm 0.0019}$ & $0.0105\,{\scriptstyle\pm 0.0009}$ & $0.307$ & $0.136$ & $0.0177\,{\scriptstyle\pm 0.0014}$ \\
SB-ECE TS & $0.0187\,{\scriptstyle\pm 0.0008}$ & $0.0160\,{\scriptstyle\pm 0.0010}$ & $0.286$ & $\underline{0.133}$ & $0.0240\,{\scriptstyle\pm 0.0014}$ \\
RCMMC & $0.0111\,{\scriptstyle\pm 0.0011}$ & $0.0105\,{\scriptstyle\pm 0.0018}$ & $\underline{0.282}$ & $0.133$ & $0.0158\,{\scriptstyle\pm 0.0008}$ \\
Conf-only & $0.0088\,{\scriptstyle\pm 0.0020}$ & $0.0096\,{\scriptstyle\pm 0.0006}$ & $0.286$ & $0.134$ & $0.0145\,{\scriptstyle\pm 0.0033}$ \\
Conf + PredEntropy & $0.0098\,{\scriptstyle\pm 0.0013}$ & $\mathbf{0.0087\,{\scriptstyle\pm 0.0018}}$ & $0.286$ & $0.133$ & $0.0168\,{\scriptstyle\pm 0.0021}$ \\
\textbf{AR-CondCal} & $0.0100\,{\scriptstyle\pm 0.0033}$ & $0.0097\,{\scriptstyle\pm 0.0010}$ & $0.297$ & $0.135$ & $\underline{0.0145\,{\scriptstyle\pm 0.0035}}$ \\
\midrule
\multicolumn{6}{l}{\textbf{Cell 4.} Block-AR --- Swin-Tiny / CIFAR-100}\\
\midrule
No calibration & $0.1345\,{\scriptstyle\pm 0.0018}$ & $0.1344\,{\scriptstyle\pm 0.0018}$ & $1.100$ & $0.401$ & $0.1487\,{\scriptstyle\pm 0.0079}$ \\
Temp. Scaling (TS) & $0.0226\,{\scriptstyle\pm 0.0019}$ & $0.0202\,{\scriptstyle\pm 0.0021}$ & $\underline{0.973}$ & $0.376$ & $0.0326\,{\scriptstyle\pm 0.0010}$ \\
Ensemble TS & $0.0226\,{\scriptstyle\pm 0.0019}$ & $0.0202\,{\scriptstyle\pm 0.0021}$ & $0.973$ & $\underline{0.376}$ & $0.0326\,{\scriptstyle\pm 0.0010}$ \\
Vector Scaling & $0.0258\,{\scriptstyle\pm 0.0044}$ & $0.0235\,{\scriptstyle\pm 0.0030}$ & $0.976$ & $0.378$ & $0.0383\,{\scriptstyle\pm 0.0030}$ \\
Classwise TS & $0.0262\,{\scriptstyle\pm 0.0034}$ & $0.0240\,{\scriptstyle\pm 0.0020}$ & $0.977$ & $0.376$ & $0.0343\,{\scriptstyle\pm 0.0034}$ \\
Parametric TS & $0.0189\,{\scriptstyle\pm 0.0037}$ & $0.0197\,{\scriptstyle\pm 0.0027}$ & $\mathbf{0.972}$ & $0.376$ & $0.0267\,{\scriptstyle\pm 0.0022}$ \\
Histogram Binning & $\mathbf{0.0139\,{\scriptstyle\pm 0.0042}}$ & $\underline{0.0157\,{\scriptstyle\pm 0.0052}}$ & $1.042$ & $0.385$ & $0.0262\,{\scriptstyle\pm 0.0060}$ \\
BBQ & $\underline{0.0141\,{\scriptstyle\pm 0.0054}}$ & $0.0201\,{\scriptstyle\pm 0.0066}$ & $1.049$ & $0.386$ & $\underline{0.0256\,{\scriptstyle\pm 0.0015}}$ \\
Isotonic Regression & $0.0184\,{\scriptstyle\pm 0.0043}$ & $0.0178\,{\scriptstyle\pm 0.0035}$ & $1.043$ & $0.385$ & $0.0288\,{\scriptstyle\pm 0.0069}$ \\
SB-ECE TS & $0.0161\,{\scriptstyle\pm 0.0048}$ & $\mathbf{0.0152\,{\scriptstyle\pm 0.0008}}$ & $0.977$ & $\mathbf{0.376}$ & $\mathbf{0.0246\,{\scriptstyle\pm 0.0010}}$ \\
RCMMC & $0.0235\,{\scriptstyle\pm 0.0030}$ & $0.0229\,{\scriptstyle\pm 0.0032}$ & $0.978$ & $0.377$ & $0.0311\,{\scriptstyle\pm 0.0035}$ \\
Conf-only & $0.0158\,{\scriptstyle\pm 0.0026}$ & $0.0172\,{\scriptstyle\pm 0.0011}$ & $0.996$ & $0.378$ & $0.0264\,{\scriptstyle\pm 0.0053}$ \\
Conf + PredEntropy & $0.0176\,{\scriptstyle\pm 0.0030}$ & $0.0174\,{\scriptstyle\pm 0.0036}$ & $0.992$ & $0.378$ & $0.0294\,{\scriptstyle\pm 0.0023}$ \\
\textbf{AR-CondCal} & $0.0161\,{\scriptstyle\pm 0.0049}$ & $0.0177\,{\scriptstyle\pm 0.0060}$ & $1.008$ & $0.380$ & $0.0300\,{\scriptstyle\pm 0.0028}$ \\
\midrule
\multicolumn{6}{l}{\textbf{Cell 5.} Block-AR --- DeiT-Small / CIFAR-100}\\
\midrule
No calibration & $0.1331\,{\scriptstyle\pm 0.0062}$ & $0.1331\,{\scriptstyle\pm 0.0062}$ & $1.332$ & $0.449$ & $0.1388\,{\scriptstyle\pm 0.0084}$ \\
Temp. Scaling (TS) & $0.0370\,{\scriptstyle\pm 0.0034}$ & $0.0371\,{\scriptstyle\pm 0.0036}$ & $1.202$ & $0.427$ & $0.0446\,{\scriptstyle\pm 0.0018}$ \\
Ensemble TS & $0.0371\,{\scriptstyle\pm 0.0034}$ & $0.0372\,{\scriptstyle\pm 0.0036}$ & $1.202$ & $0.427$ & $0.0448\,{\scriptstyle\pm 0.0019}$ \\
Vector Scaling & $0.0422\,{\scriptstyle\pm 0.0044}$ & $0.0416\,{\scriptstyle\pm 0.0048}$ & $\underline{1.201}$ & $0.427$ & $0.0529\,{\scriptstyle\pm 0.0051}$ \\
Classwise TS & $0.0412\,{\scriptstyle\pm 0.0033}$ & $0.0396\,{\scriptstyle\pm 0.0027}$ & $1.207$ & $0.427$ & $0.0511\,{\scriptstyle\pm 0.0021}$ \\
Parametric TS & $0.0362\,{\scriptstyle\pm 0.0028}$ & $0.0365\,{\scriptstyle\pm 0.0034}$ & $\mathbf{1.193}$ & $\mathbf{0.425}$ & $0.0426\,{\scriptstyle\pm 0.0024}$ \\
Histogram Binning & $0.0201\,{\scriptstyle\pm 0.0063}$ & $0.0209\,{\scriptstyle\pm 0.0070}$ & $1.279$ & $0.435$ & $0.0285\,{\scriptstyle\pm 0.0053}$ \\
BBQ & $\mathbf{0.0160\,{\scriptstyle\pm 0.0041}}$ & $\mathbf{0.0192\,{\scriptstyle\pm 0.0039}}$ & $1.287$ & $0.435$ & $\mathbf{0.0236\,{\scriptstyle\pm 0.0058}}$ \\
Isotonic Regression & $0.0213\,{\scriptstyle\pm 0.0058}$ & $0.0234\,{\scriptstyle\pm 0.0056}$ & $1.288$ & $0.435$ & $0.0291\,{\scriptstyle\pm 0.0071}$ \\
SB-ECE TS & $0.0174\,{\scriptstyle\pm 0.0031}$ & $\underline{0.0193\,{\scriptstyle\pm 0.0036}}$ & $1.206$ & $\underline{0.425}$ & $0.0276\,{\scriptstyle\pm 0.0017}$ \\
RCMMC & $0.0406\,{\scriptstyle\pm 0.0028}$ & $0.0391\,{\scriptstyle\pm 0.0028}$ & $1.205$ & $0.427$ & $0.0459\,{\scriptstyle\pm 0.0023}$ \\
Conf-only & $0.0189\,{\scriptstyle\pm 0.0025}$ & $0.0248\,{\scriptstyle\pm 0.0026}$ & $1.219$ & $0.428$ & $0.0267\,{\scriptstyle\pm 0.0087}$ \\
Conf + PredEntropy & $0.0186\,{\scriptstyle\pm 0.0025}$ & $0.0209\,{\scriptstyle\pm 0.0038}$ & $1.220$ & $0.428$ & $\underline{0.0236\,{\scriptstyle\pm 0.0043}}$ \\
\textbf{AR-CondCal} & $\underline{0.0172\,{\scriptstyle\pm 0.0034}}$ & $0.0212\,{\scriptstyle\pm 0.0031}$ & $1.228$ & $0.428$ & $0.0280\,{\scriptstyle\pm 0.0063}$ \\
\midrule
\multicolumn{6}{l}{\textbf{Cell 6.} Full-AR --- DeiT-Small / CIFAR-100}\\
\midrule
No calibration & $0.1550\,{\scriptstyle\pm 0.0067}$ & $0.1550\,{\scriptstyle\pm 0.0066}$ & $1.272$ & $0.428$ & $0.1718\,{\scriptstyle\pm 0.0079}$ \\
Temp. Scaling (TS) & $0.0350\,{\scriptstyle\pm 0.0079}$ & $0.0331\,{\scriptstyle\pm 0.0075}$ & $1.123$ & $0.398$ & $0.0454\,{\scriptstyle\pm 0.0060}$ \\
Ensemble TS & $0.0350\,{\scriptstyle\pm 0.0079}$ & $0.0332\,{\scriptstyle\pm 0.0075}$ & $1.123$ & $0.398$ & $0.0457\,{\scriptstyle\pm 0.0060}$ \\
Vector Scaling & $0.0385\,{\scriptstyle\pm 0.0029}$ & $0.0371\,{\scriptstyle\pm 0.0046}$ & $\underline{1.121}$ & $0.398$ & $0.0505\,{\scriptstyle\pm 0.0059}$ \\
Classwise TS & $0.0379\,{\scriptstyle\pm 0.0047}$ & $0.0369\,{\scriptstyle\pm 0.0054}$ & $1.130$ & $0.398$ & $0.0511\,{\scriptstyle\pm 0.0020}$ \\
Parametric TS & $0.0309\,{\scriptstyle\pm 0.0023}$ & $0.0302\,{\scriptstyle\pm 0.0054}$ & $\mathbf{1.110}$ & $\mathbf{0.395}$ & $0.0400\,{\scriptstyle\pm 0.0075}$ \\
Histogram Binning & $0.0195\,{\scriptstyle\pm 0.0059}$ & $0.0178\,{\scriptstyle\pm 0.0035}$ & $1.201$ & $0.406$ & $0.0287\,{\scriptstyle\pm 0.0036}$ \\
BBQ & $\underline{0.0142\,{\scriptstyle\pm 0.0018}}$ & $0.0233\,{\scriptstyle\pm 0.0089}$ & $1.204$ & $0.407$ & $\underline{0.0265\,{\scriptstyle\pm 0.0029}}$ \\
Isotonic Regression & $0.0193\,{\scriptstyle\pm 0.0029}$ & $0.0188\,{\scriptstyle\pm 0.0018}$ & $1.206$ & $0.406$ & $0.0313\,{\scriptstyle\pm 0.0032}$ \\
SB-ECE TS & $0.0206\,{\scriptstyle\pm 0.0006}$ & $0.0203\,{\scriptstyle\pm 0.0018}$ & $1.129$ & $\underline{0.397}$ & $0.0327\,{\scriptstyle\pm 0.0022}$ \\
RCMMC & $0.0353\,{\scriptstyle\pm 0.0032}$ & $0.0335\,{\scriptstyle\pm 0.0061}$ & $1.128$ & $0.398$ & $0.0435\,{\scriptstyle\pm 0.0074}$ \\
Conf-only & $0.0150\,{\scriptstyle\pm 0.0017}$ & $\underline{0.0160\,{\scriptstyle\pm 0.0025}}$ & $1.146$ & $0.399$ & $0.0294\,{\scriptstyle\pm 0.0065}$ \\
Conf + PredEntropy & $\mathbf{0.0124\,{\scriptstyle\pm 0.0029}}$ & $\mathbf{0.0156\,{\scriptstyle\pm 0.0020}}$ & $1.141$ & $0.399$ & $\mathbf{0.0240\,{\scriptstyle\pm 0.0022}}$ \\
\textbf{AR-CondCal} & $0.0164\,{\scriptstyle\pm 0.0030}$ & $0.0171\,{\scriptstyle\pm 0.0027}$ & $1.152$ & $0.400$ & $0.0299\,{\scriptstyle\pm 0.0043}$ \\
\midrule
\multicolumn{6}{l}{\textbf{Cell 7.} Full-AR --- Swin-Tiny / CIFAR-100}\\
\midrule
No calibration & $0.1238\,{\scriptstyle\pm 0.0005}$ & $0.1238\,{\scriptstyle\pm 0.0005}$ & $0.985$ & $0.357$ & $0.1362\,{\scriptstyle\pm 0.0039}$ \\
Temp. Scaling (TS) & $0.0226\,{\scriptstyle\pm 0.0019}$ & $0.0214\,{\scriptstyle\pm 0.0024}$ & $\underline{0.873}$ & $\underline{0.336}$ & $0.0372\,{\scriptstyle\pm 0.0012}$ \\
Ensemble TS & $0.0227\,{\scriptstyle\pm 0.0019}$ & $0.0214\,{\scriptstyle\pm 0.0024}$ & $0.873$ & $0.336$ & $0.0372\,{\scriptstyle\pm 0.0013}$ \\
Vector Scaling & $0.0261\,{\scriptstyle\pm 0.0034}$ & $0.0250\,{\scriptstyle\pm 0.0035}$ & $0.878$ & $0.338$ & $0.0404\,{\scriptstyle\pm 0.0026}$ \\
Classwise TS & $0.0254\,{\scriptstyle\pm 0.0019}$ & $0.0225\,{\scriptstyle\pm 0.0037}$ & $0.878$ & $0.336$ & $0.0361\,{\scriptstyle\pm 0.0024}$ \\
Parametric TS & $0.0212\,{\scriptstyle\pm 0.0023}$ & $0.0195\,{\scriptstyle\pm 0.0026}$ & $\mathbf{0.873}$ & $0.336$ & $\underline{0.0304\,{\scriptstyle\pm 0.0014}}$ \\
Histogram Binning & $0.0169\,{\scriptstyle\pm 0.0027}$ & $0.0182\,{\scriptstyle\pm 0.0027}$ & $0.934$ & $0.343$ & $0.0320\,{\scriptstyle\pm 0.0061}$ \\
BBQ & $0.0185\,{\scriptstyle\pm 0.0058}$ & $0.0265\,{\scriptstyle\pm 0.0060}$ & $0.948$ & $0.345$ & $0.0310\,{\scriptstyle\pm 0.0090}$ \\
Isotonic Regression & $\underline{0.0154\,{\scriptstyle\pm 0.0003}}$ & $0.0182\,{\scriptstyle\pm 0.0011}$ & $0.938$ & $0.343$ & $0.0308\,{\scriptstyle\pm 0.0037}$ \\
SB-ECE TS & $\mathbf{0.0144\,{\scriptstyle\pm 0.0006}}$ & $\mathbf{0.0156\,{\scriptstyle\pm 0.0021}}$ & $0.878$ & $\mathbf{0.335}$ & $0.0305\,{\scriptstyle\pm 0.0068}$ \\
RCMMC & $0.0235\,{\scriptstyle\pm 0.0021}$ & $0.0226\,{\scriptstyle\pm 0.0030}$ & $0.885$ & $0.337$ & $0.0338\,{\scriptstyle\pm 0.0034}$ \\
Conf-only & $0.0176\,{\scriptstyle\pm 0.0017}$ & $0.0202\,{\scriptstyle\pm 0.0016}$ & $0.891$ & $0.338$ & $0.0344\,{\scriptstyle\pm 0.0062}$ \\
Conf + PredEntropy & $0.0170\,{\scriptstyle\pm 0.0033}$ & $0.0201\,{\scriptstyle\pm 0.0036}$ & $0.887$ & $0.337$ & $0.0351\,{\scriptstyle\pm 0.0057}$ \\
\textbf{AR-CondCal} & $0.0188\,{\scriptstyle\pm 0.0024}$ & $\underline{0.0169\,{\scriptstyle\pm 0.0006}}$ & $0.899$ & $0.339$ & $\mathbf{0.0300\,{\scriptstyle\pm 0.0061}}$ \\

\end{longtable}
}

\section{Parameter overhead of the attention-residual path}
\label{app:overhead}

Tab.~\ref{tab:overhead-params} reports total parameter counts for the
three substrates (ViT-B/16, DeiT-Small, Swin-Tiny) under Baseline,
Full-AR, and Block-AR, re-constructed from the same model factories
used at training time (Appendix~\ref{app:provenance}). The AR path
adds a small learned residual module and contributes $<\!0.1\%$ of
total parameters on every substrate; Full-AR and Block-AR share the
same module (Block-AR just gates its contribution to a subset of
depths), so their parameter counts are identical. The post-hoc
AR-CondCal calibrator itself is non-parametric (a $2$-D
Nadaraya--Watson kernel fit on the calibration split; no trainable
weights), so the only additional inference cost is (i)~extracting
per-layer routing entropy $H_l(x)$, which is one softmax-and-mean per
transformer layer and is dwarfed by the attention it summarises, and
(ii)~evaluating the kernel, which is $O(n_{\mathrm{cal}})$ per test
sample in a $2$-D feature space. We report parameter counts rather
than wall-clock throughput because the paper's main comparison
(Tab.~\ref{tab:calibration_block_ar}) is a post-hoc calibration
comparison, not a training-efficiency comparison, and because target
inference hardware varies.

\begin{table}[!ht]
\centering
\setlength{\tabcolsep}{4pt}
\small
\caption{Parameter counts (millions) for the three training substrates
under Baseline, Full-AR, and Block-AR, re-constructed from the same
model factories used at training time
(Appendix~\ref{app:provenance}). Percent delta is relative to the
Baseline column of the same substrate. Full-AR and Block-AR add the
same learned residual module; they differ only in which depths the
module is activated at.}
\label{tab:overhead-params}
\begin{tabular}{lrrrrr}
\toprule
Substrate & Baseline & Full-AR & Block-AR & $\Delta$ params & $\Delta\%$ \\
\midrule
ViT-B/16    & $85.779$ & $85.816$ & $85.816$ & $+0.037$ & $+0.04\%$ \\
DeiT-Small  & $21.660$ & $21.679$ & $21.679$ & $+0.019$ & $+0.09\%$ \\
Swin-Tiny   & $27.527$ & $27.545$ & $27.545$ & $+0.018$ & $+0.07\%$ \\
\bottomrule
\end{tabular}
\end{table}

\clearpage
\section{Full-AR contrastive diagnostic (companion to Fig.~\ref{fig:priority2})}
\label{app:full-ar-pathology}

\begin{figure}[!ht]
  \centering
  \includegraphics[width=\linewidth]{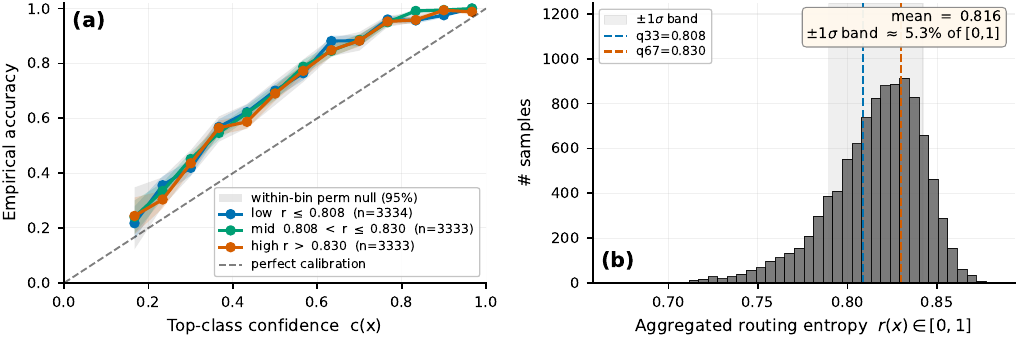}
  \caption{Routing-conditional calibration on Full-AR (contrastive).
  Same analysis and conventions as Fig.~\ref{fig:priority2} (gray band
  $=$ within-bin permutation null at $95\%$; gray $\pm 1\sigma$ in
  panel b), applied to Swin-Tiny $+$ Full-AR, CIFAR-10, ep $=$ 299.
  Maximum matched-confidence gap is $0.052$ vs Block-AR's $0.346$;
  a smaller-or-larger gap on either variant does \emph{not} mean
  one is better calibrated overall (raw global ECE is similar in
  scale, $0.162$ vs $0.149$). Quantitative Block-vs-Full comparison
  is Tab.~\ref{tab:phenomenon}.}
  \label{fig:priority2-full}
\end{figure}

\section{Robustness checks for the routing-conditional gap}
\label{app:gap-robustness}

This appendix reports four diagnostic robustness checks for the
matched-confidence routing gap on Block-AR ($0.346$) and
Full-AR ($0.052$) reported in Tab.~\ref{tab:phenomenon} /
Fig.~\ref{fig:priority2} (Swin-Tiny + CIFAR-10 seed-$0$). All
checks are computed on the same full $10{,}000$-sample CIFAR-10
test population, with the same $15$ equal-width confidence bins,
the same $\mathrm{q33}/\mathrm{q67}$ tertile cuts on aggregate
routing entropy, and the same ``$n \ge 5$ samples per shared bin''
rule used by the diagnostic pipeline. None of these checks
introduces a new empirical claim; they bound and qualify the
existing one.

\begin{figure}[!h]
  \centering
  \includegraphics[width=0.9\linewidth,trim={0 0 0 90},clip]{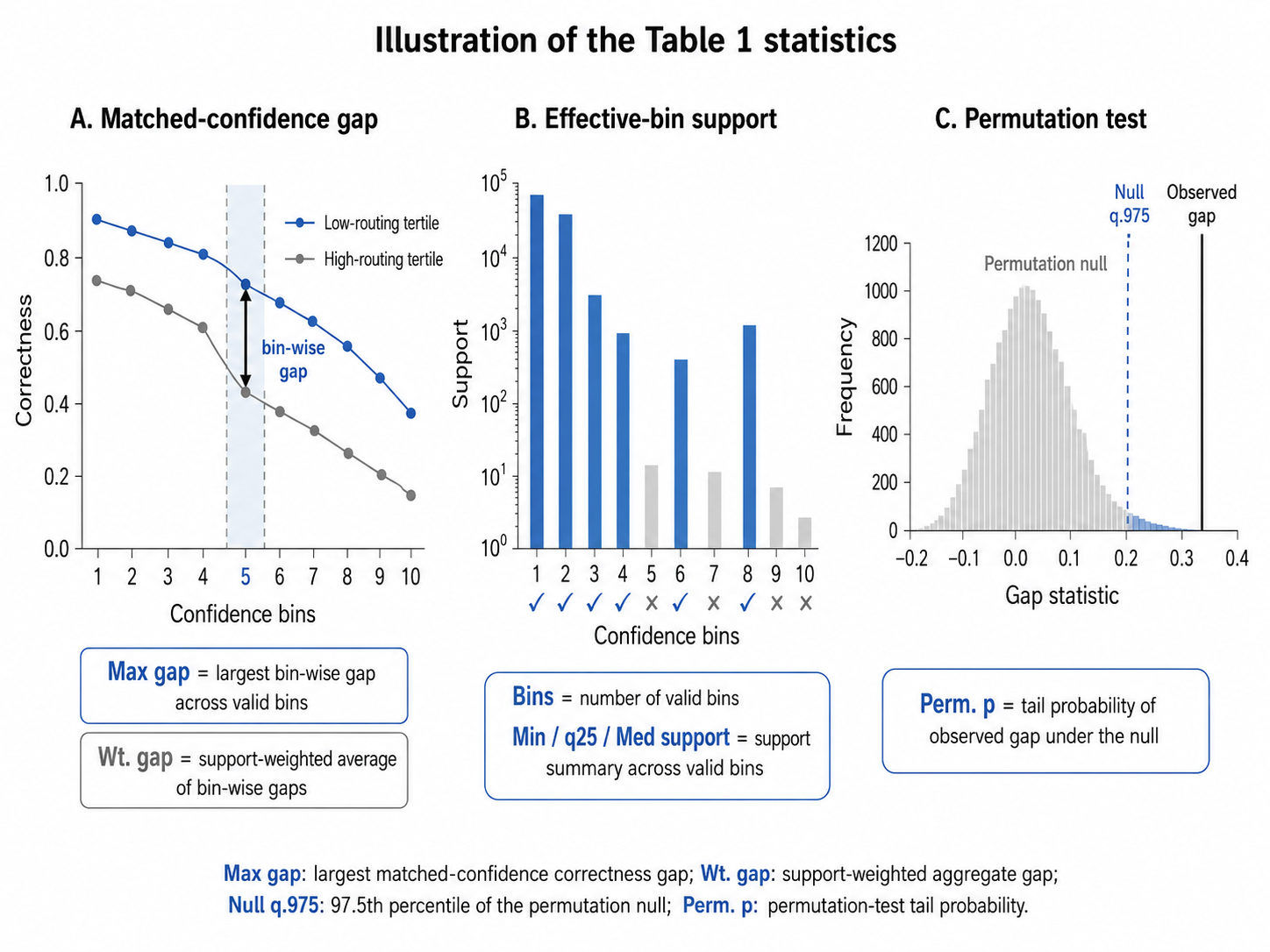}
  \caption{\textbf{Visual definition of the diagnostic statistics
    in Tab.~\ref{tab:phenomenon}.}
    \emph{(A)} Matched-confidence reliability curves for the
    low-routing-tertile (light) and high-routing-tertile (dark)
    populations across $15$ equal-width confidence bins; the
    bin-wise absolute difference defines the bin-level gap.
    \textbf{Max gap} is its maximum over shared valid bins;
    \textbf{Wt.\ gap} is the support-weighted average across the
    same bins.
    \emph{(B)} Per-bin support: only bins with $n\!\ge\!5$ samples
    in \emph{both} tertiles contribute. The \textbf{Bins} and
    Min / $q_{25}$ / Med columns of Tab.~\ref{tab:phenomenon}
    summarise this count distribution and rule out empty-tail
    artifacts.
    \emph{(C)} Within-bin routing-permutation null distribution of
    the max-gap statistic ($P\!=\!5000$ shuffles per confidence
    bin). \textbf{Null $q_{.975}$} is the $97.5$th percentile of
    this null; \textbf{Perm.\ $p$} is the upper-tail probability of
    the observed gap.}
  \label{fig:tab1-illustration}
\end{figure}

\textbf{Bootstrap CI for the max-gap statistic.}
Resampling the $10{,}000$ test instances with replacement
($B = 5000$ resamples, fixed tertile cuts $\mathrm{q33}, \mathrm{q67}$
re-applied per replicate) gives the $95\%$ percentile bootstrap CIs
in Tab.~\ref{tab:gap-robustness}: Block-AR max-gap CI
$[0.109, 0.678]$, Full-AR $[0.049, 0.216]$. Both CIs are
wide, reflecting the known instability of the max-over-bins
statistic under finite-sample resampling; the Block-AR
point estimate ($0.346$) sits in the lower half of its CI and
the Full-AR point estimate ($0.052$) sits at the lower edge of
its CI, both consistent with the high variance of the
max-over-bins statistic under a per-bin minimum-sample-count
rule ($n \ge 5$). The CIs are reported as a robustness summary
of the sampling distribution; we use the permutation test below
as the principled significance check.

\textbf{Within-confidence-bin permutation test.}
We test whether routing-tertile membership carries information
about correctness once confidence-bin membership is fixed.
Inside each of the $15$ confidence bins we permute the routing
feature $r(x)$ across samples ($P = 5000$ permutations), recompute
the max-gap statistic, and report the resulting $p$-value.
Block-AR yields $p = 0.042$ and Full-AR $p = 0.888$. The
Block-AR rejection on this seed is the only seed-level nominal
rejection in the $30$-run AR sweep
(\S\ref{sec:rq2}, Tab.~\ref{tab:ar-sweep}); the other two seeds
in the same cell give $p\!=\!0.483$ (seed-$1$) and $p\!=\!0.713$
(seed-$2$) and do not reject. We therefore treat this single
seed-level rejection as a noisy max-over-bins instability rather
than as substrate-level evidence, consistent with the
multicalibration-style formalisation in
Proposition~2 of Appendix~\ref{app:theory}: the population-level
blind spot is identifiable, and the empirical max-gap is a
high-variance estimator of it on any single seed.

\textbf{Bin-support diagnostics.}
The max-gap and the weighted-integrated gap below are computed
over the bins shared by the low- and high-routing tertiles.
Tab.~\ref{tab:gap-robustness} reports the number of shared bins
and the per-shared-bin sample-count statistics
($\min$, $25^{\text{th}}$, median); Block-AR shares $12$ of $15$
bins (Full-AR $13$ of $15$), with tens to hundreds of samples per
shared bin, so the max-gap is not driven by a near-empty bin.

\textbf{Weighted integrated gap.}
As a less-brittle summary, we report a weighted mean
$|a_{\mathrm{low}}(c) - a_{\mathrm{high}}(c)|$ across shared bins,
weighted by $w_c = \min(n_{\mathrm{low}}^c, n_{\mathrm{high}}^c)$.
The point estimates are Block-AR $0.016$ and Full-AR $0.016$
with $95\%$ percentile-bootstrap CIs $[0.012, 0.030]$ and
$[0.018, 0.043]$ respectively, overlapping. The Full-AR
point estimate ($0.0163$) sits slightly below its CI lower
bound ($0.0180$): for substrates whose per-bin gaps are
near zero, the absolute-value functional in Wt.\ gap folds
the resampled noise around small per-bin gaps onto the
positive axis, shifting the percentile distribution upward
relative to the original-sample point. The qualitative
conclusion is unaffected: the two variants are statistically
indistinguishable on this stable summary, consistent with the
``directional rather than statistically resolved'' framing in
the main body and with the $30$-run sweep in
Tab.~\ref{tab:ar-sweep}. The much smaller
absolute scale of the weighted gap (vs the max-gap) reflects
the fact that most shared bins exhibit much smaller
routing-conditional differences than the extreme bin captured
by the max statistic.

\begin{table}[!ht]
\centering
\setlength{\tabcolsep}{4pt}
\small
\caption{Robustness checks for the routing-conditional max-gap
(Swin-Tiny CIFAR-10 seed-$0$, ep $=$ 299, full $10{,}000$-sample
test set; $15$ confidence bins; $n \ge 5$ per shared bin; tertile
cuts $\mathrm{q33}, \mathrm{q67}$ of $r_{\mathrm{agg}}$). CI columns
are $95\%$ percentile bootstrap intervals over $B = 5000$
resamples; permutation $p$-value is computed over $P = 5000$
within-confidence-bin permutations. Bin-support shows the number
of bins shared by the low/high tertiles, the minimum, the
$25^{\text{th}}$ percentile, and the median sample-count per
shared bin (using $\min(n_{\mathrm{low}}, n_{\mathrm{high}})$).}
\label{tab:gap-robustness}
\begin{tabular}{lrrrr}
\toprule
Statistic & Block-AR & Full-AR & & \\
\midrule
Max gap (point) & $0.346$ & $0.052$ & & \\
\hspace{1em} $95\%$ bootstrap CI         & $[0.109,\,0.678]$ & $[0.049,\,0.216]$ & & \\
Weighted integrated gap (point) & $0.016$ & $0.016$ & & \\
\hspace{1em} $95\%$ bootstrap CI         & $[0.012,\,0.030]$ & $[0.018,\,0.043]$ & & \\
Within-bin permutation $p$ & $\mathbf{0.042}$ & $0.888$ & & \\
Shared bins (of $15$) & $12$ & $13$ & & \\
\hspace{1em} min/p25/median samples per shared bin & $11$/$86$/$143$ & $41$/$222$/$280$ & & \\
\bottomrule
\end{tabular}
\end{table}

\begin{table}[!ht]
\centering
\setlength{\tabcolsep}{4pt}
\footnotesize
\caption{\textbf{Per-seed Swin-Tiny CIFAR-10 diagnostic statistics
corresponding to the Tab.~\ref{tab:phenomenon} / Fig.~\ref{fig:priority2}
cell and its matched three-seed context.} Acc@1 is included here
for transparency; the main diagnostic quantities are the
matched-confidence Max gap, Wt.\ gap, and within-bin permutation
$p$-value. The only nominal rejection is Block-AR ($b\!=\!2$)
seed-$0$ ($p\!=\!0.042$); the other Block-AR seeds and all Full-AR
seeds do not reject. All values are computed from the same
$10{,}000$-sample CIFAR-10 test set, ep$\,=\,$299, with the same
$15$-bin equal-width / $n\!\ge\!5$ / tertile-cuts protocol as
Tab.~\ref{tab:phenomenon} ($B\!=\!P\!=\!5000$, RNG seed~$42$). Wt.\ CI
is the percentile bootstrap CI for Wt.\ gap; Bins is shared/$15$;
Min/$q_{25}$/Med is per-shared-bin support
$\min(n_{\mathrm{low}}, n_{\mathrm{high}})$.}
\label{tab:perseed-swin-c10-diagnostic}
\begin{tabular}{llcccccccc}
\toprule
Variant & Seed & Acc@1 & Max gap & Wt.\ gap & Wt.\ CI & Bins & Min/$q_{25}$/Med & Perm.\ $p$ & Reject? \\
\midrule
Block-AR ($b\!=\!2$) & $0$ & $0.903$ & $0.346$ & $0.016$ & $[0.012,\,0.030]$ & $12/15$ & $11/86/143$  & $\mathbf{0.042}$ & yes \\
Block-AR ($b\!=\!2$) & $1$ & $0.901$ & $0.133$ & $0.011$ & $[0.011,\,0.028]$ & $12/15$ & $18/117/165$ & $0.483$ & no  \\
Block-AR ($b\!=\!2$) & $2$ & $0.902$ & $0.100$ & $0.017$ & $[0.013,\,0.032]$ & $12/15$ & $25/134/171$ & $0.713$ & no  \\
\midrule
Full-AR              & $0$ & $0.724$ & $0.052$ & $0.016$ & $[0.018,\,0.043]$ & $13/15$ & $41/222/280$ & $0.888$ & no  \\
Full-AR              & $1$ & $0.921$ & $0.160$ & $0.018$ & $[0.014,\,0.030]$ & $12/15$ & $10/92/128$  & $0.512$ & no  \\
Full-AR              & $2$ & $0.924$ & $0.167$ & $0.014$ & $[0.011,\,0.026]$ & $12/15$ & $9/72/117$   & $0.578$ & no  \\
\bottomrule
\end{tabular}
\end{table}

\textbf{Correctness probe (capacity-uncontrolled).}
For completeness, the naive entropy-vector probe in
Appendix~\ref{app:mechanism} (Tab.~\ref{tab:routing-probe}) shows
a Block-vs-Full asymmetry under a capacity-uncontrolled comparison:
a $2$-layer MLP predicting $|\mathrm{conf}(x) - \mathrm{correct}(x)|$
from $(c(x), H_1(x), \ldots, H_L(x))$ adds an $R^2$ uplift over a
confidence-only linear baseline on Swin Block-AR but essentially
no uplift on Swin Full-AR. As reported in App.~\ref{app:probe-sanity},
this apparent uplift does not survive a capacity-matched
confidence-only MLP or a shuffled-routing control: the
routing-specific interpretation disappears under the matched
controls. The Block-vs-Full asymmetry in
Tab.~\ref{tab:routing-probe} therefore reflects a non-routing
property of the comparison rather than recoverable
routing-specific signal.

The robustness checks above were generated by the
diagnostic-gap-robustness pipeline from the on-disk routing
caches across all three Block-AR seeds; intermediate JSON / CSV
outputs are shipped in the supplementary material.

\section{Full method list (companion to Fig.~\ref{fig:bench-main})}
\label{app:bench-main-full}

\begin{figure}[h]
  \centering
  \includegraphics[width=\linewidth]{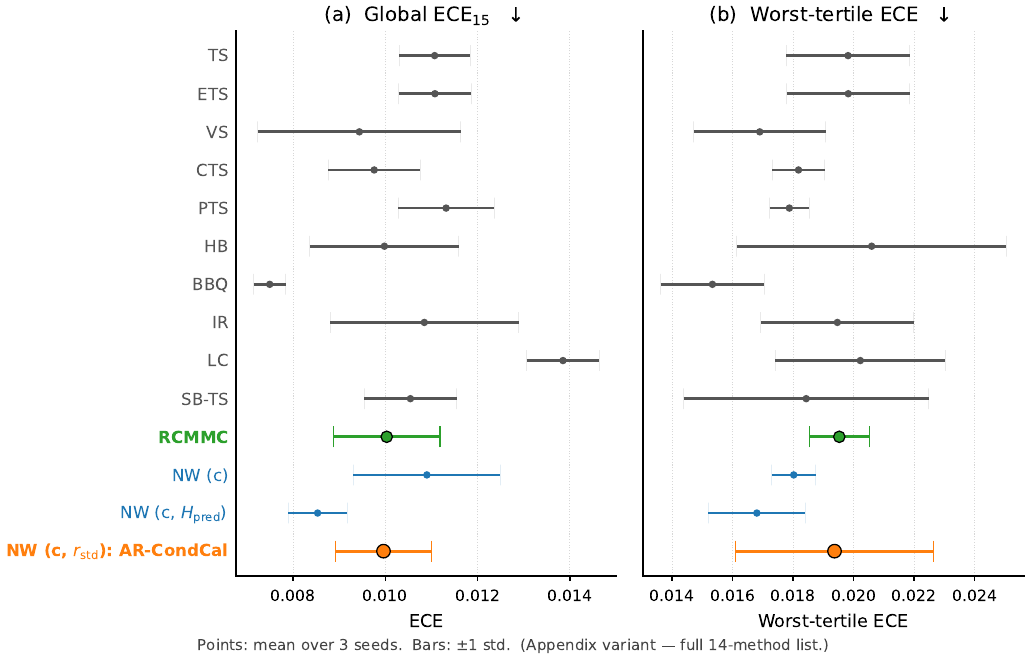}
  \caption{\textbf{Appendix variant of Fig.~\ref{fig:bench-main}:}
    full $14$-method list (same dot-and-error-bar style and the
    same $3$-seed Block-AR data). The matched-kernel \textbf{NW
    family} (blue/orange) is the same as in the main-text figure;
    grey points are the eleven non-NW calibrators (TS, ETS, VS,
    CTS, PTS, HB, BBQ, IR, LC, SB-TS, RCMMC). All $14$ methods
    have visible $\pm 1$ std bars on both panels under
    Block-AR ($b\!=\!2$) (e.g.\ TS / ETS global-ECE seed-std
    $\approx\!0.0008$, matching Tab.~\ref{tab:calibration_block_ar}). \emph{Per-seed overlay points are
    intentionally omitted to keep the figure legible at appendix
    scale.}}
  \label{fig:bench-main-full}
\end{figure}

\section{Full-AR method-comparison figure (companion to Fig.~\ref{fig:bench-main})}
\label{app:bench-full}

\begin{figure}[h]
  \centering
  \includegraphics[trim={0 0 0 12pt},clip,width=\linewidth]{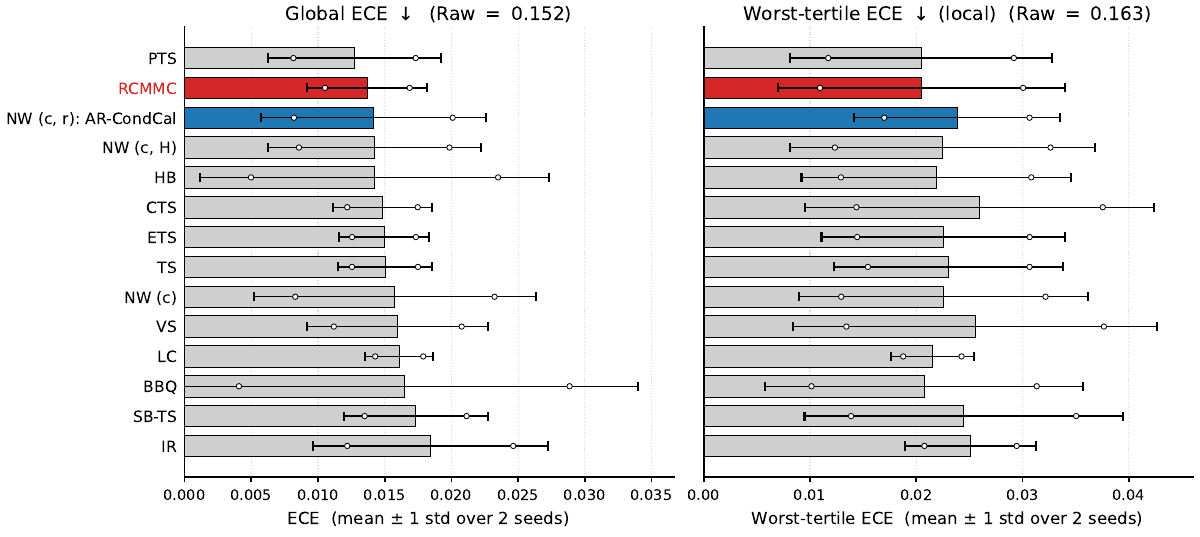}
  \caption{Method comparison on Full-AR (contrastive to
    Fig.~\ref{fig:bench-main}; same horizontal multi-seed
    style). Aggregated over $2$ Full-AR training seeds
    ($s_0,s_1$). Bar length is mean ECE; error
    bars are $\pm 1$~std; white dots are per-seed values.
    \emph{Left:} Global ECE. \emph{Right:} Worst-routing-tertile
    ECE. As on Block-AR, the per-seed std is dominated by the
    non-parametric kernel and binning methods (BBQ, HB,
    AR-CondCal) while the scalar parametric methods (TS / ETS)
    remain seed-stable; AR-CondCal's mean global ECE
    ($0.0141\!\pm\!0.0084$) sits inside the leader pack
    (PTS $0.0127\!\pm\!0.0065$, RCMMC $0.0137\!\pm\!0.0045$),
    and its mean worst-tertile ECE ($0.0238\!\pm\!0.0097$) is
    comparable to Conf-only ($0.0226\!\pm\!0.0136$). The
    $1$-D-projection-is-noise picture replicates on this
    contrastive substrate.}
  \label{fig:bench-full}
\end{figure}

\section{Bandwidth sensitivity for AR-CondCal}
\label{app:bw-sensitivity}

\S\ref{sec:rq3} reports AR-CondCal under the default Scott
bandwidth. To check whether the worst-tertile instability is
explained by that fixed bandwidth choice, we sweep eight
seed-level AR runs across five CIFAR-10 substrate / variant
groups: Swin-Tiny~$+$~Block-AR ($3$ seeds),
DeiT-Small~$+$~Block-AR ($1$ seed),
ViT-B/$16$~$+$~Block-AR ($1$ seed),
DeiT-Small~$+$~Full-AR ($2$ seeds), and
ViT-B/$16$~$+$~Full-AR ($1$ seed). We use the same
calibration-benchmark population as
Tab.~\ref{tab:calibration_block_ar}: a fixed seed-$42$
$5{,}000/5{,}000$ calibration/test split, with $15$ equal-width
confidence bins inside each $r_{\mathrm{std}}$ tertile.
We compare five bandwidth modes: $0.5\times$, $1\times$,
$2\times$ the per-axis Scott bandwidth; the multiplier from
$\{0.25, 0.5, 1, 2, 4\}$ minimising mean $5$-fold calibration
NLL (\textbf{CV-NLL}); and the multiplier minimising held-out
global ECE (\textbf{Oracle-ECE}, a deliberately optimistic
ceiling, not a selection method).

\begin{table}[h]
\centering\small
\caption{\textbf{Bandwidth sensitivity for AR-CondCal.} Pooled
mean$\,\pm\,$std across the available eight CIFAR-10 AR
checkpoints described above.}
\label{tab:bw-sensitivity}
\begin{tabular}{lcccc}
\toprule
Mode & Multiplier / selection & Global ECE & Worst-tertile ECE & NLL \\
\midrule
$0.5\!\times\!$ Scott & $0.50$                  & $0.0182\!\pm\!0.0039$ & $0.0288\!\pm\!0.0080$ & $0.4589\!\pm\!0.2307$ \\
$1\!\times\!$ Scott   & $1.00$ (paper default)  & $0.0105\!\pm\!0.0041$ & $0.0235\!\pm\!0.0074$ & $0.4348\!\pm\!0.2343$ \\
$2\!\times\!$ Scott   & $2.00$                  & $0.0159\!\pm\!0.0085$ & $0.0261\!\pm\!0.0102$ & $0.4336\!\pm\!0.2398$ \\
CV-NLL                & $5$-fold cal.\ NLL      & $0.0129\!\pm\!0.0034$ & $0.0244\!\pm\!0.0067$ & $0.4317\!\pm\!0.2361$ \\
Oracle-ECE            & held-out global ECE     & $0.0103\!\pm\!0.0037$ & $0.0232\!\pm\!0.0069$ & $0.4362\!\pm\!0.2372$ \\
\bottomrule
\end{tabular}
\end{table}

\textbf{Reading.} CV-NLL changes worst-tertile ECE by only
$+0.0009$ relative to $1\!\times\!$ Scott, well within the
cross-run standard deviation of approximately $0.007$; the
global-ECE oracle reduces worst-tertile ECE by $0.0003$ relative
to $1\!\times\!$ Scott, also within the cross-run standard
deviation. No single multiplier is consistently best across the
eight runs; the worst-tertile pattern is therefore not driven by
the default Scott bandwidth.

\textbf{Caveat.} The audit does not address anisotropic
bandwidth grids, non-Gaussian kernels, or objectives that tune
worst-tertile ECE directly; it bounds only the default-bandwidth
explanation within the kernel family evaluated.

\section{Capacity-matched audit of the full-profile probe}
\label{app:probe-sanity}

\S\ref{sec:probe} reports the audit summary. This appendix gives
the protocol and the per-model pooled $R^2$. We replicate the
naive Tab.~\ref{tab:routing-probe} probe (parametric two-layer
MLP, hidden $16$, $200$ Adam epochs at lr $10^{-2}$ /
wd $10^{-4}$, $50/50$ random split on seed $42$, target
$|\mathrm{conf}(x)\!-\!\mathrm{correct}(x)|$) and add three
controls: a closed-form ridge regressor on the same full vector
(\textsc{full-lin}); a \emph{capacity-matched confidence-only
MLP} with $c$ as sole input but the same architecture and
training schedule (\textsc{conf-mlp}); and a
\emph{shuffled-routing full-vector MLP} (\textsc{shuf-full-mlp})
where the per-layer routing profile $(H_1, \ldots, H_L)$ is
permuted across samples within the training fold while
$c$ and the target stay aligned, so the MLP sees the same
input-vector geometry as \textsc{full-mlp} but with the routing
information randomised. The sweep spans $18$ seed-level cells in
$10$ substrate $\times$ variant groups (Swin / DeiT / ViT,
Block-AR / Full-AR, CIFAR-10 and CIFAR-100; the corrected
absolute-block-size variant is used for the C-100 Swin Block-AR
cells, as noted in App.~\ref{app:per-cell-cal}).

\begin{table}[h]
\centering\small
\caption{\textbf{Capacity-matched audit of the full-profile
probe.} Pooled mean$\,\pm\,$std of held-out $R^2$ on
$|\mathrm{conf}-\mathrm{correct}|$ across the $18$ seed-level
AR cells described above.}
\label{tab:probe-sanity}
\begin{tabular}{lcc}
\toprule
Model & Input & Held-out $R^2$ \\
\midrule
\textsc{conf-lin}      & $c$              & $0.4722\!\pm\!0.177$ \\
\textsc{conf-mlp}      & $c$              & $\mathbf{0.6017\!\pm\!0.094}$ \\
\textsc{full-lin}      & $(c, H_1, \ldots, H_L)$ & $0.4719\!\pm\!0.177$ \\
\textsc{shuf-full-mlp} & $(c, \tilde H_1, \ldots, \tilde H_L)$ & $0.5616\!\pm\!0.135$ \\
\textsc{full-mlp}      & $(c, H_1, \ldots, H_L)$ & $0.5588\!\pm\!0.140$ \\
\bottomrule
\end{tabular}
\end{table}

\textbf{Reading.} \textsc{conf-mlp} achieves the highest pooled
$R^2$ and the lowest cross-cell std; \textsc{full-mlp} minus
\textsc{conf-mlp} is negative in $9/10$ substrate groups (range
$-0.21$ to $-0.004$; the c10 Swin Block-AR group is slightly
positive at $+0.022\!\pm\!0.009$, within the cross-cell std
band); \textsc{shuf-full-mlp} sits within $0.003$ of
\textsc{full-mlp}, so randomising the routing profile does not
measurably reduce held-out $R^2$. The naive
$\textsc{full-mlp} - \textsc{conf-lin}$ uplift reported in
Tab.~\ref{tab:routing-probe} (mean $+0.074\!\pm\!0.038$ on $3$
Block-AR cells; $+0.087\!\pm\!0.089$ pooled across $18$)
therefore reflects non-linear modelling of confidence and the
wider input vector, not routing-specific information.

\textbf{Caveat.} The audit does not preclude routing-specific
gains under richer probes or different objectives; under the
probes and controls evaluated here, the full-profile gain is not
routing-specific.

\section{Supporting propositions}
\label{app:theory}

This appendix contains three supporting propositions and one
quoted standard consistency fact for Nadaraya--Watson regression.
They provide population-level design motivation for the
AR-CondCal probe and clarify the conditions under which a
routing-derived second feature can or cannot help. Propositions~1,
2, and~4 are stated with their assumptions and proven in full;
Standard~fact~3 is included as a quoted standard consistency
statement for Nadaraya--Watson regression and is not claimed as a
new theorem. None of the four statements is a finite-sample
guarantee or a claim of dominance over global calibrators on any
metric.

\textbf{Setup.}
Let $(X, Y)$ be a classification input--label pair from the
population distribution, $\hat{y}(X) = \arg\max_k p_k(X)$ the argmax
prediction, and $T := \mathbb{1}[\hat{y}(X) = Y] \in \{0, 1\}$ the
binary correctness indicator. Let $c(X) = \max_k p_k(X)$ be the
top-class confidence and $r(X)$ a possibly-vector-valued,
routing-derived feature (in AR-CondCal,
$r = r_{\mathrm{std}}$, the depth-variance of per-layer routing
entropy). Define the oracle conditional correctness functions
\begin{equation}
\eta_c(c) := \mathbb{E}[T \mid c(X) = c],
\qquad
\eta_{c,r}(c, r) := \mathbb{E}[T \mid c(X) = c,\, r(X) = r].
\end{equation}
We assume $T$ has finite second moment throughout.

\textbf{Proposition 1 (oracle gain $=$ conditional explained
variance).}
\emph{The reduction in oracle squared-error risk from adding the
feature $r$ to confidence $c$ equals the $L^2$ distance between the
two oracle conditional means:}
\begin{equation}
\label{eq:explained-variance}
\mathbb{E}\bigl[(T - \eta_c(c(X)))^2\bigr]
\;-\;
\mathbb{E}\bigl[(T - \eta_{c,r}(c(X), r(X)))^2\bigr]
\;=\;
\mathbb{E}\bigl[(\eta_{c,r}(c(X), r(X)) - \eta_c(c(X)))^2\bigr] \;\ge\; 0.
\end{equation}
\emph{In particular, the oracle gain is strictly positive iff
$\eta_{c,r}(c,r) \neq \eta_c(c)$ on a positive-mass set, and is
zero iff $r$ carries no additional information about $T$ given
$c$ almost surely.}

\emph{Proof.} Writing $A := \eta_{c,r}(c(X), r(X))$ and
$B := \eta_c(c(X))$, both $A$ and $B$ are conditional expectations
of $T$ with respect to $\sigma$-algebras
$\sigma(c, r) \supseteq \sigma(c)$, so $A - B$ is
$\sigma(c, r)$-measurable and $\mathbb{E}[T - A \mid \sigma(c, r)]
= 0$. Decomposing $T - B = (T - A) + (A - B)$ and using this
orthogonality gives
$\mathbb{E}[(T - B)^2] = \mathbb{E}[(T - A)^2] + \mathbb{E}[(A - B)^2]$.
Rearranging gives~\eqref{eq:explained-variance}.~\hfill$\square$

\textbf{Remark.}
This is a Pythagorean sharpening of the standard squared-loss
ordering: the oracle benefit of adding $r$ \emph{equals} the
amount of correctness variation $r$ explains beyond confidence,
not merely a non-negative inequality. It is included to make the
empirical interpretation of the feature ablation in
Tab.~\ref{tab:feature_ablation} precise: the question is which
$r(X)$ best populates the right-hand side of
\eqref{eq:explained-variance}, i.e., which routing summary
explains the most additional correctness variation at matched
confidence on the evaluated substrate. It does \emph{not} imply
that any particular finite-sample kernel calibrator achieves
this gain, nor that aggregating over a binned ECE metric
preserves it.

\textbf{Proposition 2 (confidence-only blind spot at routing-defined
subgroups).}
\emph{Let $G : \mathcal{X} \to \{1, \dots, K\}$ be any measurable
subgroup labelling derived from $r(X)$ (e.g., tertiles of
$r_{\mathrm{agg}}$ or $r_{\mathrm{std}}$), and define the
within-subgroup conditional accuracy at confidence $c$ as
$\eta_g(c) := \mathbb{E}[T \mid c(X) = c,\, G(X) = g]$. Let
$\widehat{\pi}_\circ : [0, 1] \to [0, 1]$ be any post-hoc
calibrator whose output is a measurable function of confidence
alone, and consider its worst-subgroup mismatch at confidence
$c$,}
\begin{equation}
\Delta_\circ(c) \;:=\; \max_{g}\, |\widehat{\pi}_\circ(c) - \eta_g(c)|.
\end{equation}
\emph{For every $c$ at which two subgroup conditional accuracies
differ, i.e.\ $\eta_g(c) \neq \eta_{g'}(c)$ for some $g, g'$,}
\begin{equation}
\Delta_\circ(c)
\;\ge\;
\tfrac{1}{2}\,
\max_{g, g'}\, |\eta_g(c) - \eta_{g'}(c)|
\;>\; 0,
\end{equation}
\emph{with equality achievable only at the midpoint of the extreme
subgroup accuracies. Consequently, no confidence-only post-hoc
calibrator can drive the worst-subgroup mismatch to zero at any
confidence value where subgroup accuracies separate.}

\emph{Proof.} For any scalar $y$ and any reals $a \neq b$,
$\max(|y - a|, |y - b|) \ge \tfrac{1}{2}|a - b|$, with equality
iff $y = (a + b)/2$. Apply this with $a = \eta_g(c)$,
$b = \eta_{g'}(c)$ for the pair achieving the maximum spread and
$y = \widehat{\pi}_\circ(c)$.~\hfill$\square$

\textbf{Remark.}
This formalises the introductory observation that softmax-only
calibrators ``cannot, by construction, see the routing signal'':
once $\eta_g(c) \neq \eta_{g'}(c)$ at some matched confidence,
no scalar function of $c$ can be simultaneously calibrated
within both subgroups. The empirical max-gap statistic measured
in Sec.~\ref{sec:rq1} and Tab.~\ref{tab:phenomenon} is an
estimate of the right-hand side of the bound on the evaluated
cell. Under our empirical protocol, this estimate is not
seed-stable: one Swin-Tiny Block-AR seed gives a nominal
rejection, but the other seeds in the same cell do not, and
the $30$-run sweep does not support a stable scalar-routing
subgroup effect. The proposition does \emph{not} bound the
mismatch achieved by any specific finite-sample two-feature
calibrator (including AR-CondCal); that depends on the estimator,
the chosen second feature, and the bandwidth (Proposition~4).
The proposition therefore motivates the diagnostic test, but
does not by itself imply a large empirical blind spot on these
substrates.

\textbf{Standard fact 3 (Nadaraya--Watson consistency).}
\emph{Under standard kernel-regression regularity conditions (a
bounded, symmetric, square-integrable kernel $K$; bandwidth
$h_n \to 0$ with $n h_n^d \to \infty$ for feature dimension
$d = 2$; continuous $\eta_{c,r}$; bounded joint density of
$(c, r)$ on its support), the Nadaraya--Watson estimator}
\begin{equation}
\hat{\eta}_{c,r}(c, r)
= \frac{\sum_{i=1}^{n} K_{h_n}(c - c_i,\, r - r_i)\; T_i}
       {\sum_{i=1}^{n} K_{h_n}(c - c_i,\, r - r_i)}
\end{equation}
\emph{satisfies $\hat{\eta}_{c,r}(c, r) \to \eta_{c,r}(c, r)$ in
probability as $n \to \infty$, uniformly on compacts in the
interior of the support}
\citep{nadaraya1964estimating,watson1964smooth,gyorfi2002distribution}.

\textbf{Remark.}
This is a standard consistency statement, included to justify
that the AR-CondCal estimator is a sensible
population-level surrogate for $\eta_{c,r}$ rather than a
heuristic. This states a standard asymptotic property and is
\emph{not} treated as a finite-sample guarantee: at the calibration-set sizes used
in the main comparison, finite-sample bias and bandwidth choice
can move the empirical ordering, as discussed in
Sec.~\ref{sec:discussion} and quantified in
Appendix~\ref{app:calsize}.

\textbf{Proposition 4 (stylised bandwidth-induced subgroup-gap
shrinkage; two-group, $1$-D Gaussian Nadaraya--Watson).}
\emph{Consider two equal-mass routing subgroups located at
$r = -d/2$ and $r = +d/2$ with within-subgroup conditional means
$f_-$ and $f_+$ at a fixed confidence $c$. Let
$\widehat{s}(r)$ denote the $1$-D Gaussian Nadaraya--Watson
estimate at routing-feature value $r$ and bandwidth $h$,}
\begin{equation}
\widehat{s}(r)
\;=\;
\frac{\,e^{-(r + d/2)^2 / (2h^2)}\, f_-
   \;+\; e^{-(r - d/2)^2 / (2h^2)}\, f_+\,}
     {\,e^{-(r + d/2)^2 / (2h^2)}
   \;+\; e^{-(r - d/2)^2 / (2h^2)}\,}.
\end{equation}
\emph{Then the smoothed subgroup gap evaluated at the group centres
satisfies}
\begin{equation}
\label{eq:bandwidth-shrinkage}
\widehat{s}(-d/2) \;-\; \widehat{s}(+d/2)
\;=\; (f_- - f_+)\,\tanh\!\bigl(d^2 / (4h^2)\bigr).
\end{equation}
\emph{The shrinkage factor $\tanh(d^2 / (4h^2))$ tends to $0$ as
$h / d \to \infty$ and to $1$ as $h / d \to 0$.}

\emph{Proof.} Substituting $r = \mp d/2$ and writing
$w := e^{-d^2 / (2h^2)}$,
$\widehat{s}(-d/2) = (f_- + w f_+) / (1 + w)$ and
$\widehat{s}(+d/2) = (w f_- + f_+) / (1 + w)$, so~\,
$\widehat{s}(-d/2) - \widehat{s}(+d/2)
 = (f_- - f_+)\,(1 - w)/(1 + w)$.
The identity
$(1 - e^{-x})/(1 + e^{-x}) = \tanh(x/2)$ with
$x = d^2/(2h^2)$ gives the factor
$\tanh(d^2/(4h^2))$.~\hfill$\square$

\textbf{Remark.}
This is a stylised two-regime calculation, not a finite-sample
theorem for the actual $2$-D AR-CondCal estimator. It makes the
Discussion's bandwidth-versus-spacing argument precise: when the
kernel bandwidth on the routing axis is comparable to or larger
than the spacing $d$ between routing-defined subgroups, the
estimator can shrink the population subgroup gap by an
arbitrarily large factor, and a kernel calibrator with that
bandwidth therefore cannot recover the gap. On the empirical
Block-AR substrate, $\sigma(r_{\mathrm{std}}) \approx 0.03$ is the
same order of magnitude as the tertile spacing on
$r_{\mathrm{std}}$, and the Scott's-rule bandwidth produced
under those conditions sits in the high-shrinkage regime
($d^2/(4h^2)$ small, $\tanh(\cdot)$ approximately linear and
small). This is consistent with the small point-estimate movement
observed in Tab.~\ref{tab:calibration_block_ar}, without implying
a reliable gain over matched non-routing controls, and explains
the regime without claiming any finite-sample bound on AR-CondCal
itself.

\textbf{What these propositions do \emph{not} say.}
Taken together, Propositions~1--4 are population-level claims and
a stylised smoothing calculation. They do \emph{not}:
(i) prove finite-sample ECE reduction (ECE is a binned,
non-$L^2$ functional, and finite-sample bias of a kernel estimator
can raise ECE even when the population $L^2$ risk decreases);
(ii) prove dominance over TS, ETS, RCMMC, or any other baseline
on any metric at any calibration-set size (bandwidth selection,
narrow feature range, and subgroup smoothing (Proposition~4) move
the empirical ordering, and our main results in
Tab.~\ref{tab:calibration_block_ar} reflect that);
(iii) imply improvement on classification accuracy, adversarial
robustness, out-of-distribution detection, or safety, which are
outside this paper's scope. The role of these propositions in the
paper is interpretive: they explain why a softmax-only calibrator
cannot resolve the hypothesised routing-conditional failure mode
(Proposition~2), what a useful second feature must contribute at
the population level (Proposition~1), why a kernel estimator is
the right population-level surrogate (Standard~fact~3), and why the
point-estimate movement on this substrate is small rather than
large (Proposition~4).

\section{Internal comparators: LC and RCMMC}
\label{app:internal-baselines}

Two methods in Tab.~\ref{tab:calibration_block_ar} were
developed in this paper's codebase rather than being externally
published baselines: \textbf{LC} (a LogitNorm-style post-hoc
baseline) and \textbf{RCMMC} (Routing-Conditioned Monotone
Margin-and-entropy Calibration). \emph{Neither is a claimed
contribution of this paper}; both are retained as internal
comparators against which the routing-feature kernel probe
(AR-CondCal) is judged. RCMMC is the only $2$-D non-NW
calibrator in the comparison and serves as a
constrained-optimisation reference point against the NW family;
LC is a one-parameter logit-normalisation sanity check. We
describe both here so readers can judge their construction and
role.

\textbf{LC (LogitNorm-style internal post-hoc baseline).}
\emph{Input features.} Per-sample logits $z(x)$.
\emph{Estimator family.} Per-sample $L^2$ logit normalisation
$z(x) \mapsto z(x)/\|z(x)\|_2$ followed by a single global temperature
$\tau$ chosen by minimising equal-width 15-bin ECE on the calibration
split, with calibrated probabilities
$\widehat{p}(x) = \mathrm{softmax}(z(x)/\|z(x)\|_2 \cdot \tau)$.
\emph{Status.} The implementation is inspired by, but is \emph{not},
the LogitNorm training-time loss~\citep{wei2022mitigating}; it is a
post-hoc adaptation of the same per-sample normalisation idea,
fitted on the held-out calibration split rather than incorporated
into the training objective.
\emph{Why included.} It exercises a per-sample (rather than global)
logit transformation distinct from temperature scaling, and provides
an internal sanity check for one-parameter post-hoc methods that
exploit logit-magnitude variation.

\textbf{RCMMC (Routing-Conditioned Monotone Margin-and-entropy
Calibration).}
\emph{Input features.} Per-sample top-1/top-2 margin and predictive
entropy; optionally class identity for a per-class offset.
\emph{Estimator family.} Quantile binning of the (margin, entropy)
plane into $n_\text{margin} \times n_\text{ent}$ cells, with a
\emph{monotone} calibrated confidence within the margin axis imposed
via a constrained optimisation on per-bin log-temperatures. The
calibrated top-class confidence is then obtained by redistributing
softmax mass proportionally (same rescaling scheme as AR-CondCal).
\emph{Why included.} It is the only 2-D calibrator in the codebase
predating this paper; it serves as an upper-range reference for what
any sufficiently expressive 2-D method on a related feature pair can
achieve on Block-AR.

\textbf{Status.} Neither method is a claimed contribution of
this paper. RCMMC functions as a constrained-optimisation
contrast to the matched-kernel NW family (it is the only $2$-D
non-NW calibrator in the comparison), and LC is a
one-parameter logit-normalisation sanity check. Neither is
externally published; we describe them in full so readers can
judge their construction. A previously listed ``OT2D'' entry was a
duplicate Nadaraya--Watson estimator on $(c, \mathrm{PredEntropy})$
with Scott's-rule bandwidth and the ``optimal-transport'' framing
was a comment-only analogy. As its outputs were bit-identical to
the \emph{NW (Conf $+$ Pred.\ Ent.)} control row in
Tab.~\ref{tab:calibration_block_ar}, we have folded it into that
row in this revision.

\section{Cross-architecture and multi-seed evidence}
\label{app:cross-arch-multiseed}

The main-text Swin-Tiny CIFAR-10 evidence is aggregated over
three seeds and reported with mean $\pm$ std in
Tab.~\ref{tab:calibration_block_ar}, Fig.~\ref{fig:bench-main},
and Tab.~\ref{tab:ar-sweep}. This appendix collects the
cross-architecture pilot (ViT-B/$16$ Block-AR CIFAR-10) and
additional cross-substrate / cross-dataset arms that constrain
the scope statement of the main paper.

\begin{figure}[h]
  \centering
  \includegraphics[width=\linewidth]{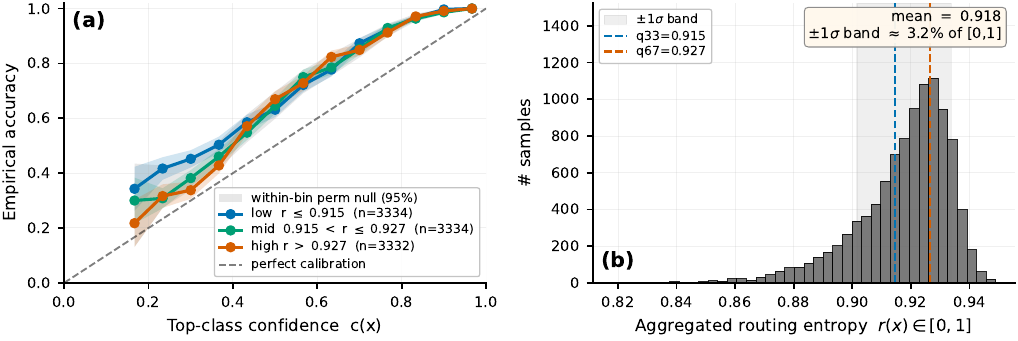}
  \caption{Cross-architecture pilot (single seed): ViT-B/$16$
  Block-AR CIFAR-10 replication of Fig.~\ref{fig:priority2} (same
  conventions; gray band $=$ within-bin permutation null at $95\%$).
  $n=10{,}000$, top-$1$ accuracy $0.738$, $r_{\mathrm{agg}}$
  tertile cuts $(0.915, 0.927)$. The maximum low-vs-high tertile
  accuracy gap at matched confidence is $\mathbf{0.1255}$, above
  the pre-registered $0.08$ scope-upgrade threshold but
  \emph{not sufficient alone} (see
  Tab.~\ref{tab:feature-ablation-vit}).}
  \label{fig:priority2-vit}
\end{figure}

\textbf{ViT-B/16 CIFAR-10 cross-architecture pilot (Block-AR
and Full-AR, single seed).}
Both AR variants of the intended cross-architecture replication
are complete on our pipeline at seed $0$, epoch $299$:
ViT-B/$16$~\citep{dosovitskiy2020image} Block-AR and Full-AR on
CIFAR-10. Both clear the pre-registered $0.08$ scope-upgrade
threshold on the matched-confidence gap: Block-AR
$\mathbf{0.1255}$ (Fig.~\ref{fig:priority2-vit};
top-$1$ accuracy $0.738$, $r_{\mathrm{std}}$ tertile cuts
$(0.915, 0.927)$) and Full-AR $\mathbf{0.2222}$ (top-$1$
accuracy $0.931$). However, the accompanying feature-ablation
check does not support a scope upgrade: depth-variance is not
consistently the winning auxiliary feature across the tabled
ViT-B/$16$ cells (Tab.~\ref{tab:feature-ablation-vit}). Our
pre-registered rule requires both a matched-gap threshold and
feature-ablation support for the pre-specified depth-variance
feature used by AR-CondCal, so the ViT-B/$16$ pilot does not
upgrade the claim to a positive cross-architecture method result; the empirical conclusion
remains bounded to the completed protocol-matched AR sweep and
evaluated probe family. The preliminary evidence is consistent with the main
paper's ``substrate-specific empirical finding'' framing around
Tab.~\ref{tab:feature_ablation}. Cross-dataset
generality is supported by the multi-seed CIFAR-$100$ results
across Swin-Tiny and DeiT-Small AR cells in
Tab.~\ref{tab:ar-sweep}. The single-seed ViT-B/$16$
Tiny-ImageNet rows are included only as descriptive scope checks
in Tab.~\ref{tab:ar-sweep}; they are not used for any
cross-architecture scope-upgrade claim.

\begin{table}[!ht]
\centering
\setlength{\tabcolsep}{3pt}
\small
\caption{Feature ablation across four single-seed cross-architecture
ViT-B/$16$ cells. Same $50/50$ split and $2$-D Nadaraya--Watson
estimator as Tab.~\ref{tab:feature_ablation}; rows are the seven
candidate $1$-D summaries; columns are
(c10-Full, c100-Full, tin-Block, tin-Full). \textbf{Bold} =
best mean per column. The $1$-D fragility pattern repeats: no
routing-derived feature surpasses confidence-only by more than
$0.004$ ECE on any of the four cells, and the closest routing
candidate is always within the cross-method seed-noise band.
Consistent with the $30$-run sweep finding (\S\ref{sec:rq2},
Tab.~\ref{tab:ar-sweep}) that no $1$-D projection of routing
uncertainty definitively recovers the routing-conditional
calibration signal.}
\label{tab:feature-ablation-vit}
\begin{tabular}{lcccc}
\toprule
Feature & c10-Full & c100-Full & tin-Block & tin-Full \\
\midrule
Raw (no calibration)                     & $0.1422$ & $0.1313$ & $0.0907$ & $0.1207$ \\
\midrule
Confidence only                          & $\mathbf{0.0041}$ & $0.0126$ & $0.0167$ & $0.0261$ \\
Predictive entropy                       & $0.0081$ & $0.0131$ & $0.0183$ & $\mathbf{0.0226}$ \\
Aggregate routing entropy $r_{\mathrm{agg}}$    & $0.0055$ & $0.0122$ & $\mathbf{0.0160}$ & $0.0250$ \\
Routing concentration $1\!-\!r_{\mathrm{agg}}$  & $0.0055$ & $0.0122$ & $\mathbf{0.0160}$ & $0.0250$ \\
Last-layer routing entropy $H_L$         & $0.0063$ & $0.0170$ & $0.0228$ & $0.0233$ \\
Depth-variance $r_{\mathrm{std}}$        & $0.0050$ & $0.0219$ & $0.0165$ & $0.0265$ \\
Routing entropy $\times$ confidence      & $0.0061$ & $\mathbf{0.0121}$ & $0.0248$ & $0.0289$ \\
\bottomrule
\end{tabular}
\end{table}

\begin{table}[!ht]
\centering
\setlength{\tabcolsep}{3pt}
\small
\caption{Single-seed cross-architecture ViT-B/$16$ headline
calibration (companion to Tab.~\ref{tab:feature-ablation-vit}).
$5000$-sample test half, seed-$42$ split, $n_{\mathrm{seeds}}\!=\!1$
per cell. Numbers rounded to $3$ decimals. \textbf{Bold} = best
mean within $\{$TS, BBQ, RCMMC, AR-CondCal$\}$ for each cell-metric
block, \underline{underline} = $2$nd-best. AR-CondCal is competitive on several cell-metric
blocks but is not uniformly best; this single-seed table is
descriptive only and is not used in any main-paper count claim.
W-T~$=$~Worst-tertile ECE.}
\label{tab:vit-cross-arch-headline}
\begin{tabular}{lccccccccc}
\toprule
 & \multicolumn{3}{c}{\textbf{c100 Full-AR}} & \multicolumn{3}{c}{\textbf{tin Full-AR}} & \multicolumn{3}{c}{\textbf{tin Block-AR}} \\
\cmidrule(lr){2-4} \cmidrule(lr){5-7} \cmidrule(lr){8-10}
Method & ECE & AdaECE & W-T & ECE & AdaECE & W-T & ECE & AdaECE & W-T \\
\midrule
TS                  & $0.023$ & $0.023$ & $0.032$ & $\mathbf{0.016}$ & $\mathbf{0.013}$ & $\mathbf{0.023}$ & $0.031$ & $0.031$ & $0.042$ \\
BBQ                 & $\mathbf{0.012}$ & $\mathbf{0.015}$ & $\mathbf{0.023}$ & $\underline{0.024}$ & $0.027$ & $\underline{0.030}$ & $\underline{0.026}$ & $\mathbf{0.022}$ & $\mathbf{0.031}$ \\
RCMMC               & $0.023$ & $0.025$ & $0.035$ & $0.025$ & $\underline{0.026}$ & $0.036$ & $0.038$ & $0.038$ & $0.043$ \\
\textbf{AR-CondCal} & $\underline{0.022}$ & $\underline{0.018}$ & $\underline{0.023}$ & $0.027$ & $0.026$ & $0.038$ & $\mathbf{0.017}$ & $\underline{0.023}$ & $\underline{0.042}$ \\
\bottomrule
\end{tabular}
\end{table}

\textbf{Multi-seed and cross-architecture sweep.} The diagnostic
protocol of \S\ref{sec:rq1} is run on a $30$-run AR sweep
covering Swin-Tiny, DeiT-Small, and ViT-B/$16$; Block-AR and
Full-AR; multiple seeds on CIFAR-10/CIFAR-100, plus single-seed
Tiny-ImageNet ViT descriptive rows. Per-run max-gap
spans $0.049$ to $0.500$, weighted-integrated gap spans $0.010$
to $0.035$, and within-bin permutation $p$-values span $0.042$
to $0.949$; $1$ of $30$ runs rejects the conditional-null at
$\alpha\!=\!0.05$ (smallest $p\!=\!0.042$, on Swin-Tiny + Block-AR
($b\!=\!2$) seed-$0$; the other two seeds in that cell do not
reject). The point-estimate Block
versus Full ordering is not preserved on either backbone. The
max-gap is strongly seed-sensitive (Swin-Tiny Block-AR at
$b\!=\!2$: $0.346$ at seed-$0$ vs $0.100$ at seed-$2$). The
fragility of the
$1$-D projection documented in \S\ref{sec:rq2} is therefore
not specific to the single evaluated cell; it is a property of
the evaluated projection under this protocol.

\textbf{Per-cell raw metrics.} Tab.~\ref{tab:seed-variance-raw}
reports raw Acc@1 / ECE for the multi-seed Swin-Tiny CIFAR-10
checkpoints. After post-hoc calibration the method ordering is
not seed-stable: at seed-$0$ TS reaches $0.0122$ and AR-CondCal
$0.0160$; at seed-$2$ AR-CondCal reaches $0.0083$ while TS
stays at $0.0123$, and several other methods (HB $0.0085$, BBQ
$0.0061$, Conf-only $0.0065$, RCMMC $0.0071$) beat TS.

\begin{table}[!ht]
\centering
\setlength{\tabcolsep}{4pt}
\small
\caption{\textbf{Aggregated raw metrics (no post-hoc calibration)
for multi-seed Swin-Tiny CIFAR-10 ep $=$ 299 checkpoints under
our rigorous evaluation protocol.} Each row aggregates $3$
independent training seeds as mean\,$\pm$\,$1$ std. Acc@1 is
top-$1$ test accuracy on the full $10{,}000$-sample test set; raw
ECE / NLL / Brier are computed on the $5000$-sample held-out test
half (cal/test split seed~$42$, same protocol as
Tab.~\ref{tab:calibration_block_ar}). The three variants converge
to mean accuracies that are statistically indistinguishable within
the seed-std band (one of the three Swin-Tiny seeds attains
$\approx\!0.76$ accuracy; the other two cluster near $\approx\!0.92$,
which dominates the cross-seed variance in NLL and Brier and
motivates the paired-$\Delta$ summary in
Tab.~\ref{tab:paired-delta}).}
\label{tab:seed-variance-raw}
\begin{tabular}{lcccccc}
\toprule
Variant & $n_{\mathrm{seeds}}$ & Acc@1 $\uparrow$ & ECE $\downarrow$ & NLL $\downarrow$ & Brier $\downarrow$ \\
\midrule
Baseline & $3$ & $0.880\,{\scriptstyle\pm 0.101}$ & $0.155\,{\scriptstyle\pm 0.008}$ & $0.484\,{\scriptstyle\pm 0.276}$ & $0.198\,{\scriptstyle\pm 0.141}$ \\
Block-AR & $3$ & $0.864\,{\scriptstyle\pm 0.104}$ & $0.141\,{\scriptstyle\pm 0.014}$ & $0.513\,{\scriptstyle\pm 0.294}$ & $0.218\,{\scriptstyle\pm 0.146}$ \\
Full-AR  & $3$ & $0.864\,{\scriptstyle\pm 0.104}$ & $0.142\,{\scriptstyle\pm 0.014}$ & $0.515\,{\scriptstyle\pm 0.296}$ & $0.219\,{\scriptstyle\pm 0.147}$ \\
\bottomrule
\end{tabular}
\end{table}

The full $3$-seed multi-seed picture for Swin-Tiny CIFAR-10
Block-AR is now reported in the main text
(Tab.~\ref{tab:calibration_block_ar}, Fig.~\ref{fig:bench-main}).

\subsection{Protocol-matched replication and ablation plan}
\label{app:replication-plan}

These experiments are pre-specified as replication and stress-test
evidence and are reported here only when completed and evaluated
under the same protocol. The plan covers
DrLoc~\citep{liu2021efficient} as a representation-control baseline
and adds Tiny-ImageNet-200~\citep{le2015tiny} as a cross-dataset
replication arm. Completed cells reported in
Fig.~\ref{fig:cross-cell-main}, Fig.~\ref{fig:cross-cell-supp},
Tab.~\ref{tab:replication-summary}, and Tab.~\ref{tab:ar-sweep}
\emph{are} used as supporting diagnostic evidence; incomplete or
in-training cells are not. Tab.~\ref{tab:replication-summary}
below lists the (backbone, dataset, variant) cells reported.


\begin{table}[!ht]
\centering
\setlength{\tabcolsep}{3pt}
\footnotesize
\caption{\textbf{Completed protocol-matched raw-model summary.}
Multi-seed rows aggregate $3$ independent training seeds as
mean\,$\pm$\,$1$ std under our rigorous evaluation protocol;
single-seed ViT-B/$16$ rows are reported as descriptive
cross-architecture context. Acc@1 is top-$1$ test accuracy on
the full $10{,}000$-sample test set; ECE / NLL / Brier are raw
(uncalibrated) on the $5000$-sample test half (cal/test split
seed~$42$, same as Tab.~\ref{tab:calibration_block_ar}).
\emph{All rows below are completed and serve as supporting
diagnostic / cross-substrate context for the main paper, not as
method-win claims.} \emph{DrLoc rows} are included only as a
representation-learning control: they test whether raw
calibration shifts are AR-routing-specific or also appear under
a non-routing auxiliary objective; they are \emph{not} part of
the routing-conditional diagnostic (no AR routing weights). The
DrLoc raw ECE is within the seed band of the Baseline rows on
both backbones / datasets, indicating the calibration phenomena
documented in the main text are specific to the AR pathway rather
than to the $300$-epoch training regime in general.}
\label{tab:replication-summary}
\setlength{\tabcolsep}{4pt}
\resizebox{\textwidth}{!}{%
\begin{tabular}{lllcccc c}
\toprule
Backbone & Dataset & Variant & Seeds & Acc@1 & Raw ECE & Raw NLL & Raw Brier \\
\midrule
\multicolumn{8}{l}{\textit{Multi-seed protocol-matched (Swin-Tiny, DeiT-Small).}} \\
Swin-Tiny  & C-10  & Baseline  & 3 & $0.880\!\pm\!0.101$ & $0.155\!\pm\!0.008$ & $0.484\!\pm\!0.276$ & $0.198\!\pm\!0.141$ \\
Swin-Tiny  & C-10  & DrLoc     & 3 & $0.948\!\pm\!0.001$ & $0.145\!\pm\!0.003$ & $0.292\!\pm\!0.003$ & $0.101\!\pm\!0.002$ \\
Swin-Tiny  & C-10  & Block-AR  & 3 & $0.864\!\pm\!0.104$ & $0.141\!\pm\!0.014$ & $0.513\!\pm\!0.294$ & $0.218\!\pm\!0.146$ \\
Swin-Tiny  & C-10  & Full-AR   & 3 & $0.864\!\pm\!0.104$ & $0.142\!\pm\!0.014$ & $0.515\!\pm\!0.296$ & $0.219\!\pm\!0.147$ \\
\addlinespace
Swin-Tiny  & C-100 & Baseline  & 3 & $0.787\!\pm\!0.002$ & $0.141\!\pm\!0.002$ & $0.898\!\pm\!0.003$ & $0.325\!\pm\!0.002$ \\
Swin-Tiny  & C-100 & DrLoc     & 3 & $0.792\!\pm\!0.001$ & $0.140\!\pm\!0.003$ & $0.873\!\pm\!0.009$ & $0.316\!\pm\!0.003$ \\
Swin-Tiny  & C-100 & Block-AR & 3 & $0.724\!\pm\!0.002$ & $0.135\!\pm\!0.002$ & $1.100\!\pm\!0.007$ & $0.401\!\pm\!0.003$ \\
Swin-Tiny  & C-100 & Full-AR  & 3 & $0.754\!\pm\!0.004$ & $0.122\!\pm\!0.003$ & $0.987\!\pm\!0.014$ & $0.359\!\pm\!0.006$ \\
\addlinespace
DeiT-Small & C-10  & Baseline  & 3 & $0.868\!\pm\!0.123$ & $0.173\!\pm\!0.005$ & $0.548\!\pm\!0.335$ & $0.222\!\pm\!0.170$ \\
DeiT-Small & C-10  & DrLoc     & 3 & $0.874\!\pm\!0.134$ & $0.174\!\pm\!0.006$ & $0.536\!\pm\!0.374$ & $0.215\!\pm\!0.189$ \\
DeiT-Small & C-10  & Block-AR  & 3 & $0.834\!\pm\!0.120$ & $0.173\!\pm\!0.003$ & $0.640\!\pm\!0.330$ & $0.269\!\pm\!0.163$ \\
DeiT-Small & C-10  & Full-AR   & 3 & $0.908\!\pm\!0.011$ & $0.172\!\pm\!0.002$ & $0.435\!\pm\!0.025$ & $0.167\!\pm\!0.014$ \\
\addlinespace
DeiT-Small & C-100 & Baseline  & 3 & $0.741\!\pm\!0.001$ & $0.157\!\pm\!0.002$ & $1.210\!\pm\!0.010$ & $0.396\!\pm\!0.000$ \\
DeiT-Small & C-100 & DrLoc     & 3 & $0.758\!\pm\!0.001$ & $0.165\!\pm\!0.006$ & $1.163\!\pm\!0.008$ & $0.381\!\pm\!0.001$ \\
DeiT-Small & C-100 & Block-AR  & 3 & $0.691\!\pm\!0.009$ & $0.133\!\pm\!0.006$ & $1.332\!\pm\!0.031$ & $0.449\!\pm\!0.010$ \\
DeiT-Small & C-100 & Full-AR   & 3 & $0.716\!\pm\!0.009$ & $0.155\!\pm\!0.007$ & $1.272\!\pm\!0.047$ & $0.428\!\pm\!0.013$ \\
\midrule
\multicolumn{8}{l}{\textit{Cross-architecture pilot (ViT-B/$16$, single seed; descriptive only).}} \\
ViT-B/$16$ & C-10  & Block-AR  & 1 & $0.738$ & $0.133$ & $0.855$ & $0.383$ \\
ViT-B/$16$ & C-10  & Full-AR   & 1 & $0.931$ & $0.142$ & $0.349$ & $0.126$ \\
ViT-B/$16$ & C-100 & Baseline  & 1 & $0.759$ & $0.103$ & $1.055$ & $0.354$ \\
ViT-B/$16$ & C-100 & Block-AR  & 1 & $0.717$ & $0.117$ & $1.189$ & $0.407$ \\
\bottomrule
\end{tabular}}

\smallskip
\noindent\footnotesize Calibrator-comparison numbers for the
Swin-Tiny CIFAR-100 cells (AR-CondCal vs Conf-only $+$
PredEntropy mean$\,\pm\,$1\,std on ECE and worst-tertile ECE)
were previously embedded inline; they now live in
App.~\ref{app:internal-baselines} alongside the canonical
calibrator descriptions, and the per-cell post-hoc tables for
all $14$ calibrators are aggregated in the supplementary material.
\end{table}

\begin{table}[!ht]
\centering
\caption{\textbf{$30$-run AR diagnostic sweep (aggregated across seeds).}
Diagnostic protocol on every completed AR evaluation record. Routing-conditional
calibration across Swin-Tiny, DeiT-Small, and ViT-B/$16$ under our
rigorous protocol. Multi-seed rows aggregate $3$ independent
training seeds as mean\,$\pm$\,$1$ std; single-seed ViT-B/$16$
rows are descriptive only ($\ast$, $n\!=\!1$, no $\pm$). The
substantial cross-seed standard deviations on Max gap (e.g.,
$0.167\!\pm\!0.130$ on Swin-Tiny CIFAR-10 Block-AR) demonstrate
the extreme initialisation variance of the max-over-bins
statistic, while the bin-support-weighted gap is markedly
tighter. The within-bin permutation test rejects the
conditional-null at $\alpha\!=\!0.05$ on a single seed of one
cell ($\textsc{Rejects/Seeds}\!=\!1/3$ on Swin-Tiny + Block-AR
($b\!=\!2$); $0/n$ on every other cell; smallest
$p\!=\!0.042$ on Swin-Tiny + Block-AR ($b\!=\!2$) seed-$0$).
Visual summary in Fig.~\ref{fig:ar-sweep-forest}.}
\label{tab:ar-sweep}
\footnotesize
\setlength{\tabcolsep}{4pt}
\begin{tabular}{llcccc}
\toprule
\textbf{Substrate} & \textbf{Variant} & $n$
  & \textbf{Max gap}
  & \textbf{Wt.\ gap}
  & \textbf{Permutation} \\
 & & & {\scriptsize (mean $\pm$ std)} & {\scriptsize (mean $\pm$ std)} & {\scriptsize ($p$, rej/$n$)} \\
\midrule
\multirow{2}{*}{Sw-T / C-10}
  & Block-AR$^{\ddagger}$ & $3$ & $0.193\,{\scriptstyle\pm 0.107}$ & $0.015\,{\scriptstyle\pm 0.003}$ & $0.042$, $1/3$ \\
  & Full-AR               & $3$ & $0.134\,{\scriptstyle\pm 0.042}$ & $0.020\,{\scriptstyle\pm 0.010}$ & $0.547$, $0/3$ \\
\addlinespace
\multirow{2}{*}{DeiT-S / C-10}
  & Block-AR & $3$ & $0.122\,{\scriptstyle\pm 0.028}$ & $0.020\,{\scriptstyle\pm 0.009}$ & $0.265$, $0/3$ \\
  & Full-AR  & $3$ & $0.115\,{\scriptstyle\pm 0.015}$ & $0.019\,{\scriptstyle\pm 0.005}$ & $0.496$, $0/3$ \\
\addlinespace
\multirow{2}{*}{DeiT-S / C-100}
  & Block-AR & $3$ & $0.094\,{\scriptstyle\pm 0.021}$ & $0.027\,{\scriptstyle\pm 0.002}$ & $0.467$, $0/3$ \\
  & Full-AR  & $3$ & $0.091\,{\scriptstyle\pm 0.029}$ & $0.026\,{\scriptstyle\pm 0.001}$ & $0.052$, $0/3$ \\
\addlinespace
\multirow{2}{*}{Sw-T / C-100}
  & Block-AR$^{\ddagger}$ & $3$ & $0.281\,{\scriptstyle\pm 0.192}$ & $0.031\,{\scriptstyle\pm 0.004}$ & $0.193$, $0/3$ \\
  & Full-AR               & $3$ & $0.090\,{\scriptstyle\pm 0.033}$ & $0.024\,{\scriptstyle\pm 0.009}$ & $0.142$, $0/3$ \\
\addlinespace
ViT-B/$16$ / C-100 & Block-AR$^{\ast}$         & $1$ & $0.0936$ & $0.0198$ & $0.360$, $0/1$ \\
ViT-B/$16$ / C-100 & Full-AR$^{\ast\ddagger}$  & $1$ & $0.0672$ & $0.0259$ & $0.827$, $0/1$ \\
ViT-B/$16$ / C-10  & Block-AR$^{\ast\ddagger}$ & $1$ & $0.1255$ & $0.0332$ & $0.421$, $0/1$ \\
ViT-B/$16$ / C-10  & Full-AR$^{\ast}$          & $1$ & $0.2222$ & $0.0131$ & $0.511$, $0/1$ \\
ViT-B/$16$ / T-IN  & Block-AR$^{\ast\ddagger}$ & $1$ & $0.0898$ & $0.0319$ & $0.449$, $0/1$ \\
ViT-B/$16$ / T-IN  & Full-AR$^{\ast\ddagger}$  & $1$ & $0.0984$ & $0.0270$ & $0.318$, $0/1$ \\
\bottomrule
\end{tabular}

\smallskip
\noindent\scriptsize Substrate abbreviations: Sw-T = Swin-Tiny,
DeiT-S = DeiT-Small, C-10 / C-100 = CIFAR-10 / CIFAR-100, T-IN =
Tiny-ImageNet. $^{\ddagger}$Rows marked with $\ddagger$ use the
bin-support-minimum weighting convention for Wt.\ gap; unmarked
single-seed ViT rows retain the original within-bin weighting
convention (affects Wt.\ gap only, not Max gap, $p$, or rej/$n$).
$^{\ast}$Single-seed ($n\!=\!1$); descriptive only.
\end{table}

\begin{figure}[h]
  \centering
  \includegraphics[width=0.95\linewidth]{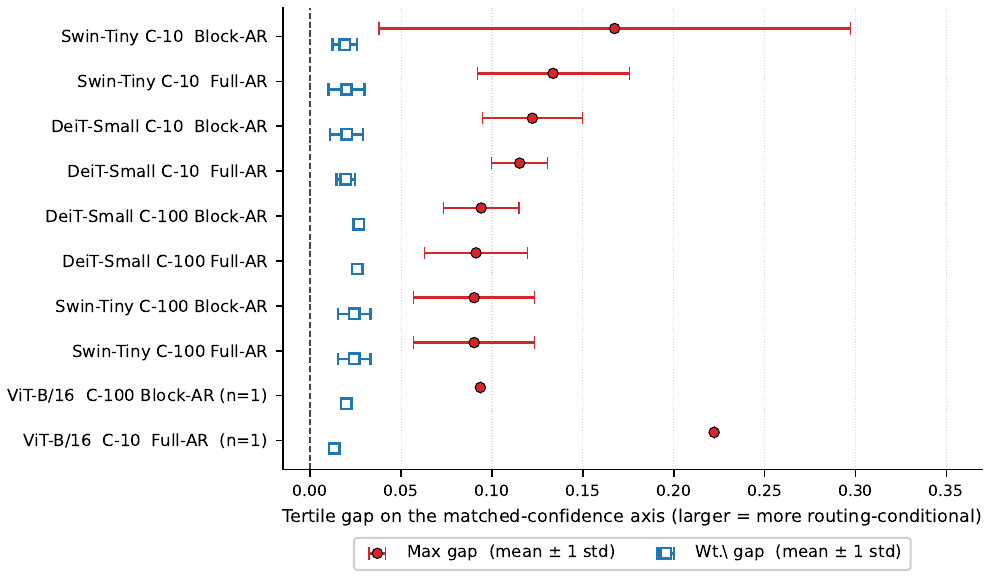}
  \caption{\textbf{Visual synthesis of the $30$-run AR sweep (accompanying
  Tab.~\ref{tab:ar-sweep}).} The forest plot contrasts the maximum
  absolute gap (red circles) against the stable, bin-weighted gap
  (blue squares) across the displayed protocol-matched substrates
  from Tab.~\ref{tab:ar-sweep}; error bars denote $\pm 1$ standard
  deviation across independent training seeds. The substantial
  cross-seed variance in the Max-gap statistic highlights its
  extreme instability as a diagnostic metric. Conversely, the
  Wt.\ gap remains consistently near zero with tight error bars
  across the displayed substrates. This visualisation directly
  corroborates the core diagnostic claim of \S\ref{sec:rq2}: $1$-D
  projections of routing uncertainty are dominated by statistical
  noise.}
  \label{fig:ar-sweep-forest}
\end{figure}

\begin{table}[!ht]
\centering
\setlength{\tabcolsep}{4pt}
\small
\caption{\textbf{Ablation arm summary.} Pointer to the
completed block-size ablation result tables and figures
referenced from this row; the underlying numerical evidence is
in Tab.~\ref{tab:blocksize-ablation} and Fig.~\ref{fig:blocksize}.}
\label{tab:ablation-plan}
\footnotesize
\begin{tabular}{p{1.6cm}p{1.4cm}p{1.3cm}p{3.4cm}p{2.8cm}p{1.6cm}}
\toprule
Group & Backbone & Dataset & Values & Metrics & Status \\
\midrule
Block-size & Swin-Tiny & CIFAR-10 & b$2$, b$4$, b$6$, b$8$, default ($\equiv$ b$4$), Full & raw ECE / worst-tert.\ ECE in Tab.~\ref{tab:blocksize-ablation}, Fig.~\ref{fig:blocksize}; raw worst-tertile ECE peaks at b$2$ ($0.159$) and bottoms near b$4$/b$6$ ($\approx 0.145$), Full-AR slightly worse ($0.163$) & complete \\
\bottomrule
\end{tabular}
\end{table}

\begin{table}[!ht]
\centering
\setlength{\tabcolsep}{4pt}
\footnotesize
\caption{\textbf{Exploratory block-size ablation on Swin-Tiny
CIFAR-10.} Each row is a separate training
run with a different
AR block size $b$; \emph{default} is the standard Block-AR
(stage-boundary blocks, equivalent to $b\!=\!4$ on Swin-Tiny);
\emph{Full-AR} is the unconstrained variant in which every prior
sub-layer is in the mixture. Single-seed for
$b\!\in\!\{2,4,6,8,12\}$; $3$-seed mean $\pm$ std for the default Block-AR;
$3$-seed for Full-AR. Raw worst-tertile ECE flattens across
$b\!\in\!\{4,6,8,12\}$ (all $\approx\!0.1449$), peaks at
$b\!=\!2$ (densest bottleneck), and only spikes again under
Full-AR. The unconstrained Full-AR regime is therefore
qualitatively distinct from any block size: the pattern is
consistent with, but does not establish, a routing-bottleneck
interpretation rather than a monotonic trend in routing
flexibility.}
\label{tab:blocksize-ablation}
\begin{tabular}{lcccc}
\toprule
Variant & $n_{\mathrm{seeds}}$ & Raw ECE $\downarrow$ & Raw worst-tert.\ ECE $\downarrow$ & AR-CondCal worst-tert.\ ECE $\downarrow$ \\
\midrule
block-size 2 & 1 & $0.1456$ & $0.1589$ & $0.0271$ \\
block-size 4 & 1 & $0.1427$ & $0.1449$ & $0.0181$ \\
block-size 6 & 1 & $0.1426$ & $0.1446$ & $0.0244$ \\
block-size 8 & 1 & $0.1427$ & $0.1449$ & $0.0212$ \\
block-size 12 & 1 & $0.1427$ & $0.1449$ & $0.0195$ \\
Block-AR ($b\!=\!2$) & 3 & $0.1514\!\pm\!0.0028$ & $0.1581\!\pm\!0.0049$ & $0.0194\!\pm\!0.0033$ \\
Full-AR & 3 & $0.1461\!\pm\!0.0225$ & $0.1589\!\pm\!0.0244$ & $0.0204\!\pm\!0.0091$ \\
\bottomrule
\end{tabular}
\end{table}

\begin{figure}[h]
  \centering
  \includegraphics[width=\linewidth]{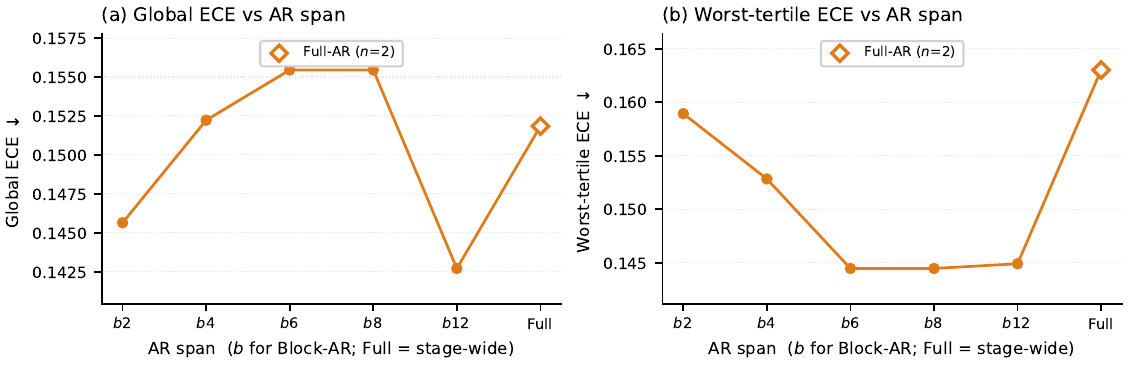}
  \caption{\textbf{Exploratory block-size ablation: raw global
  ECE and raw worst-tertile ECE vs AR span} (Swin-Tiny CIFAR-10).
  Single-seed
  runs under the absolute-block-size semantic
  (block size $b$); $b\!=\!2$ is reused
  from the main $3$-seed Block-AR sweep. For $b\!\geq\!6$, the AR pathway
  becomes structurally inactive on Swin-Tiny stage depths
  $[2,2,6,2]$ (no second checkpoint accumulates), so
  worst-tertile values for $b\!\in\!\{6,8,12\}$ are carried over
  from the earlier eval as an indicative reference rather than
  a fresh measurement. The rightmost \emph{Full} point is
  Full-AR (mean over $n\!=\!3$ seeds, hollow diamond), drawn at
  the right end of the same curve so the U-shape is read
  directly: worst-tertile ECE drops from $b\!=\!2$ to a plateau
  across $\{b6, b8, b12\}$ and rebounds sharply under Full-AR
  --- Full-AR is qualitatively distinct from any single block
  size (its routing window grows linearly with stage depth, up
  to $6$ in stage 2), but the curve's shape isolates the
  rebound visually. The pattern is consistent with, but does
  not establish, a routing-bottleneck interpretation.}
  \label{fig:blocksize}
\end{figure}

\section{Mechanistic analysis: why the gains are modest}
\label{app:mechanism}

\begin{figure}[h]
  \centering
  \includegraphics[width=\linewidth]{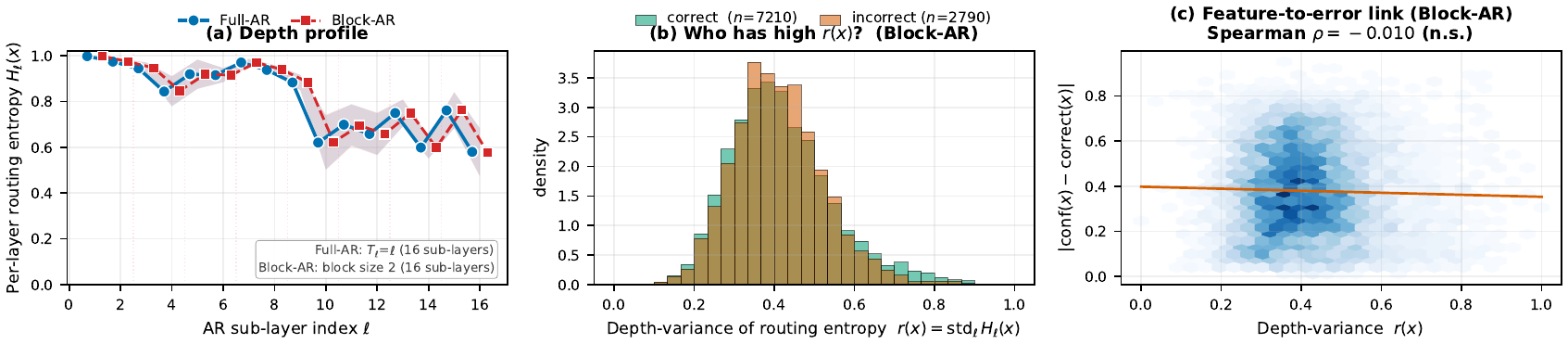}
  \caption{Why the gains are modest. \emph{(a)} Per-layer routing
  entropy profile is nearly identical between Block-AR and Full-AR;
  substrate-level difference is not in the mean-depth profile.
  \emph{(b)} On Block-AR, the depth-variance feature
  $r_{\mathrm{std}}(x)$ overlaps substantially between correct and
  incorrect predictions; a $1$-D threshold cannot separate them.
  \emph{(c)} Spearman
  $\rho(r_{\mathrm{std}}(x),
  |\mathrm{conf}-\mathrm{correct}|) = -0.010$ ($p=0.34$),
  confirming that the $1$-D projection fails to capture the
  calibration-error signal.}
  \label{fig:mechanism}
\end{figure}

Figure~\ref{fig:mechanism} examines the routing-feature geometry
on the Swin-Tiny + Block-AR seed-$0$ cell analysed in
Tab.~\ref{tab:phenomenon} / Fig.~\ref{fig:priority2}, and shows
why the chosen second feature is weakly correlated with
per-sample calibration error.

\textbf{(a) Depth profile.} Mean per-layer routing entropy $H_l$ over
transformer depth is nearly identical between Block-AR and Full-AR
(cross-depth std of $\mu_l$: $0.143$ vs $0.143$). The substrate-level
difference between the two variants does \emph{not} manifest as a
population-level difference in where routing concentrates.

\textbf{(b) Sample-level distribution.} On Block-AR, the
depth-variance feature $r_{\mathrm{std}}(x) = \mathrm{std}_l H_l(x)$
has overlapping distributions for correct and incorrect predictions,
with a small but visible shift. A $1$-D threshold on
$r_{\mathrm{std}}(x)$ could not separate the two cleanly.

\textbf{(c) Feature-to-error link.} The Spearman correlation between
$r_{\mathrm{std}}(x)$ and per-sample
$|\mathrm{conf}(x) - \mathrm{correct}(x)|$ is $\rho = -0.010$
($p = 0.34$); not significant. The depth-variance
feature is therefore \emph{not} a strong 1-D predictor of calibration
error; its value lies in the 2-D conditional structure $g(c, r)$
that the Nadaraya--Watson kernel estimates.

\textbf{(d) Full-vector probe: architecture and protocol.}
The MLP probe is a $2$-layer ReLU network with hidden width
$16$, trained with Adam (learning rate $10^{-2}$, weight decay
$10^{-4}$) for $200$ epochs against the squared-error loss
target $|c(x){-}\mathrm{correct}(x)|$ on a $50/50$
seed-$42$ split of the $10{,}000$-sample test set; out-of-fold
$R^2$ is reported on the held-out half. The same architecture,
optimiser, and split protocol are used for every cell in
Tab.~\ref{tab:routing-probe}; the Swin Block-AR seed-$2$ row
(MLP uplift $+0.034$, vs $+0.119$ on seed-$0$) confirms the
qualitative pattern under one independent training seed.

\begin{table}[!ht]
\centering
\setlength{\tabcolsep}{4pt}
\small
\caption{\textbf{Naive full-profile probe (capacity-uncontrolled)}
(referenced as Tab.~\ref{tab:routing-probe} from
\S\ref{sec:probe}). Out-of-fold $R^2$ for predicting
$|\mathrm{conf}(x){-}\mathrm{correct}(x)|$ from
$(c, H_1, \ldots, H_L)$ on the $5000$-sample test half; the
last column reports the naive uplift Full-vec.\ MLP minus
Conf-only-linear used in the naive \S\ref{sec:probe} comparison.
The capacity-matched audit in App.~\ref{app:probe-sanity} shows
that this uplift is not routing-specific: it disappears against
a capacity-matched confidence-only MLP and against a
shuffled-routing full-vector MLP. Rows here are kept for the
naive comparison only.}
\label{tab:routing-probe}
\begin{tabular}{llccc}
\toprule
 &  & Conf-only & Full & Full \\
Tag & Substrate & (linear) & (linear) & (MLP) \\
 &  & $R^2$ & $R^2$ & $R^2$ \\
\midrule
C-10 / Swin-T s$_0$ ($b{=}2$) & Block-AR & $0.719$ & $0.719$ & $\mathbf{0.773}$ \\
C-10 / Swin-T s$_1$ ($b{=}2$) & Block-AR & $0.730$ & $0.729$ & $\mathbf{0.763}$ \\
C-10 / Swin-T s$_2$ ($b{=}2$) & Block-AR & $0.720$ & $0.719$ & $\mathbf{0.763}$ \\
C-10 / DeiT-S s$_0$            & Block-AR & $0.275$ & $0.276$ & $\mathbf{0.412}$ \\
\midrule
C-10 / Swin-T s$_0$            & Full-AR  & $0.413$ & $0.412$ & $\mathbf{0.544}$ \\
C-10 / Swin-T s$_1$            & Full-AR  & $0.717$ & $0.716$ & $0.717$ \\
C-10 / DeiT-S s$_0$            & Full-AR  & $0.653$ & $0.652$ & $0.649$ \\
C-10 / DeiT-S s$_1$            & Full-AR  & $0.658$ & $0.655$ & $0.654$ \\
\bottomrule
\end{tabular}
\end{table}

Taken together, these four panels explain why AR-CondCal's ECE
movements over 1-D baselines are within seed variance and why the
per-method seed-std bands in Tab.~\ref{tab:calibration_block_ar}
overlap across methods. The naive full-entropy-vector MLP can
appear to improve over a linear confidence baseline, but as the
capacity-controlled audit in App.~\ref{app:probe-sanity}
establishes, this apparent uplift disappears under capacity-matched
confidence-only and shuffled-routing controls; we therefore do not
read the per-layer entropy vector as carrying recoverable
routing-specific signal beyond confidence under the probes
evaluated here.

\begin{figure}[h]
  \centering
  \textbf{(a) CIFAR-10}\\[-0.3em]
  \includegraphics[width=\linewidth]{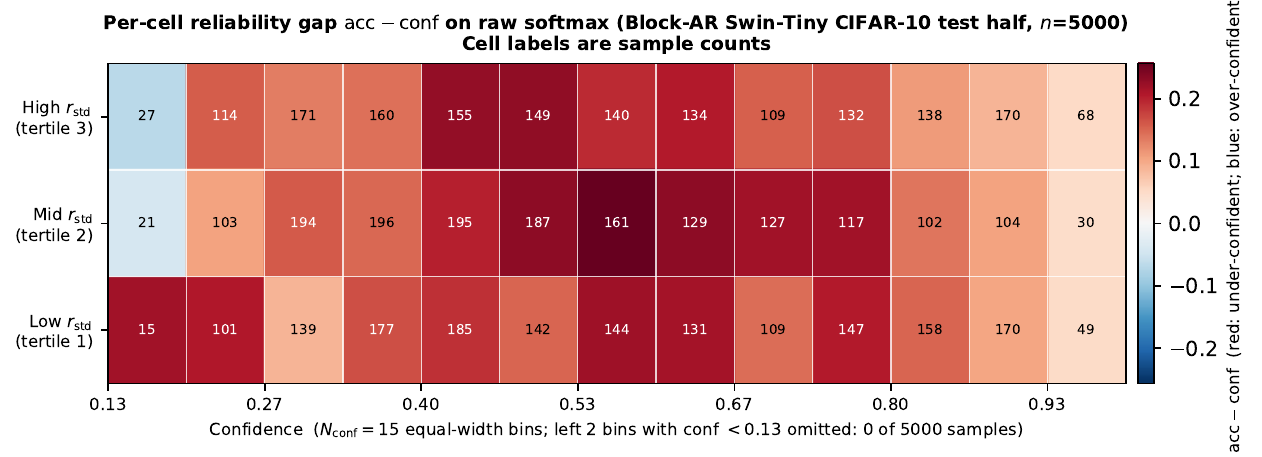}\\[0.5em]
  \textbf{(b) CIFAR-100}\\[-0.3em]
  \includegraphics[width=\linewidth]{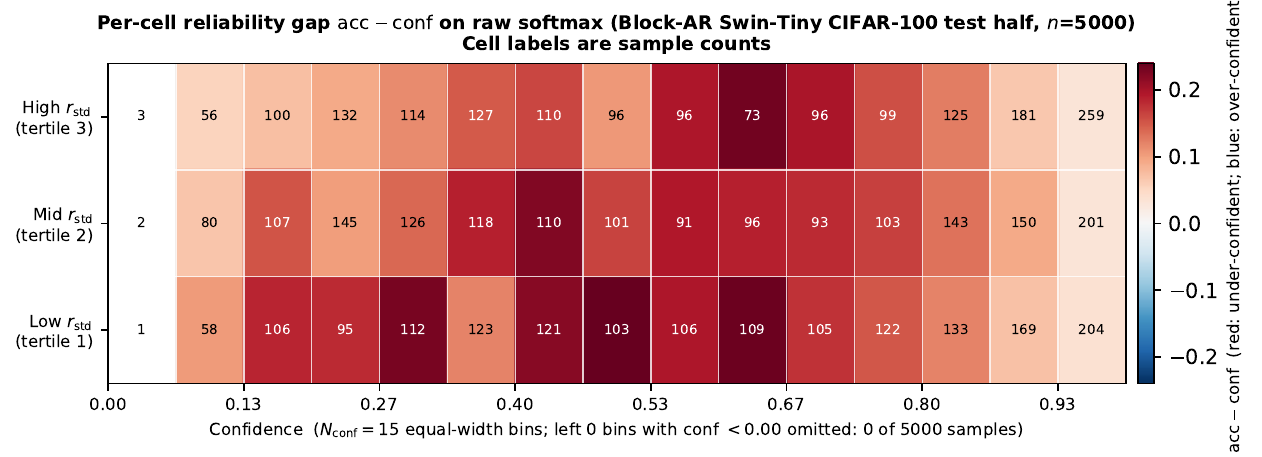}
  \caption{2-D reliability heatmaps on the raw softmax of Block-AR
  Swin-Tiny on (a) CIFAR-10 and (b) CIFAR-100, both on the
  $n\!=\!5000$ test half. Rows are tertiles of the depth-variance
  feature $r_{\mathrm{std}}(x)$ used by AR-CondCal; columns are the
  $15$ equal-width confidence bins (CIFAR-10 trims the $2$
  unpopulated leftmost bins, conf $<\!0.13$). Colour is the per-cell
  reliability gap $\mathrm{acc}-\mathrm{conf}$ (red: under-confident,
  blue: over-confident); cell annotations are sample counts; ``$\cdot$''
  marks cells with $<\!5$ samples (rendered with no gap value to
  suppress single-sample noise). \emph{Panel (a):} bin-weighted
  per-tertile ECE $0.161/\mathbf{0.181}/0.156$ (low/mid/high), which
  reproduces the $\mathbf{0.1814}$ raw worst-tertile ECE of
  Tab.~\ref{tab:calibration_block_ar} to four decimals (empty-bin
  trimming has no numerical effect). \emph{Panel (b):} bin-weighted
  per-tertile ECE $\mathbf{0.151}/0.134/0.112$ (low/mid/high) on
  CIFAR-100; the worst tertile is the \emph{low} band
  here, the opposite end of the $r_{\mathrm{std}}$ axis from CIFAR-10,
  reinforcing the $r_{\mathrm{std}}$ axis carries no fixed
  miscalibration direction across substrates.}
  \label{fig:reliability-2d}
\end{figure}

Figure~\ref{fig:reliability-2d} makes the \emph{per-cell} reliability
structure that underlies the worst-tertile metric directly visible.
Each cell is one (confidence bin, $r_{\mathrm{std}}$ tertile) pair on
the test half; its colour is $\mathrm{acc}-\mathrm{conf}$ in that cell,
and its annotation is the sample count feeding the bin. Three things
are worth noting. (i)~The dominant pattern on raw softmax is
\emph{under}-confidence in the mid-confidence range (red cells around
$\mathrm{conf}\!\in\![0.27, 0.67]$), not the over-confidence regime
most commonly flagged in the literature. (ii)~The mid-tertile row
concentrates the largest gaps, which is the row responsible for
worst-tertile ECE $=0.181$ on the raw softmax (matches the
\texttt{Raw} row of Tab.~\ref{tab:calibration_block_ar}); the
worst-tertile-ECE metric is therefore picking up a localised
reliability pattern in this cell, rather than only a binning
artifact. (iii)~High-confidence bins
(rightmost two columns) are relatively under-populated ($\le 170$
samples per cell) and show smaller gaps uniformly across tertiles.
This is the regime where scalar-temperature methods like TS
concentrate their fix. A kernel calibrator bandwidth on the $r_{\mathrm{std}}$ axis of
order the tertile spacing therefore smooths the worst row toward the
adjacent two, trading local correction for global mean: this is the
concrete geometry behind the ``bandwidth $\ge$ tertile spacing''
argument in Sec.~\ref{sec:discussion}.

\begin{table}[!ht]
\centering
\setlength{\tabcolsep}{3.5pt}
\small
\caption{Feature ablation across seven candidate $1$-D routing summaries (Swin-Tiny seed-$0$, $b\!=\!2$ for Block-AR, $2$-D Nadaraya--Watson estimator with Scott's-rule bandwidth, only the second feature differs). $\Delta$ columns are Block-AR ECE differences vs the Conf-only and Conf$+$PE controls within each dataset (positive $=$ this row achieves a lower ECE than the baseline). The point-estimate ECE range across the seven $1$-D summaries is narrow within every substrate ($0.0061$ on C-10 Block-AR, $0.0050$ on C-10 Full-AR, $0.0030$ on C-100 Block-AR, $0.0071$ on C-100 Full-AR), consistent with \S\ref{sec:rq2}: no scalar projection of routing uncertainty meaningfully separates from a confidence-only baseline once the $1$-D projection has discarded the across-depth coupling. The winning $1$-D summary changes across substrates, so $r_{\mathrm{std}}$ is best read as a fixed depth-heterogeneity probe used by AR-CondCal rather than a post-hoc selected winner.}
\label{tab:feature_ablation}
\begin{tabular}{p{4.5cm}rrrr}
\toprule
Feature & Block-AR ECE $\downarrow$ & Full-AR ECE $\downarrow$ & $\Delta$ vs Conf-only & $\Delta$ vs Conf+PE \\
\midrule
\multicolumn{5}{l}{\textbf{CIFAR-10} (Swin-Tiny seed-$0$, $b\!=\!2$ for Block-AR)} \\
\midrule
Confidence only & 0.0131 & 0.0232 & +0.0000 & $-$0.0039 \\
Predictive entropy & 0.0092 & 0.0199 & +0.0039 & +0.0000 \\
Aggregate routing entropy $r_{\mathrm{agg}}$ & 0.0099 & 0.0241 & +0.0032 & $-$0.0007 \\
Last-layer routing entropy $H_L$ & 0.0075 & 0.0191 & +0.0056 & +0.0017 \\
Routing concentration $1\!-\!r_{\mathrm{agg}}$ & 0.0099 & 0.0241 & +0.0032 & $-$0.0007 \\
Routing entropy $\times$ confidence & 0.0136 & 0.0207 & $-$0.0005 & $-$0.0044 \\
Depth-variance $r_{\mathrm{std}}$ & 0.0114 & 0.0201 & +0.0017 & $-$0.0022 \\
\midrule
\multicolumn{5}{l}{\textbf{CIFAR-100} (Swin-Tiny seed-$0$, $b\!=\!2$ for Block-AR)} \\
\midrule
Confidence only & 0.0145 & 0.0191 & +0.0000 & +0.0015 \\
Predictive entropy & 0.0160 & 0.0156 & $-$0.0015 & +0.0000 \\
Aggregate routing entropy $r_{\mathrm{agg}}$ & 0.0151 & 0.0197 & $-$0.0006 & +0.0009 \\
Last-layer routing entropy $H_L$ & 0.0157 & 0.0218 & $-$0.0012 & +0.0003 \\
Routing concentration $1\!-\!r_{\mathrm{agg}}$ & 0.0151 & 0.0197 & $-$0.0006 & +0.0009 \\
Routing entropy $\times$ confidence & 0.0130 & 0.0147 & +0.0015 & +0.0030 \\
Depth-variance $r_{\mathrm{std}}$ & 0.0137 & 0.0212 & +0.0008 & +0.0023 \\
\bottomrule
\end{tabular}
\end{table}

\subsection{Extended discussion of \S\ref{sec:discussion}}
\label{app:paradigm}

\textbf{Why scalar routing projections fail.} AR-CondCal's
failure is consistent with the loss induced by scalar
projection. Compressing $(H_1, \ldots, H_L)$ to a single scalar
(e.g., $r_{\mathrm{std}}$, or alternatives in
Tab.~\ref{tab:feature_ablation}) leaves a $1$-D projection
dominated by pipeline and seed noise (\S\ref{sec:rq2}). Distance-based kernel smoothing on this
projection then smooths over adjacent subgroups
(Proposition~4, App.~\ref{app:theory}: bandwidth-induced
subgroup-gap shrinkage by $\tanh(d^2/4h^2)$, and on Block-AR
$\sigma(r_{\mathrm{std}})\!\approx\!0.03$ places the Scott's-rule
bandwidth in the high-shrinkage regime). The $1$-D Spearman
correlation
$\rho(r_{\mathrm{std}}, |\mathrm{conf}{-}\mathrm{correct}|)\!=\!{-}0.010$
($p\!=\!0.34$) confirms that no monotonic scalar reading of the
routing vector recovers the signal.

\textbf{The full-profile probe also does not isolate
routing-specific signal.} Bypassing the $1$-D projection with a
full-vector MLP appears to give a positive $R^2$ uplift over a
linear confidence baseline. The capacity-matched audit in
App.~\ref{app:probe-sanity} shows this apparent uplift is not
routing-specific: a confidence-only MLP at the same architecture
matches or beats the full-vector MLP across all $10$ evaluated
substrate groups, and shuffling the routing profile across the
training fold leaves the held-out $R^2$ essentially unchanged. We
therefore do not claim that $(c, H_1, \ldots, H_L)$ exposes a
clean calibration signal under the probes tested. Whether other
internal-dynamics descriptors or non-AR
routing mechanisms expose cleaner signal under the same
diagnostic is left for future work.

\section{Cross-dataset replication of the temporal pilot
(companion to \S\ref{sec:probe})}
\label{app:temporal-cross-dataset}

The appendix-only temporal pilot complements the mechanistic
autopsy around Fig.~\ref{fig:mechanistic-autopsy} and
App.~\ref{app:mechanism}: deep AR layers undergo routing-entropy
collapse during training, and last-$K$-epoch temporal ensembling
does not rescue worst-tertile calibration on the tested cells.
The pilot is established on Swin-Tiny~$+$~Block-AR seed-$1$, the
cell whose mechanistic autopsy is shown in
Fig.~\ref{fig:mechanistic-autopsy}. We re-ran the same temporal
pilot (stride-$4$ ckpt sampling over $300$ epochs; $K\!=\!5$
ensemble of the last $5$ sampled epochs) on the corresponding
Swin-Tiny~$+$~Block-AR CIFAR-100 cell (seed-$1$). The two key
qualitative signatures replicate verbatim.

\begin{figure}[h]
  \centering
  \includegraphics[width=\linewidth]{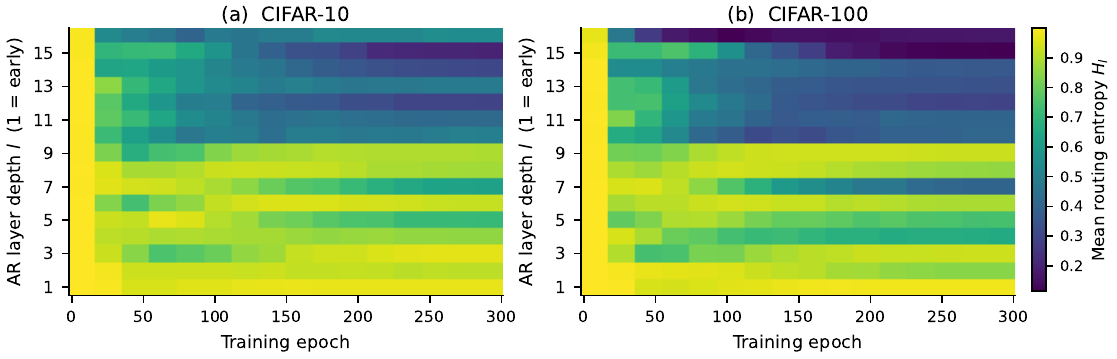}
  \caption{\textbf{Routing-entropy evolution heatmaps across
  training} for Swin-Tiny~$+$~Block-AR seed-$1$ on
  \textbf{(a)}~CIFAR-10 and \textbf{(b)}~CIFAR-100.
  Both panels share the same colour scale. The deep-layer
  collapse to near-deterministic routing emerges at the same
  qualitative stage of training and the early layers retain
  dispersion in both datasets, indicating the collapse is
  consistent with an AR-pathway effect under this training
  regime rather than a CIFAR-10-specific artifact.}
  \label{fig:temporal-heatmaps}
\end{figure}

\begin{figure}[h]
  \centering
  \includegraphics[width=\linewidth]{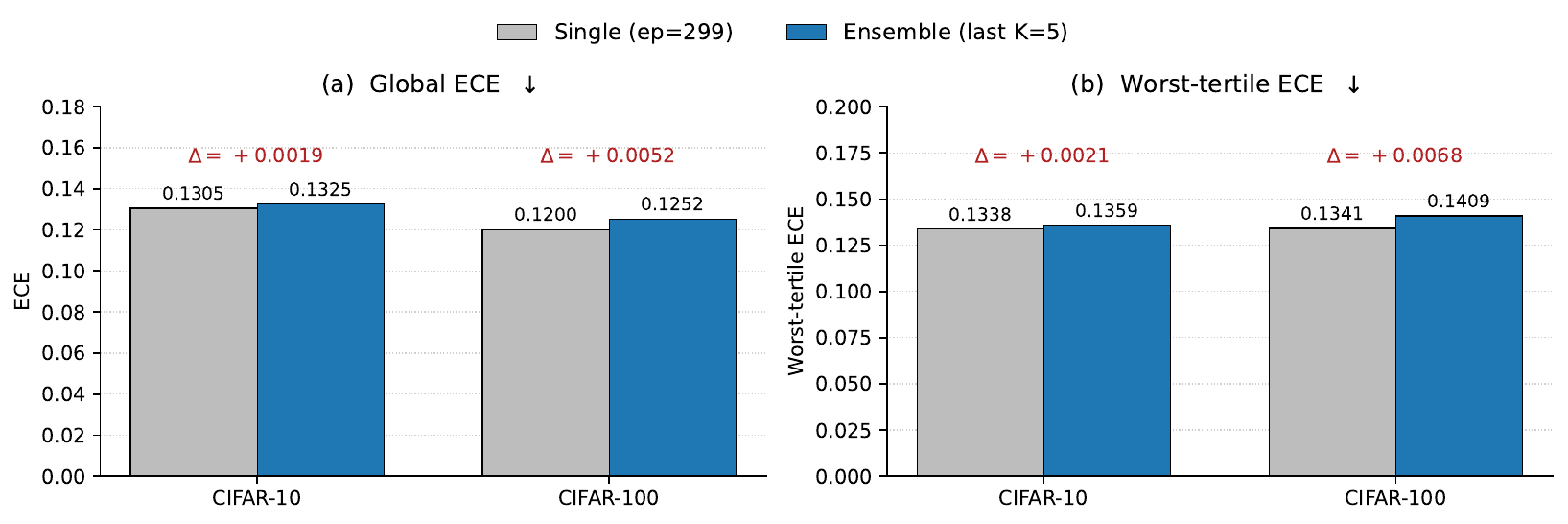}
  \caption{\textbf{Last-$K$-epoch ensemble vs single ep$=$299
  baseline} on the same two cells. Each panel reports the
  raw-softmax metric (no calibrator fit). The ensemble does
  \emph{not} reduce either metric; the worst-tertile delta is
  $+0.0021$ on CIFAR-10 and $+0.0068$ on CIFAR-100 (both worse
  than the single-epoch baseline). Temporal smoothing of
  raw-softmax outputs therefore does not reduce worst-tertile
  ECE on either tested dataset.}
  \label{fig:temporal-ensemble}
\end{figure}

A consistent direction of effect on both datasets is used here,
together with Fig.~\ref{fig:mechanistic-autopsy} and
App.~\ref{app:mechanism}, to characterise the collapse as tied to
the AR routing mechanism under this training regime, rather than
to a single-dataset coincidence.

\begin{figure}[h]
    \centering
    \includegraphics[width=\textwidth]{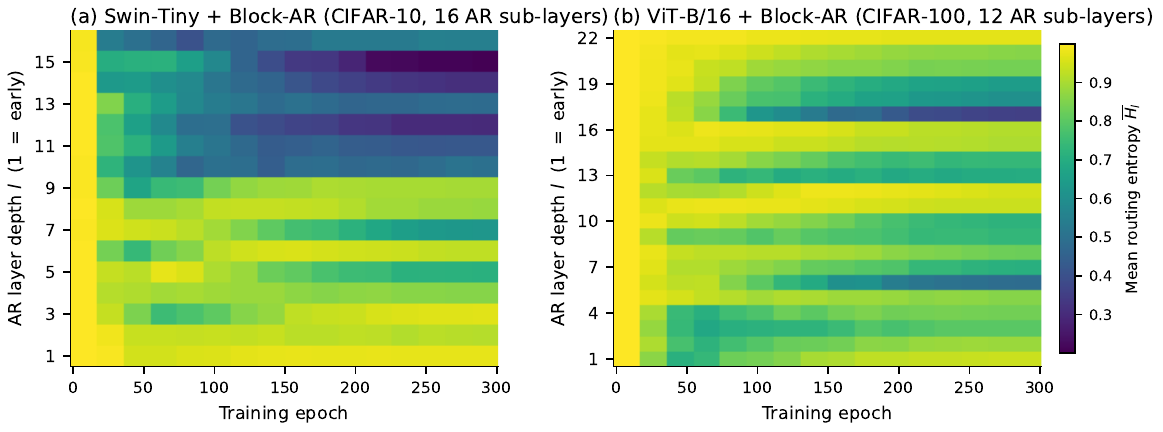}
    \caption{\textbf{Cross-architecture replication of deep-layer
    routing collapse.} (a) Swin-Tiny $+$ Block-AR (CIFAR-10, $16$ AR
    sub-layers). (b) ViT-B/$16$ $+$ Block-AR (CIFAR-100, $12$ AR
    sub-layers). Both the hierarchical Swin architecture and the
    homogeneous ViT architecture exhibit severe routing-entropy
    collapse in deeper sub-layers as training progresses. This is
    consistent with the narrative in \S\ref{sec:probe} that the
    collapse is tied to the AR routing mechanism under this
    training regime, rather than to a specific backbone's stage
    design.}
    \label{fig:cross-arch-heatmap}
\end{figure}

\section{External-runs audit summary (appendix-only)}
\label{app:external}

We additionally include two external checkpoint collections (a
ViT-B/16 finetune suite and a scratch-trained custom small-ViT
suite on CIFAR-10 / CIFAR-100 / Tiny-ImageNet) produced by an
independent training pipeline under different protocol
conventions. They are reported here only as cross-substrate
sanity checks and do not support any main-paper claim. The
finding consistent with \S\ref{sec:rq2} is that point-estimate
ordering of AR variants is unstable across these
incompatibilities; absolute ECE numbers are not directly
comparable to our main protocol.

\begin{table}[!ht]
\centering
\setlength{\tabcolsep}{5pt}
\small
\caption{External reproduction on ViT-B/16 \emph{finetune} (imported
from an independent training pipeline; see
Appendix~\ref{app:external} and the code/result provenance note in
Appendix~\ref{app:provenance}). These numbers were produced under a
different calibration-split protocol (dev-split from train, not our
$50/50$ test-split with seed~$42$), different metric and
routing-feature implementations, and early-stopped on validation ECE
rather than taken at the last epoch. They are \emph{not} directly
comparable to Tab.~\ref{tab:calibration_block_ar}; we include them
only as a cross-regime sanity check. ``--'' marks cells where no
external run was available.}
\label{tab:external-ft-vit-b16}
\begin{tabular}{llrrrr}
\toprule
Variant & Dataset & Acc $\uparrow$ & ECE $\downarrow$ & NLL $\downarrow$ & Brier $\downarrow$ \\
\midrule
Baseline & CIFAR-10 & 0.985 & 0.0045 & 0.048 & 0.023 \\
Block-AR & CIFAR-10 & 0.986 & 0.0042 & 0.049 & 0.022 \\
Full-AR & CIFAR-10 & 0.986 & 0.0034 & 0.048 & 0.022 \\
Baseline & CIFAR-100 & 0.910 & 0.0127 & 0.317 & 0.135 \\
Block-AR & CIFAR-100 & 0.914 & 0.0099 & 0.301 & 0.129 \\
Full-AR & CIFAR-100 & 0.909 & 0.0194 & 0.321 & 0.136 \\
Baseline & Tiny-ImageNet & 0.885 & 0.0120 & 0.460 & 0.166 \\
Block-AR & Tiny-ImageNet & 0.887 & 0.0127 & 0.452 & 0.164 \\
Full-AR & Tiny-ImageNet & 0.888 & 0.0107 & 0.457 & 0.164 \\
\bottomrule
\end{tabular}
\end{table}

\begin{table}[!ht]
\centering
\setlength{\tabcolsep}{5pt}
\small
\caption{External reproduction on a \emph{scratch-trained ViT}
(imported from an independent training pipeline). The backbone is a
custom scratch ViT shipped by that pipeline, not a conventional
ViT-B/16 (which in our implementation is only available with a
pre-trained initialisation). The training regime matches the main
paper (scratch, not finetune), but the calibration-split protocol
(dev-split $0.10$ from train, not our $50/50$ seed-$42$),
routing-feature implementation (aggregate routing entropy rather than
the per-layer depth-variance feature used in
Sec.~\ref{sec:method-arcondcal}), and checkpoint-selection rule (best
by val-ECE rather than last epoch) all differ. They are \emph{not}
directly comparable to Tab.~\ref{tab:calibration_block_ar}; we
include them as cross-substrate evidence only. Missing cells are
marked ``--''.}
\label{tab:external-scr-vit}
\begin{tabular}{llrrrr}
\toprule
Variant & Dataset & Acc $\uparrow$ & ECE $\downarrow$ & NLL $\downarrow$ & Brier $\downarrow$ \\
\midrule
Baseline & CIFAR-10 & -- & -- & -- & -- \\
Block-AR & CIFAR-10 & 0.775 & 0.1765 & 0.810 & 0.352 \\
Full-AR & CIFAR-10 & 0.739 & 0.1930 & 0.919 & 0.406 \\
Baseline & CIFAR-100 & 0.667 & 0.1470 & 1.396 & 0.471 \\
Block-AR & CIFAR-100 & 0.677 & 0.1614 & 1.365 & 0.466 \\
Full-AR & CIFAR-100 & 0.666 & 0.1567 & 1.396 & 0.474 \\
Baseline & Tiny-ImageNet & 0.486 & 0.1454 & 2.304 & 0.673 \\
Block-AR & Tiny-ImageNet & 0.504 & 0.1619 & 2.249 & 0.664 \\
Full-AR & Tiny-ImageNet & 0.483 & 0.1667 & 2.349 & 0.689 \\
\bottomrule
\end{tabular}
\end{table}

Point-estimate ECE on the two external tables is directionally
consistent with the main paper's framing (no strong ordering
survives across datasets; cross-substrate ECE on the scratch-ViT
suite sits in the $[0.145,\,0.167]$ band with no clear winner), and
none of it strengthens any main-paper claim.

\textbf{Compressed external-run robustness audit.}
We additionally ran six external-cell stress tests on the same two
external suites ($40$ external cells in total). For brevity, the
underlying figures are deferred to supplementary material; the
findings, which collectively reinforce the paper's conservative
framing rather than upgrading any main claim, are summarised in
Tab.~\ref{tab:ext-robustness-compressed}.

\begin{table}[!ht]
\centering
\setlength{\tabcolsep}{4pt}
\small
\caption{\textbf{Compressed external-run audit (appendix-only).}
Headline numbers from six external-pipeline stress tests on the
same two suites used in Tab.~\ref{tab:external-ft-vit-b16} /
Tab.~\ref{tab:external-scr-vit}. None of these is protocol-matched
main-paper evidence; each is reported only to bound the scope of
external-cell interpretations a reviewer might attach.}
\label{tab:ext-robustness-compressed}
\begin{tabular}{p{4.0cm}p{8.4cm}}
\toprule
Check & Headline finding \\
\midrule
Cross-regime sanity (ViT-B/$16$ FT, $9$ cells) & Accuracy / ECE / NLL / Brier vary by dataset; no consistent variant ordering survives across cells. \\
Cross-substrate sanity (custom small-ViT scratch, $7/9$ cells) & Across CIFAR-100 / Tiny-ImageNet the three variants lie in the $[0.145, 0.167]$ ECE band; no clear winner. \\
Block-size ablation (small-ViT scratch C100) & Top-$1$ accuracy flat at $0.677$; ECE decreases monotonically from $0.166$ at bl1 to $0.151$ at bl3; not a Swin-Tiny / ViT-B/$16$ claim. \\
Calibration-split variance (DeiT-Tiny FT C100, $10$ random splits) & BBQ takes rank-$1$ on $7/10$ splits, AR-CondCal on $2/10$; mean ECE $0.0137$ (AR-CondCal) vs $0.0118$ (BBQ). Single-split external ordering is not a reliable summary. \\
Bootstrap on $3$ external Block-AR cells & $95\%$ bootstrap CIs of routing features overlap a random-null baseline; logit-norm cleanly exceeds the noise band on $2/3$ cells. External signal not statistically resolved. \\
Cross-cell ranking ($40$ external cells, seed $42$) & Rank-$1$ frequency on global ECE: BBQ $25\%$, AR-CondCal $17.5\%$, HB $15\%$, Conf-only $12.5\%$. AR-CondCal is competitive on a subset of cells but not an overall winner. \\
Cross-substrate Block-vs-Full ($5$ paired external cells) & Both variants' bootstrap CIs overlap the noise null; no significant Block-vs-Full ordering survives on these external substrates under a more stable bootstrap summary. \\
Routing vs logit-norm incremental lift ($10$ external cells) & Adding routing-entropy on top of logit-norm hurts ECE on $7/10$ cells, hurts NLL / Brier on $9/10$ cells; helps by $\leq 0.004$ ECE on $3/10$. \\
Calibration-set-size sensitivity (DeiT-Tiny FT C100) & TS plateaus around ECE $0.025$ independent of $N$; AR-CondCal exceeds TS only at $N \geq 500$ on this single cell; Conf-only ($1$-D) dominates NLL at every $N$. Qualitatively mirrors the Swin-Tiny calibration-set-size pattern in Fig.~\ref{fig:calsize}; not a general method-win claim. \\
\bottomrule
\end{tabular}
\end{table}

\subsection{Generality check on vanilla substrates}
\label{app:generality-vanilla}

To probe whether the scalar-vs-non-linear-vector gap of
\S\ref{sec:probe} reflects a property of the AR routing-entropy
descriptor or a more general property of internal-state probing,
we ran the same MLP-probe diagnostic on three \emph{non-routing}
ImageNet-1k pretrained backbones, each evaluated on the
ImageNet-LT test split ($n\!=\!5{,}000$ subsample, seed $42$,
single seed). The vector descriptor is the per-block hidden-state
summary obtained by hooking each backbone's natural block class
(Swin-Base: $24$ \texttt{SwinTransformerBlock}s; ViT-Large: $24$
attention \texttt{Block}s; DeiT-Base: $12$ attention
\texttt{Block}s) --- it is \emph{not} the AR per-layer routing
entropy, since none of these backbones contains
Attention-Residual layers.
\begin{center}
\setlength{\tabcolsep}{4pt}
\small
\begin{tabular}{lccccc}
\toprule
Backbone & Acc@$1$ & Base ECE & Conf-only $R^{2}$ & MLP $R^{2}$ & Uplift \\
\midrule
Swin-Base   & $0.846$ & $0.086$ & $0.381$ & $0.513$ & $+0.131$ \\
ViT-Large   & $0.843$ & $0.025$ & $0.377$ & $0.468$ & $+0.091$ \\
DeiT-Base   & $0.835$ & $0.061$ & $0.297$ & $0.393$ & $+0.096$ \\
\bottomrule
\end{tabular}
\end{center}
The full-vector MLP achieves $R^{2}$ uplifts ranging from
$+0.091$ (ViT-Large) to $+0.131$ (Swin-Base) over the
confidence-only \emph{linear} baseline. This is the same
capacity-uncontrolled comparison that App.~\ref{app:probe-sanity}
shows is not routing-specific on AR substrates; we did not run
the capacity-matched controls on these non-AR backbones (no
routing trajectory is defined to shuffle), so these numbers are
reported only for transparency, not as evidence of non-linear
internal-state signal recovery. The within-bin
permutation null is \emph{not} rejected on Swin-Base
($p\!=\!0.531$) or DeiT-Base ($p\!=\!0.346$), but \emph{is}
rejected on ViT-Large ($p\!=\!0.021$). We report this only as a
single-seed parallel observation; these numbers do not feed any
main-paper count claim.

\subsection{Split-stability and ranking-stability subsection (compressed)}
\label{app:ext-robustness}

The full figures for this subsection (calibration-split
variance, bootstrap, cross-cell ranking) are deferred to
supplementary material; the corresponding headline numbers appear in
Tab.~\ref{tab:ext-robustness-compressed} above. None of these
external-cell results is used to strengthen any main-paper claim.

\section{Training and evaluation details}
\label{app:details}
All \emph{main-paper} models (i.e., the Swin-Tiny, DeiT-Small, and
ViT-B/16 cells produced by our own training pipeline, as named in
Appendix~\ref{app:provenance}) are trained from scratch for $300$
epochs with mixed-precision training (AMP at
O1)~\citep{micikevicius2017mixed}, a cosine-annealed learning-rate
schedule~\citep{loshchilov2016sgdr}, and a $20$-epoch
warmup~\citep{goyal2017accurate}; full hyperparameters are reproduced
in the released code. Calibration is evaluated on a $50/50$
calibration/test split of the official test set with fixed
seed~$42$. Equal-width ECE uses $15$ bins unless otherwise noted.
The separate external result collections audited in
Appendix~\ref{app:external} were produced by an independent
training pipeline under a different protocol (different calibration
split, different checkpoint-selection rule, and (for the ViT-B/16
rows there) a finetune rather than scratch regime); those
settings are documented in Appendix~\ref{app:external} and are not
covered by the description above.

\section{Implementation notes}
\label{app:impl}
Routing weights $\alpha_l$ are extracted by temporarily enabling a cache on
each Attention-Residual sub-layer module; no re-training is required.
The reported AR-CondCal runs use a label-free min--max rescaling of
the routing cache before the fixed 50/50 split. This transform uses
routing values only and no labels; the Nadaraya--Watson fit,
correctness targets, and Scott bandwidth statistics are computed on
$\mathcal{D}_{\mathrm{cal}}$. A cal-only-min/max audit produced
bit-identical Tab.~\ref{tab:calibration_block_ar} values for the
matched-kernel NW rows, so the paper's matched-control conclusions are
unchanged. The main AR-CondCal row uses Scott's-rule bandwidths;
Appendix~\ref{app:bw-sensitivity} reports a bandwidth-sensitivity
sweep over Scott multiples, CV-NLL, and a global-ECE oracle. The AR-CondCal
bisection target $\tilde c(x)$ is projected to
$[1/K\!+\!\epsilon,\, 1\!-\!\epsilon]$ with $\epsilon = 10^{-6}$
before solving for $\tau(x)$ (\S\ref{sec:ar-condcal}); on the
evaluated cells the pooled clipping rate is $0.08\%$ total.

\section{Reproducibility checklist}
\label{app:reproducibility}
\textbf{Training.} 300 epochs on Swin-Tiny / DeiT-Small /
ViT-B/$16$, batch $128$, AdamW (lr $5\!\times\!10^{-4}$, wd
$0.05$), cosine schedule with $20$-epoch
warmup~\citep{loshchilov2016sgdr,goyal2017accurate}, AMP at
O1~\citep{micikevicius2017mixed}.
\textbf{Evaluation.} A $50/50$ cal/test split of the standard
test set with seed $42$; bootstrap CIs on the test half use
$500$ resamples; permutation tests use $5{,}000$ resamples
(\S\ref{sec:rq1}).
\textbf{AR-CondCal.} 2-D Nadaraya--Watson on
$(c, r_{\mathrm{std}})$, per-axis Scott's-rule bandwidths on
$\mathcal{D}_{\mathrm{cal}}$ (\S\ref{sec:ar-condcal}); the
kernel-predicted target is projected to
$[1/K\!+\!\epsilon,\, 1\!-\!\epsilon]$ before bisection.
\textbf{Diagnostic MLP probe.} 2-layer ReLU MLP, hidden $16$,
Adam lr $10^{-2}$ wd $10^{-4}$, $200$ epochs, target
$|\mathrm{conf}(x)\!-\!\mathrm{correct}(x)|$, single $50/50$
split with seed $42$ matching Tab.~\ref{tab:routing-probe}.
\textbf{Bandwidth sensitivity.} Multipliers $\{0.25, 0.5, 1, 2,
4\}$; CV-NLL uses $5$-fold cal-set CV; Oracle-ECE selected on
the held-out test half (App.~\ref{app:bw-sensitivity}).

\section{Code and result provenance}
\label{app:provenance}

To distinguish the main-paper results from the external appendix-only
evidence audited in Appendix~\ref{app:external}, we name here the two
codebases used to produce the main-body numbers and the metadata that
a reader can verify against the supplementary artifact.

\textbf{Main-paper training.}
Every checkpoint that feeds the main-body figures and tables,
including Tab.~\ref{tab:phenomenon},
Tab.~\ref{tab:calibration_block_ar}, Fig.~\ref{fig:priority2},
Fig.~\ref{fig:bench-main}, Fig.~\ref{fig:mechanistic-autopsy},
and the appendix diagnostic sweep in
Tab.~\ref{tab:replication-summary} / Tab.~\ref{tab:ar-sweep} /
Tab.~\ref{tab:routing-probe}, was trained from scratch on our
internal training codebase. The training entrypoint reads the
model-family-specific YAML configuration; every hyperparameter
visible in Appendix~\ref{app:details} is set in that YAML, not
overridden from the command line. Seed, batch size, and
checkpoint-selection rule (last epoch) are fixed in the YAML and do
not vary across the broader set of main-paper training runs
used for raw-model and seed-variance analyses.

\textbf{Routing extraction and calibration evaluation.}
Routing weights, the per-layer routing entropy $H_l$, the
depth-variance feature $r_{\mathrm{std}}(x)$, the equal-width /
adaptive-mass / smooth ECE statistics, the worst-tertile ECE, and
the $500$-resample bootstrap CIs are produced by an evaluation
module within the same internal codebase, organised as a
separate code path from the training entrypoint: it consumes
checkpoints and routing-weight caches and writes metrics JSON
files, with no training state of its own. Routing extraction,
calibration benchmarking with bootstrap CIs, and paper-table
assembly are described, with their input/output contracts, in
the supplementary README.

\textbf{Commit-level reproducibility.}
The supplementary artifact for this submission records the repository
commit hash at the time each result family was generated. Unlike the external runs audited in
Appendix~\ref{app:external} (whose metadata carries a null training
commit and therefore cannot be rebuilt exactly by a third party),
every main-body number in this paper is tied to a specific
commit of the internal codebase above, and the
scripts, configuration files, and seeds needed to rebuild the number
from scratch are all included. We hold the external appendix-only
evidence to a weaker standard only because that is what the external
metadata supports; we hold our own numbers to this stronger standard
because that is what the submission can substantiate.

\textbf{Thresholds fixed in advance.}
Any future cross-architecture scope upgrade beyond the completed
protocol-matched AR sweep is gated on pre-registered thresholds
(a matched-confidence tertile gap of at least $0.08$ and
feature-ablation support for the pre-specified depth-variance
feature used by AR-CondCal) fixed in the submitted draft and
not adjusted after any corresponding training run completes. The
preliminary ViT-B/$16$ cell in Appendix~\ref{app:cross-arch-multiseed} already
fails the second half of this rule, so we do not claim a scope
upgrade in this submission.


\end{document}